\documentclass[10pt,twocolumn,letterpaper]{article}

\usepackage{iccv}
\usepackage{times}
\usepackage{epsfig}
\usepackage{graphicx}
\usepackage{amsmath}
\usepackage{amssymb}

\usepackage{kotex}
\usepackage{booktabs}
\usepackage{subfigure}
\usepackage{float}
\usepackage{comment}
\usepackage{mathtools}
\usepackage{amsthm}
\usepackage{color}
\usepackage{xcolor}
\usepackage{wrapfig}
\usepackage{ragged2e}
\usepackage{enumitem}
\usepackage{soul}
\usepackage[normalem]{ulem} %\sout to strike out
\usepackage{multirow}
\usepackage{xspace}
\usepackage{makecell}
\usepackage{array}
\usepackage{pifont}% http://ctan.org/pkg/pifont
\usepackage{csquotes}
\usepackage{balance}
\usepackage{algorithmic}
\usepackage{amsfonts}
\usepackage{algorithm}
\usepackage[bottom]{footmisc}
\usepackage[pagebackref=true,breaklinks=true,letterpaper=true,colorlinks,bookmarks=false]{hyperref}

\iccvfinalcopy % *** Uncomment this line for the final submission

\def\iccvPaperID{9353} % *** Enter the ICCV Paper ID here

\ificcvfinal\fi

\begin{document}

%%%%%%%%% TITLE
\title{Influential Rank: A New Perspective of Post-training \\ for Robust Model against Noisy Labels}

\author{Seulki Park$^1$\quad Hwanjun Song$^{2}$ \quad Daeho Um$^1$ \quad Dae Ung Jo$^{1}$ \quad Sangdoo Yun \quad Jin Young Choi$^1$\\
% {\footnotesize $^1$ASRI, Dept. of Electrical and Computer Eng., Seoul National University, South Korea} \\
% {\footnotesize $^2$Personal Robotics Lab., Dept. of Electrical and Electronic Eng., Imperial College London, UK}\\
% {\footnotesize $^3$Intelligent Robotics Lab., School of Computer Science, University of Birmingham, UK}\\
% {\footnotesize $^4$Clova AI Research, NAVER Corp., South Korea}\\
{\hspace{1cm}$^1$ASRI, ECE., Seoul National University}
{\hspace{1cm}$^2$NAVER AI Lab}\\
{\footnotesize {seulki.park@snu.ac.kr, ghkswns91@gmail.com, \{daehoum1, mardaewoon\}@snu.ac.kr, sangdoo.yun@navercorp.com, 
jychoi@snu.ac.kr}}
% {\tt\scriptsize jwchoi.pil@gmail.com, \{hj.chang,t.fischer,y.demiris\}@imperial.ac.uk, \{yunsd101,kyuewang,jy.jeong,jychoi\}@snu.ac.kr}
% For a paper whose authors are all at the same institution,
% omit the following lines up until the closing ``}''.
% Additional authors and addresses can be added with ``\and'',
% just like the second author.
% To save space, use either the email address or home page, not both
}

\maketitle
% Remove page # from the first page of camera-ready.
\ificcvfinal\thispagestyle{empty}\fi

%%% new command

% for to-do list
\newlist{todolist}{itemize}{2}
\setlist[todolist]{label=$\square$}
\newcommand{\cmark}{\ding{51}}%
\newcommand{\xmark}{\ding{55}}%
\newcommand{\done}{\rlap{$\square$}{\raisebox{2pt}{\large\hspace{1pt}\cmark}}%
\hspace{-2.5pt}}
\newcommand{\wontfix}{\rlap{$\square$}{\large\hspace{1pt}\xmark}}

\newcommand{\overbar}[1]{\mkern 1.5mu\overline{\mkern-1.5mu#1\mkern-1.5mu}\mkern 1.5mu}
\newcommand{\eqdef}{\overset{\mathrm{def}}{=\joinrel=}}
\newcommand{\our}{Influential Rank\xspace} 
\newcommand{\ourbold}{\textbf{Influential Rank}\xspace} 

\newcommand{\floor}[1]{\left\lfloor #1 \right\rfloor}
\newcommand{\ceil}[1]{\left\lceil #1 \right\rceil}
\renewcommand{\algorithmicrequire}{\textsc{Input:}}
\renewcommand{\algorithmicensure}{\textsc{Output:}}
\renewcommand{\algorithmiccomment}[1]{/*~#1~*/}
\newcommand{\INDSTATE}[1][1]{\STATE\hspace{#1\algorithmicindent}}

\theoremstyle{definition}
\newtheorem{definition}{Definition}[section]
\newcommand\jy[1]{{\color{magenta}#1}}
\newcommand\sk[1]{{\color{blue}#1}}
\newcommand\dw[1]{{\color{orange}#1}}
\newcolumntype{L}[1]{>{\raggedright\let\newline\\\arraybackslash\hspace{0pt}}m{#1}}
\newcolumntype{X}[1]{>{\centering\let\newline\\\arraybackslash\hspace{0pt}}p{#1}}
% %%%%%%%%% ABSTRACT
%%%%%%
Deep neural network can easily overfit to even noisy labels due to its high capacity, which degrades the generalization performance of a model. 
To overcome this issue, we propose a new approach for learning from noisy labels (LNL) via \textit{post-training}, which can significantly improve the generalization performance of any pre-trained model on noisy label data.
To this end, we rather exploit the overfitting property of a trained model to identify mislabeled samples.
Specifically, our post-training approach gradually removes samples with high influence on the decision boundary and refines the decision boundary to improve generalization performance. 
Our post-training approach creates great synergies when combined with the existing LNL methods. 
Experimental results on various real-world and synthetic benchmark datasets demonstrate the validity of our approach in diverse realistic scenarios.
\vspace{-0.3cm}

% %%%%%%%%% BODY TEXT
%%%%  \todo update recent papers
\section{Introduction}
\vspace{-0.1cm}
% 1문단: Why LNL is important research field?
The current deep learning %breakthroughs have largely been due to `data'
has made a huge breakthrough because of `data'. Thus, many researchers in both academia and industry endeavor to obtain considerable data.
%One of the most popular methods to collect data today is crowd-sourcing\,\cite{ref:snow_amt_2008, ref:crowd_survey_2011} (e.g., Amazon Mechanical Turk) because it is cheap and quick.
However, real-world data inevitably contain some proportion of incorrectly labeled data, owing to perceptual ambiguity, or errors from human or machine annotations. %errors, or errors from automatic annotations.
These noisy labels negatively affect the generalization performance of a trained model since a deep neural network\,(DNN) can easily {overfit} to even noisy labels due to its high capacity\,\cite{ref:zhang_iclr_2017}.
Therefore, learning from noisy labels\,(LNL) has received much attention in recent years\,\cite{ref:hendrycks_gimpel_nips2018, ref:xu_ldmi_nips2019, ref:song_selfie_2019_icml, ref:zheng_chen_icml_2020, ref:volminnet_2021_icml, ref:Cheng_2022_CVPR, ref:Iscen_2022_CVPR} due to the increasing need to handle noisy labels in practice.

\begin{figure}[t!]
\centering
\includegraphics[width=0.9\columnwidth]{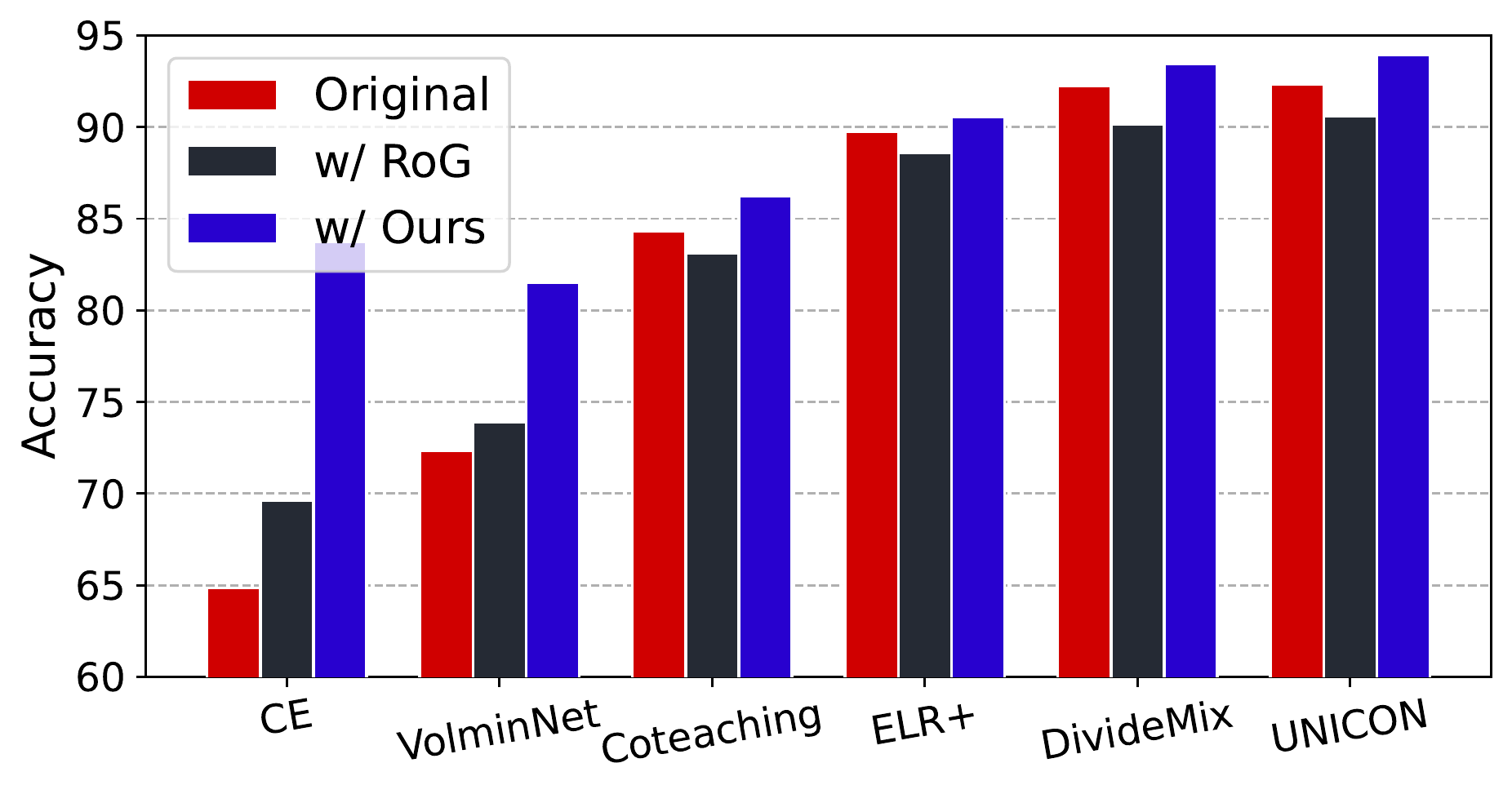}
\vspace*{-0.2cm}
\caption{\textbf{Test accuracy improvement over various methods on CIFAR-10N (Worst).}
As a post-training method, our proposed Influential Rank can improve various pre-trained models by large margin, compared to the post-processing baseline method, RoG~\cite{ref:rog_2019_icml}. The used CIFAR-10N (Worst) is a human-annotated real-world noisy dataset with about 40\% noise rate\,\cite{ref:cifar-n_2021}.
}
\label{fig:intro_comparison}
\vspace*{-0.6cm}
\end{figure}

% 최근에 pre-trained model에 대해 inference를 robust 하게 하는 post-inference? 방식이 제안됨. -> 얘는 모델을 직접 개선시키지 않고, gaussian 가정을 사용하기 때문에 corrupt 많이 된 모델에 대해서는 성능 개선에 한계가 있음.
% -> 이러한 문제들을 해결하기 위해 new post training based on overfitting property 를 제안함.
To handle noisy-label problem, prior literature aims to distinguish between clean and mislabeled data, and use this information to train a robust classifier during training.
To this end, prior works mainly rely on the assumption that the clean labels are more likely to have smaller losses before the model is overfitted~\cite{ref:arpit_memoriz_icml2017}.
However, due to the high capacity of deep neural networks (DNNs), DNNs can fit even noisy labels~\cite{ref:zhang_iclr_2017}; thus it is challenging to correctly detect mislabeled data during training.
Hence, various methods have been proposed to use more robust models before overfitting, such as leveraging the model with early stopping~\cite{ref:song2019does, ref:li2020gradient}, or using multiple networks with co-training for sample selection~\cite{ref:coteacing_nips_18, ref:coteachingplus_icml2019, ref:dividemix_iclr_2019}.
%Although these methods achieve a robust classification accuracy, they do not completely overcome the overfitting to noisy data due to their high capacity in learning~\cite{ref:zhang_iclr_2017}.

%3문단: 우리 방법 간단히 설명
%To overcome the aforementioned problems, 
Here, we introduce a different perspective against the mainstream research. 
We propose a new post-training LNL approach,
%In this paper, we propose a new LNL approach via \textit{post-training}
which can synergize with the model trained using prior robust methods, further enhancing the generalization capability of the model.
Given a pre-trained model, the proposed post-training scheme refines the model by exploiting the `overfitting property' of mislabeled samples.
`Overfitting property' of mislabeled samples is derived from two following intuitions. (1) Mislabeled samples are more likely to distort the decision boundary than clean samples. Thus removing the mislabeled samples is likely to sway the decision boundary significantly.
%, as shown in Figure~\ref{fig:intuition} (b). 
%However, since clean samples occupy the majority in a class, removing a few clean points hardly changes the model.
(2) The overfitted model predicts poorly on unseen data, and the mislabeled sample is usually the main culprit for the model to classify new data with incorrect labels.
%(Figure~\ref{fig:intuition} (c and d)).
The details on these intuitions are discussed in Section \ref{sec:inf_intuition}.

These intuitions on overfitting motivate us to propose a novel method named \ourbold, which leverages the samples' influence on the decision boundary and on unseen samples to enhance robustness.
%Specifically, the influence score is an estimate of the change in the model's parameters when the example does not affect the model training. 
%Specifically, the 
To this end, we propose {\it overfitting score on model} (OSM) and {\it overfitting score on data} (OSD).
OSM measures the influence of a training sample on changes in model parameters, and OSD measures the inconsistency of the sample's influence on the classification prediction for a small number of clean validation data.
Based on OSM and OSD, \our updates the trained model by removing high influential samples and mitigating their negative influence on the classifier.

% 4문단: 우리 방법의 장점들 & 실험 결과.
% confirmation bias에 낫다. 
% 1 post training 방법 처음 제안했다. 
% 1. post-training 이기 때문에 일반 모델이건, LNL 모델이건 어디에도 붙일 수 있다.
% 기존 LNL과 다르게 - orthogonal. 도움된다.
% 2. LNL방식 써도 효과적으로 제거되지 않고 남아있는 noisy label을 찾아내고 모델을 개선시킬 수 있어서, 그냥 일반 CE에 붙여도 다른 LNL 모델보다 좋은 성능을 보이기도 하며, 
% 기존 LNL 모델에 보완적으로 추가성능을 올릴 수도 있다.
% (비디오에도 가능)
% 3. 오버피팅 특성을 이용해서 decision boundary를 완화시키기 때문에 꼭 noisy label이 아니더라도, clean dataset에 사용할 경우 regularization 효과가 있는 것을 발견했다. 
Since the post-training provides a new information (\emph{i.e}., sample's influence) to any pre-trained models, \our can effectively improve robustness of existing LNL methods. %models.
Through extensive experiments on multiple benchmark data sets, we demonstrate the validity of our method, and show that \our can improve the performance of the model consistently whether or not it is pre-trained with LNL methods, as shown in Figure~\ref{fig:intro_comparison}.   
Furthermore, we show that \our is useful in two different applications other than LNL. The proposed overfitting scores can be effective for (1) data cleansing that filters out erroneous examples in \textit{real-world video data} and (2) \textit{regularization} that boosts the classification performance on clean data. 

Our key contributions can be summarized as follows:

\begin{itemize}[leftmargin=*,noitemsep,topsep=0pt]
\item \textbf{Post-training}: \our is a novel post-training approach for LNL, which leverages the \textit{overfitting scores} of training examples on the decision boundary.
\item \textbf{Practicability}: \our is applicable to any pre-trained models and works synergistically with other existing LNL methods.
%are used in pre-training.
%\item \textbf{Robustness}: \our achieves considerably lower test error on multiple real-world and synthetic noisy datasets compared with state-of-the-art methods.
\item \textbf{Extensibility}: \our can be easily extended to cleansing video dataset and a regularization for reducing overfitting arising from  clean but  influential samples.
\end{itemize}
%}

\begin{figure*}[t!]
\centering
    \subfigure[Overfitted model]
    {\includegraphics[width=0.24\linewidth]{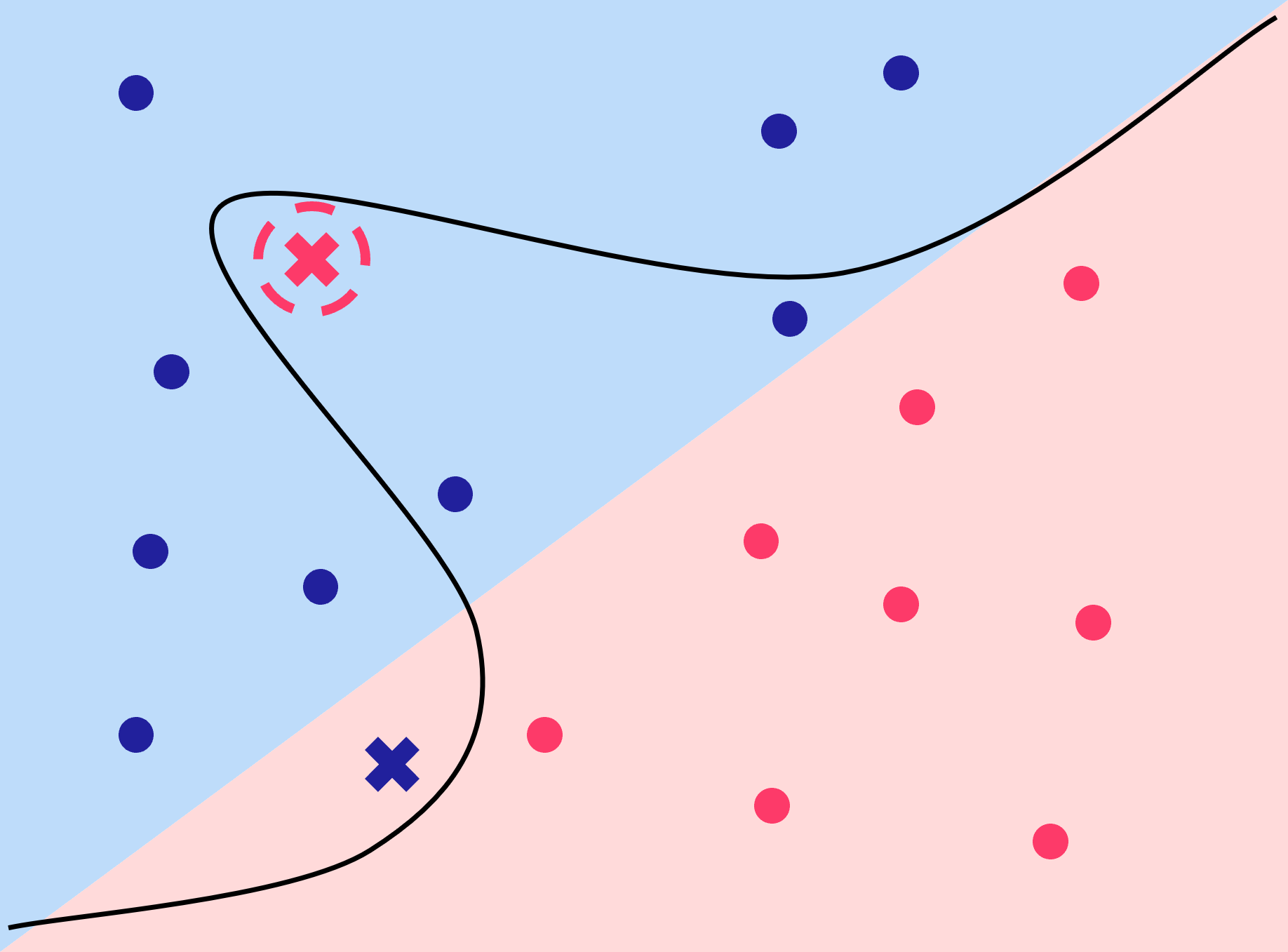}}
    %\hspace{0.05cm}
    \subfigure[Influence on model]{
        \includegraphics[width=0.24\linewidth]{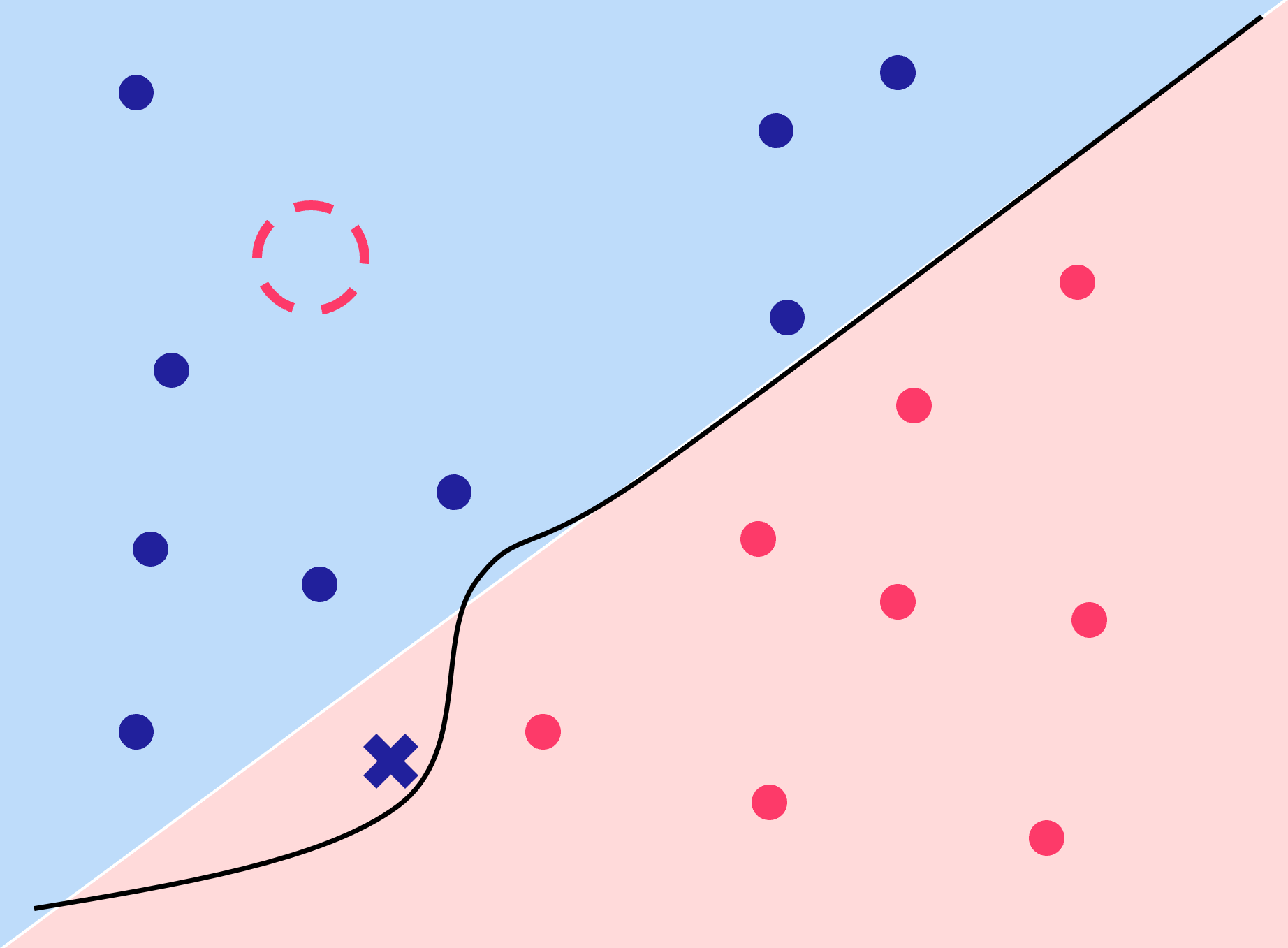}}
    %\hspace{0.05cm}
    \subfigure[Influence on data (blue)]{
        \includegraphics[width=0.24\linewidth]{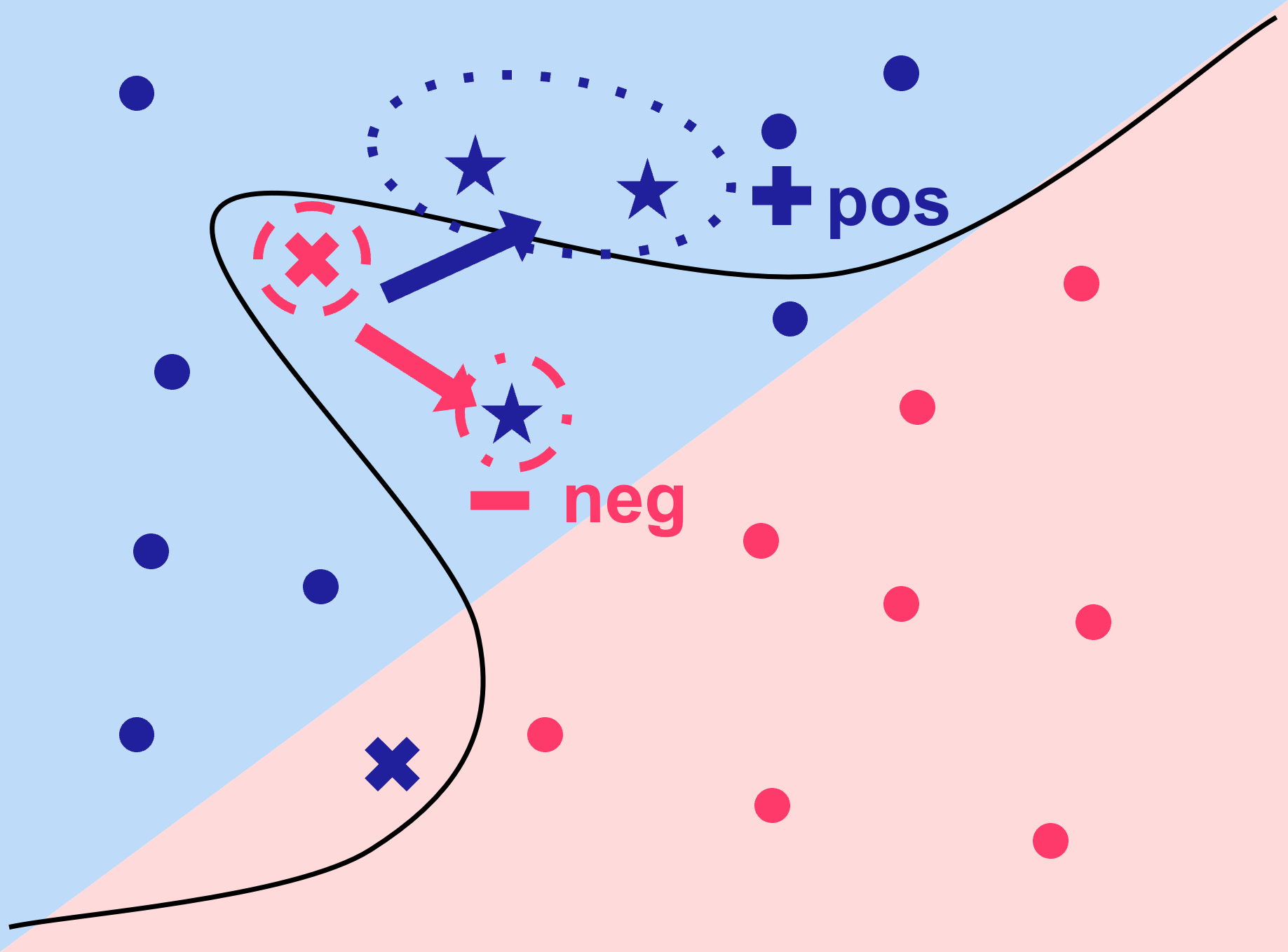}}
    %\hspace{0.05cm}
    \subfigure[Influence on data (red)]{
        \includegraphics[width=0.24\linewidth]{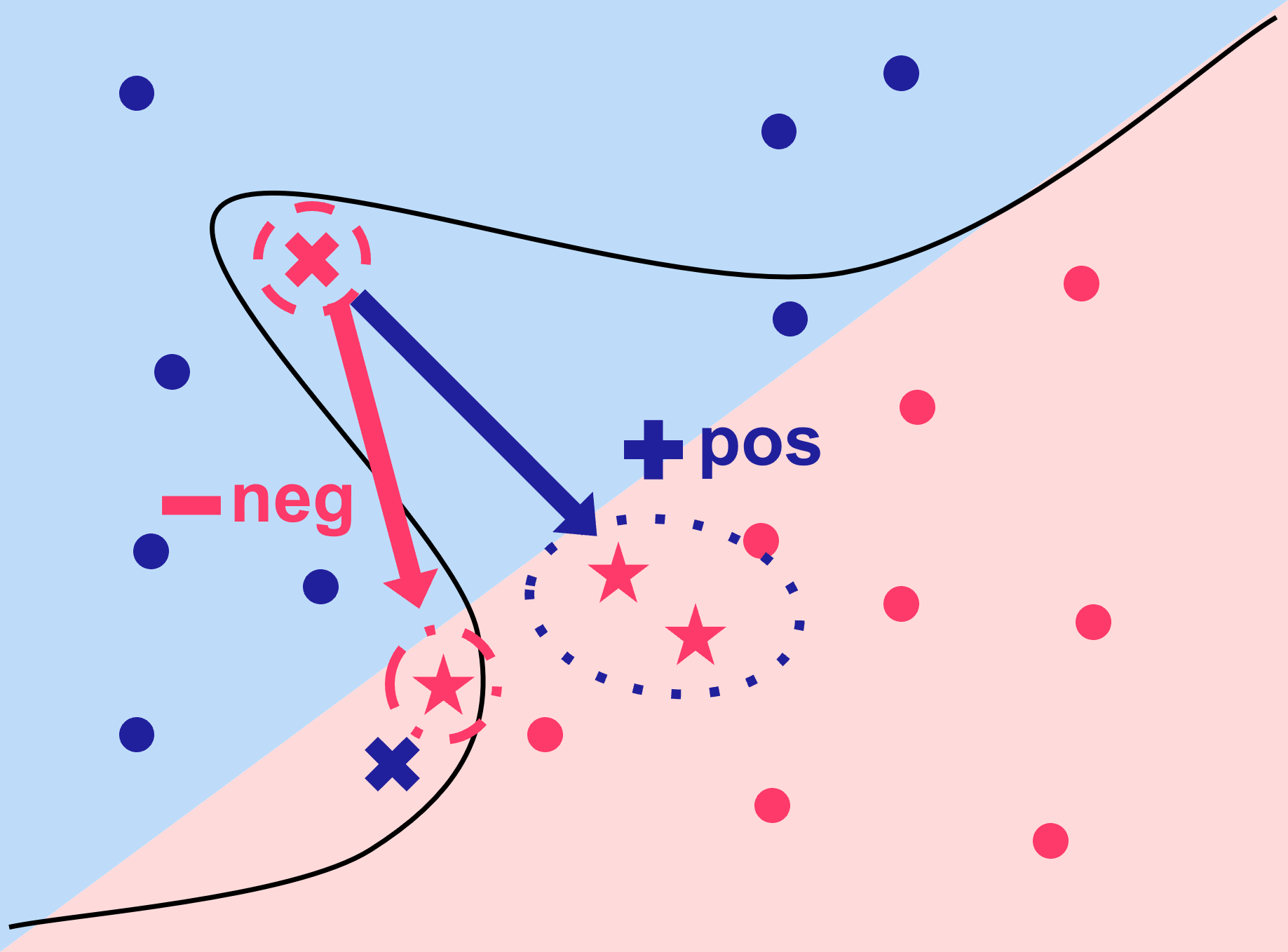}}
    %\vspace{-0.3cm}
\caption{\textbf{Our intuition.} The red and blue points belong to different classes in binary classification.
The $\times$ marks indicate mislabeled data.
(a) Due to the mislabeled samples ({$\times$}), the model is overfitted. 
(b) {$\times$} significantly affects the model because if the sample is removed, the parameter of the model is substantially changed.
(c and d) Assume clean validation data ($\bigstar$) are given.
The noisy-label sample ({$\times$}) exerts both positive and negative influences on correctly classifying the validation data in the same class, even when distances are near. 
The noisy-label data tend to have inconsistent effects on data within the same class.  
%Therefore, observing the training sample's abnormal influence on the model or clean validation data can provide an important clue for detecting noisy labels.
}
\label{fig:intuition}
\vspace*{-0.50cm}
\end{figure*}

\vspace*{-0.1cm}
\section{Related Work}
\vspace*{-0.1cm}

\subsection{Deep Learning with Noisy Labels}
Learning with noisy labels has two main research directions. 
One is to find and use only clean labels for training, and the other is to directly train a robust model on noisy labels. For a more thorough study on this topic, we refer the reader to survey\,\cite{ref:song2020learning, ref:frenay2013classification}.

\smallskip\smallskip
\noindent
\textbf{Noise-cleaning Approach.}
Most noise-cleaning approaches focus on finding small-loss examples before overfitting because DNNs learn easy samples first and gradually learn difficult samples  \cite{ref:arpit_memoriz_icml2017}.
To prevent overfitting of a neural network, some methods simultaneously train two neural networks and select small-loss examples\,\cite{ref:coteacing_nips_18, ref:coteachingplus_icml2019, ref:zhou2020robust, ref:song2019does, ref:Sun_2022_CVPR}, while others train a network guided by a teacher network\,\cite{ref:mentornet_icml_18, ref:zhang2020distilling}.
Meanwhile, O2U-Net\,\cite{ref:o2unet_iccv_19} adjusts the learning rate to take the model from overfitting to underfitting cyclically and records the losses of each example during the iterations. DivideMix\,\cite{ref:dividemix_iclr_2019} and SELF\,\cite{ref:nguyen2019self} incorporate semi-supervised learning with the small-loss trick for better sample selection.
Recently, UNICON~\cite{ref:unicon_2022} proposed uniform clean sample selection algorithm to tackle the class imbalance problem induced by prior sample selection methods.

\smallskip\smallskip
\noindent
\textbf{Noise-tolerant Approach.}
The noise-tolerant approach aims to train a robust model on a noisy-label dataset without removing the noise.
Some methods design noise-robust losses  \cite{ref:liu2020peer, ref:ma2020normalized, ref:zhang_gce_nips2018, ref:xu_ldmi_nips2019, ref:Iscen_2022_CVPR}, and others attempt to reweight losses  \cite{ref:lnl_nips2013, ref:patrini_qu_cvpr2017}.
Despite their theoretical justification, these approaches require mathematical assumptions or prior knowledge, such as known noise rates and class-conditional noise transition matrices, which make them challenging in practice. 
To tackle the difficulty in estimating transition matrix, Cheng et al.~\cite{ref:Cheng_2022_CVPR} recently proposed manifold-regularized transition matrix estimation method.
Meanwhile, there are more recent efforts to add a noise adaptation layer or relabel data\,\cite{ref:goldberger_NAL_iclr2017, ref:song_selfie_2019_icml, ref:chen2020beyond}, but they do not perform well especially when numerous classes exist or noise rates are heavy.
%Although Goldberger et al.  \cite{ref:goldberger_NAL_iclr2017} attempted to estimate a noise transition matrix by adding a noise adaptation layer, it is difficult to estimate the noise transition probability when numerous classes exist or noise rates are heavy.
%
The main difference from existing works is that, while they follow a paradigm of `learning from scratch' to prevent overfitting and regard small-loss examples as clean examples, we rather leverage the overfitting property under the `post-training' paradigm, determining the most abnormal influential examples to rule out.
Although RoG\,\cite{ref:rog_2019_icml} builds a simple robust generative classifier on top of pre-trained models as post-processing, it has limited performance gain since it does not make any change to the target model and makes a strong assumption on the distribution of feature representation.
%As a post-processing method, RoG\,\cite{ref:rog_2019_icml} builds a simple robust generative classifier on top of pre-trained models as post-processing, %and NPC~\cite{ref:Bae_2022_ICML} proposed noisy prediction calibration, which estimates the transition from a noisy prediction to a correct prediction via generative process.
%While they improve the pre-trained classifier performance, they have limited performance gain since they do not make explicit changes to the target model. 
%Also, RoG makes a strong assumption on the distribution of feature representation.
\looseness=-1

% Some works use a small number of clean-label samples to detect corrupted data and train a robust model \cite{ref:zhang_wright_aaai2018, ref:sprl_zhang_aaai2020}. 
% The purpose of using clean samples in this approach is fundamentally different from that of our method.
% In the existing approach, the clean samples are used directly for training. But in our method, the clean samples are used to assess the reliability of training data for the trained model. 

%\vspace*{-0.1cm}
%\subsection{Influence Function}

\begin{comment}
\begin{figure}[t]
\centering
\subfigure[Trained model]{\includegraphics[width=0.45\columnwidth]{fig/1.png}}
\subfigure[Influence on model]{\includegraphics[width=0.45\columnwidth]{fig/2.png}} 
\caption{ \textbf{Intuition.}
The red and blue points belong to different classes in binary classification.
The $\times$ marks indicate noisy-label data.
(a) Due to the noisy labels({$\times$}), the model is overfitted. 
(b) {$\times$} significantly affects the model because if the point is removed, the parameter of the model is substantially changed.
}
\label{fig:algo_intuition}
\end{figure}
\end{comment}

\vspace*{-0.1cm}
\subsection{Influence Function}
\vspace*{-0.1cm}
Finding influential examples in a dataset has been studied for decades in robust statistics  \cite{ref:hampel1986robust, cook_influence}.
Recently, a few attempts have been made to apply the idea to neural networks\,\cite{ref:FANN_2003, ref:Koh_Liang_2017}.
%
%The influence function has the potential to be used as a good measure of robustness, but it has not been widely used in DNNs because few suggestions on how to use the influence function have been made.
%
%First, it is difficult to interpret the results of the influence function because the result is a  $k$-dimensional vector, which is the number of model parameters.
Recently, \cite{ref:Koh_Liang_2017} used influence functions to understand the effect of a training example on a test example.
%However, it is difficult to find the pattern on the whole dataset or the model from the individual influence of each training example on a test example.
Although Koh and Liang~\cite{ref:Koh_Liang_2017} showed the possibility of finding mislabels by using the influence function from email spam classification, their method requires human intervention to check and fix the examples, which is not practical. On the other hand, we propose a new criterion, overfitting scores, and a novel post-training algorithm that does not require human intervention. In addition, we extend the method to a new multimedia data, \emph{i.e.}, a video set, and we newly discover that the proposed method can have a regularization effect on clean data.
Meanwhile, \cite{ref:hara_sgdinf_nips2019} proposed stochastic gradient descent\,(SGD) influence that can infer the influential examples for models trained with SGD.
However, this method is limited to optimization by SGD and requires to store the parameters of the model at every step, requiring huge memory consumption for DNNs.
%However, their usage is limited to examining the impact of each example on others and also requires extremely high computation due to their iterative Hessian calculations, which are not feasible in training DNNs.
%
%
%To the best of our knowledge, this is the first work to study the potential impact of the influence function for overcoming noisy labels. 
%On the other hand, our post-training scheme gets its benefit without huge computation and space costs since only a few Hessian computations are required just before the post-training phase, rather than learning from scratch. % in an iterative manner. 

%However, this method is limited to optimization by SGD and must store the parameters of the model at every step, which requires substantial memory for DNNs.
%In addition, how many influential points to remove is not well defined (i.e., the top-\textit{m} influential examples are removed).
%We propose a novel method to use the influence function to identify overfitting and solve LNL problem. 

\vspace*{-0.1cm}
\section{Influential Rank}
\vspace*{-0.1cm}
Our idea is to leverage the property of an overfitted model for post-processing.
First, we present the observations that motivated our method in Section \ref{sec:inf_intuition}.
%, and review the influence function in Section~\ref{sec:preliminary}.
Then, we propose two novel criteria in Section \ref{sec:overfitting_score}, and we describe the overall scheme of robust post-training with overfitting scores, referred to as \our{}\,(Section \ref{sec:algorithm}). 
Finally, we empirically verify the effectiveness of the proposed criteria in post-training from a toy example\,(Section~\ref{sec:toyexample}).
% In the presence of noisy labels, mislabeled examples are critical to the degradation of the classification performance. It is attributed to \emph{overfitting} due to the extremely high capacity of DNNs to fit even noisy labels. Hence, the mislabeled examples exert great influence on the overfitted model. From this intuition, we propose to use the influence of the example to recover the overfitted decision boundary to be correct via post-training.
%
%We first introduce the influence score of a training example in the perspective of example's overfitting\,(Section \ref{sec:overfitting_model}). Here, 

\subsection{Intuition}\label{sec:inf_intuition}
%In the presence of noisy labels, mislabeled examples are critical to the degradation of the classification performance. 
%It is attributed to \emph{overfitting} due to the extremely high capacity of DNNs to fit even noisy labels. 
%\dw{\sout{It is known that DNNs have a high capacity to memorize data, and}}
%It is known that DNNs learn ``easy'' data first and can memorize ``hard'' instances as training processes \cite{ref:arpit_memoriz_icml2017}.
%Thus, in noisy-label circumstances, DNNs are likely to learn the pattern of clean labels first and eventually fit the noisy labels. 
% \dw{\sout{DNNs can learn feature representations of the data well.}}
%Moreover, common features over broad classes are learned in the lower layers, and the class-specific features are learned in the higher layers \cite{ref:zeiler_deconv_eccv2014}. 
%Then, to fit the mislabeled data against the feature representations learned from the clean data, overfitting would likely occur more frequently in the higher layers and the final fully connected layer, which largely contribute to the decision boundary. 
%Hence, the noisy-label data are critical to the degradation of the classification performance. 
%In this perspective, our objective is to find and remove the mislabeled samples causing the overfitted decision boundary.
%noisy labels 
%in the overfitted region around the decision boundary.
%To this end, we need to answer a question: How can we find the overfitted region and determine the noisy labels?

Our post-training algorithm is based on two following intuitions. 
Mislabeled samples are likely to significantly distort the decision boundary, and to cause misclassification of nearby correctly labeled samples.
%\dw{제안하는 finding 알고리즘은 두 가지 intuition에 근거한다. overfitting을 유도하는 mislabeled sample들은 (1) decision boundary의 큰 왜곡을 발생시킨다. (2) 주변의 correctly labeled sample들의 오분류를 야기한다.}
Figure \ref{fig:intuition} illustrates our intuition.
The red and blue points belong to different classes for binary classification, and the pink and light blue background indicates the ground-truth feature embedding space.
% The decision boundary is distorted by overfitting the noisy-label data, as shown by the black line. 
% In Figure \ref{fig:intuition}(a), the model (black line) overfits the noisy labels ($\times$ mark).
Black line denotes a decision boundary predicted by the model.
In Figure \ref{fig:intuition}(a) the model overfits the mislabeld samples ({\color{red} $\mathbb{\times}$} mark), thus the decision boundary is distorted compared to the ground-truth boundary.
When the mislabeled sample ({\color{red} $\mathbb{\times}$} mark) is removed, the trained model is substantially changed (Figure \ref{fig:intuition}(b)). 
% That is, the noisy label can exert great influence on the overfitted model.
That is, the noisy label can exert great influence on the decision boundary of the model. %This influence is referred to as ``influence on model''.

In addition,  to evaluate whether a training sample causes a significantly overfitted classifier, we can use a small number of clean validation data.
We consider a few validation data points\footnote{We use only 5 data per class.} (${\color{blue}\bigstar}$ marks) as shown in Figure  \ref{fig:intuition}(c).
Because the fitted decision boundary is distorted toward the blue region to include the noisy label ({\color{red} $\mathbb{\times}$} mark), the ${\color{blue}\bigstar}$ enclosed by a red dotted circle is wrongly classified into the red class.
Thus, the noisy label ({\color{red} $\mathbb{\times}$} mark) causes a clean sample to be misclassified (i.e., negative influence).
%negatively affects the  ${\bigstar}$ marks enclosed by the red-dotted circle in Figure \ref{fig:intuition}(c)). 
Meanwhile, the validation samples upper the line (blue-dotted circle) are correctly classified that it can be said that the boundary created by this mislabeled sample ({\color{red} $\mathbb{\times}$}) has a positive influence on properly classifying other samples.
%The sample {$\times$} positively influences the validation samples upper the line (blue-dotted circle).
Therefore, the noisy label is likely to have inconsistent influences on the clean validation samples, although their distances are near each other. %This influence is referred to as ``influence on data''
The same claim can apply to the validation samples ({\color{red}$\bigstar$}) in the other ({\color{red} red}) category in Figure \ref{fig:intuition}(d).
We verify the inconsistent influences of noisy labels in Section~\ref{subsec:validity_od}.
% \dw{In addition, we can use a small number of clean validation data to evaluate whether a training sample is mislabeled. 
% We consider a few validation data points ($\bigstar$ marks) as shown in Figure  \ref{fig:intuition}(c). One validation data (${\bigstar}$ enclosed by a red dotted circle) is wrongly classified while the other (${\bigstar}$ enclosed by a blue dotted circle) is correctly classified.
% 반면에, 이러이러한 샘플들은 주변의 validation data들이 consistently and correctly 분류된다. (그림 추가 필요)
% 따라서 noisy sample들은 주변의 validation data들에 대해서 classifier가 분류를 잘 못하게 만든다.
% We referred to it as ``influence on data''.}

From this observation, we present two novel criteria that measure the abnormal influences of a training sample. 
One is to measure how much a training sample affects the overfitting of model parameters, referred to as the {\it overfitting score on model}, and the other measures how inconsistently a training sample affects the classification of clean validation data, which is referred to as the {\it overfitting score on data}.

\subsection{Overfitting Scores}\label{sec:overfitting_score}

To identify overfitting on individual points for detecting noisy labels, we utilize two influence functions in 
~\cite{ref:Koh_Liang_2017}. One is to measure the influence of an example $(x, y)$ on the model $f(x, \hat{\theta})$ trained on the dataset $\mathcal{D}$ via loss function $\ell(y, f(x, \theta))$, given by \begin{equation}
{\mathcal I_M}(x;\hat{\theta}) = -H_{\hat{\theta}}^{-1}
\left.\triangledown_\theta \ell({y}, f(x, \theta))\right|_{\theta=\hat{\theta}},
\label{eq:Iparams}
\end{equation}
where $H_{\hat{\theta}} \eqdef \frac{1}{|{\mathcal{D}}|}\sum_{(x, {y}) \in {\mathcal{D}}} \left.\triangledown_\theta^2\ell\big({y}, f(x, \theta)\big)\right|_{\theta=\hat{\theta}}$. The other is to measure the influence of a training sample $(x_i, y_i)$ on a test sample $(x_t, y_t)$, given by 
\begin{equation}
{\mathcal I_D}(x_i, x_t;\hat{\theta}) = \triangledown_\theta \ell({y_t}, f(x_t, \hat{\theta}))^{\top}{\mathcal I_M}(x_i;\hat{\theta}).
\label{eq:Itest}
\vspace{-3mm}
\end{equation}

\subsubsection{Overfitting Score on Model }
${\cal I}_M(x;\hat{\theta})$ can be used to estimate the effect of a noisy label on an overfitted model (Figure \ref{fig:intuition}(b)).
However, ${\cal I}_M(x;\hat{\theta})$ is a $p$-dimensional vector, where $p$ is the number of model parameters. 
Thus, to measure the strength of the influence of a training point $(x_i, y_i)$, we use $\|{\cal I}_M(x_i;\hat{\theta})\|$ as a metric.
Using this metric, we define 
%\begin{definition}({\scshape Overfitting Score on the Model})
{\it overfitting score on model} (OSM) ${\mathcal O_M}(x_i; \hat{\theta})$ as the model\,(parameter)'s potential change caused by ignoring the example $x_i$ for training, 
\begin{equation}
{\mathcal O_M} (x_i;\hat{\theta}) = \frac{\|{\mathcal I}_M(x_i;\hat{\theta})\|-\mu_{x \in \mathcal{D}}\big(\|{\mathcal I}_M(x;\hat{\theta})\|\big)}{\sigma_{x \in \mathcal{D}} \big(\|{\mathcal I}_M(x;\hat{\theta})\|\big)},
\label{eq:OMscore}
\end{equation}
where $\mu_{x \in \mathcal{D}}(\cdot)$ and $\sigma_{x \in \mathcal{D}}(\cdot)$ denote  mean and standard deviation of $\|{\cal I}_M(x;\hat{\theta})\|$ over $x \in \mathcal{D}$, respectively. 
%\qed
%\label{def:om_score}
%\end{definition}

% \vspace*{-0.05cm}
OSM ${\mathcal O_M} (x_i;\hat{\theta})$ measures a normalized global influence of a training sample $x_i$ on the entire parameters. 
As in Figure~\ref{fig:intuition}(b), the noisy samples are likely to locate near the decision boundary, therefore, they will exhibit a higher OSM than examples with clean labels.

\vspace*{-0.2cm}
\subsubsection{Overfitting Score on Data}\label{sec:overfitting_data}
\vspace*{-0.1cm}
In contrast to a well-generalized decision boundary, an overfitted decision boundary by a mislabeled sample makes the mislabeled sample inconsistently affect clean validation samples, even though the validation samples belong to the same class (Figure \ref{fig:intuition}(c) and \ref{fig:intuition}(d)).
Here, an influence of a training sample on a validation sample indicates how much a classification result of the validation sample changes after removing the training sample. 
Therefore, we suggest overfitting score on data (OSD) as the within-class influence consistency of a training sample $x_i$ on $m$ clean validation samples in $\mathcal{D}_k=\{x_1^v, \cdots x_{m}^v\}$ in the $k$-th  class.
Utilizing (\ref{eq:Itest}), 
%we estimate the within-class influence consistency by measuring the variance of ${\cal I}_D(x_i, x_j^v; \hat{\theta})$ for all $x_j^v$ in the validation set of the $k$th class.
%This function represents the amount of validation error on $(x_j^v, y_j^v)$ after $(x_i, y_i)$ is removed from the training set. 
%To measure the within-class influence consistency of $x_i$ to the $k$th class, we propose to use the variance of ${\cal I}_D(x_i, x_j^v; \hat{\theta})$ for all $x_j^v$ in the validation set of the $k$th class, which is denoted by $\sigma_{k}^2({\cal I}_D(x_i, x^v; \hat{\theta}))$. % \ref{sup:OSD}).
%Then, to estimate samples with relatively high influences, we normalize $\sigma_{k}({\cal I}_D(\cdot; \hat{\theta}))$.
%\begin{definition}({\scshape Overfitting Score on Data}) OSD ${\mathcal O_D^k} (x_i;\hat{\theta})$ 
%as the within-class influence consistency of a training sample $x_i$ on $m$ clean validation samples $x_1^v, \cdots x_{m}^v$ 
OSD ${\mathcal O_D^k} (x_i;\hat{\theta})$ in the $k$-th class is defined by
\begin{equation}
\small
{\mathcal O_D^k} (x_i;\hat{\theta}) = \frac{\sigma_k\big({\cal I}_D\big(x_i, x^v; \hat{\theta}\big)\big) - \mu\big(\sigma_k\big({\cal I}_D\big(x, x^v; \hat{\theta}\big)\big)\big)}{\sigma\big(\sigma_k\big({\cal I}_D\big(x, x^v; \hat{\theta}\big)\big)\big)},
\label{eq:ODscore}
\end{equation}
where $\sigma_k(\cdot)$ is standard deviation of ${\cal I}_D\big(x, x^v; \hat{\theta}\big)$ over $x^v \in \mathcal{D}_k$, whereas $\mu(\cdot)$ and $\sigma(\cdot)$ denote mean and standard deviation of $\sigma_k(\cdot)$ over $k$. 
%\qed
%\label{def:od_score}
%\end{definition}

\vspace*{-0.1cm}
\subsection{Post-processing with Influential Rank} \label{sec:algorithm}
\vspace*{-0.1cm}
Algorithm \ref{alg:proposed_algorithm} outlines the overall procedure of \our{}.
%\dw{post-processing이 크게 두단계 (influential sample 제외, 제외된 sample을 labeling correction 후 포함하여 재학습) 로 이루어 진 것 같은데, subsection이나, bold체로 두 단계를 구분해주는 것은 어떨까요?}
% Given a pre-trained model, \our repeats updating the model after excluding highly influential examples\,(\emph{i.e.} potentially mislabeled examples) for a fixed number of post-training epochs, which is much smaller than the total training epochs of the pre-trained model. 
Given a pre-trained model, \our updates the model parameter with the training dataset excluding highly influential examples\,(\emph{i.e.}, potentially mislabeled examples) for a fixed number of post-training epochs, which is much smaller than the total training epochs of the pre-trained model.
Specifically, given a pre-trained model $\hat{\theta}_{0}$, we calculate ${\mathcal O_M} (x_i;\hat{\theta})$ for the whole training dataset ${\mathcal{D}}$ (Line 3).
Since our goal is to exclude examples that have high scores for both ${\cal O}_M (x_i;\hat{\theta})$ and ${\mathcal O_D^k} (x_i;\hat{\theta})$, we compute ${\mathcal O_D^k} (x_i;\hat{\theta})$ for the training samples whose ${\mathcal O_M} (x_i;\hat{\theta})$ are higher than the mean (\emph{i.e.}, 0) for efficient computation.
%We justify the choice of the mean (\emph{i.e.}, 0) in .

% To automatically quantify the number of influential samples that need to be eliminated, we assume a two-modality Gaussian mixture model\,(GMM).
% We fit the two-modality GMM to ${\mathcal O_D^k} (x_i;\hat{\theta})$ using the Expectation-Maximization algorithm. 
% To select enough candidates, we assume the training samples whose ${\mathcal O_D^k} (x_i;\hat{\theta})$ are higher than the smaller mean of the Gaussian component as {\it noisy candidates}.
% Then, we decide the final influential samples if a noisy candidate is inconsistent for the validation data of more than $\gamma$ classes in common (\emph{i.e.}, more than $\gamma$ classes have \textit{consensus} that the noisy candidate is inconsistent). 

%\dw{윗 문단을 조금 paraphrasing 해봤습니다. 
To automatically quantify the number of influential samples that need to be eliminated, we assume a two-modality Gaussian mixture model\,(GMM). 
First, we fit the two-modality GMM to ${\mathcal O_D^k} (x_i;\hat{\theta})$ using the Expectation-Maximization algorithm. 
Next, we select the training samples whose ${\mathcal O_D^k} (x_i;\hat{\theta})$ are higher than the smaller mean of the Gaussian component (Line 6).
We referred to those samples as {\it noisy candidates}. 
Then, we decide the final influential samples if a noisy candidate is inconsistent for %the validation data of 
more than $\gamma$ classes in common (\emph{i.e.}, more than $\gamma$ classes have \textit{consensus} that the noisy candidate is inconsistent), which are referred as {\it noisy-probable samples} (Line 8).

After removing all the noisy-probable samples, the model is retrained for a small number of epochs using the new training set (Line 9, 10).
If a meaningful improvement in the classification accuracy occurs, the noisy-probable samples are eliminated from the training set, and the algorithm is repeated. 
Otherwise, the noisy-probable samples are not removed, and the algorithm stops.

When the algorithm finishes, new labels of the removed samples are predicted by the classifier in the last iteration.
Simply, we replace the labels of the noisy data with the newly corrected labels.
Then, among the corrected training data, the new clean dataset includes only the data whose softmax outputs are higher than $S$(prediction threshold).
Then, the model is newly trained on the new clean dataset and is evaluated for the test dataset.

This iterative design allows to remove more mislabeled examples in an iterative manner. As the model evolves, \our{} can incrementally find hard-to-identify mislabeled examples that could not be detected in the previous round. 
Especially under the high noise-level circumstances\,\emph{e.g.}, 70\% of label noise, multi-round post-training achieves significant performance gains. %according to our analysis in Section \ref{sec:exp_multi}.

\setlength{\textfloatsep}{10pt}% Remove \textfloatsep
\begin{algorithm}[t!]
\caption{\our{}}
\label{alg:proposed_algorithm}
\begin{algorithmic}[1]
\REQUIRE { ${\mathcal{D}}$: data, $\hat{\theta}_{0}$: pre-trained model, $epochs$: post-training epochs, $\gamma$: consensus number}
\ENSURE { $\theta_r$: model parameter after post-training}
\STATE $\mathcal{C}\leftarrow {\mathcal{D}}$ ~~~\COMMENT{$\mathcal{C}$ is entire clean samples in ${\mathcal{D}}$}
\STATE {\bf repeat} 
\STATE \,\,\,\,${\mathcal{D}_M} \leftarrow \{ x_i |\{{\mathcal O}_M(x_i;\hat{\theta}_{r})\}_{i=1}^{n}\! \geq 0 \} $ // Compute  Eq.\,\eqref{eq:OMscore}\!\!\!\!\!\!\!\!
\STATE \,\,\,\,{\bf for} class $k=1$ {\bf to} $K$ {\bf do}
\STATE \,\,\,\,\,\,\,\,\COMMENT{Compute  Eq.\,\eqref{eq:ODscore} and fit GMM $(G_{low}, G_{high})$}
\STATE \,\,\,\,\,\,\,\ ${\mathcal{D}_D^k} \leftarrow \{ x_i | \{{\mathcal O}_D^k(x_i;\hat{\theta}_{r})\}_{i \in {\mathcal{D}_M}} \geq \mu(G_{low}) \}$
\STATE \,\,\,\,{\bf end for}
\STATE \,\,\,\,$\mathcal{S} \leftarrow \{x_i | \sum_{k=1}^K 1[x_i \in {\mathcal D}_D^k] \geq \gamma \} $ 
\STATE \,\,\,\,$\mathcal{C}\leftarrow \mathcal{C} - \mathcal{S}$ ~~~\COMMENT{Update the clean set}
\STATE \,\,\,\,Post-train $\hat{\theta}_{r}$ on the refined clean set $\mathcal{C}$ for $epochs$
\STATE {\bf until} acc$(\hat{\theta}_{r})$ saturates
\STATE {\bf return} $\hat{\theta}_{r}$
\end{algorithmic}
\end{algorithm}

%%%% TO-do decision boundary 변화하는 그림 추가하기 %%%%%%%%%%%%%%
\begin{figure*}[t]
\centering
    \subfigure[Ours \small{(Clean Model)}]
     {\includegraphics[width=0.24\linewidth]{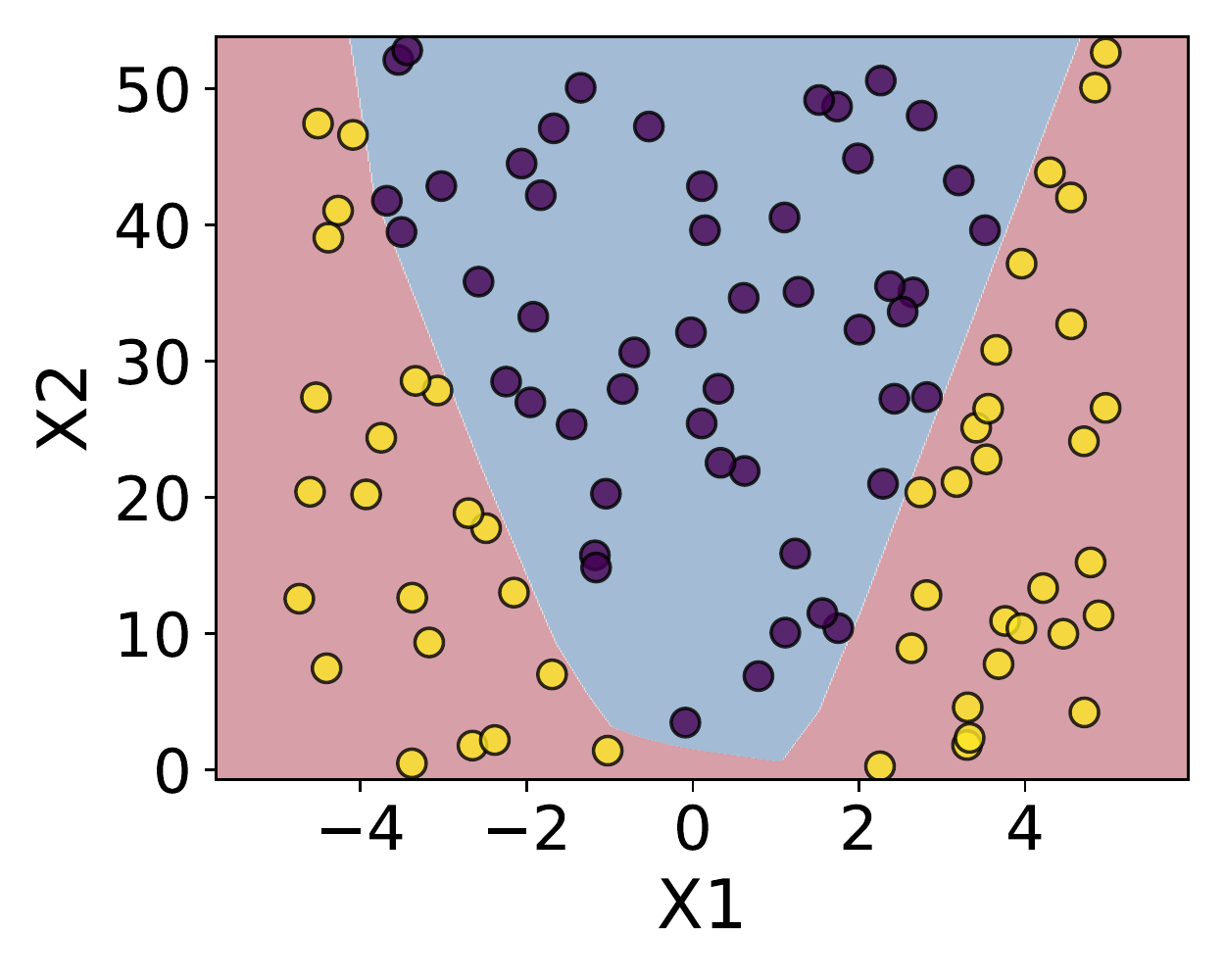}}
     \subfigure[Ours\small{(Corrupted Model)}]
     {\includegraphics[width=0.24\linewidth]{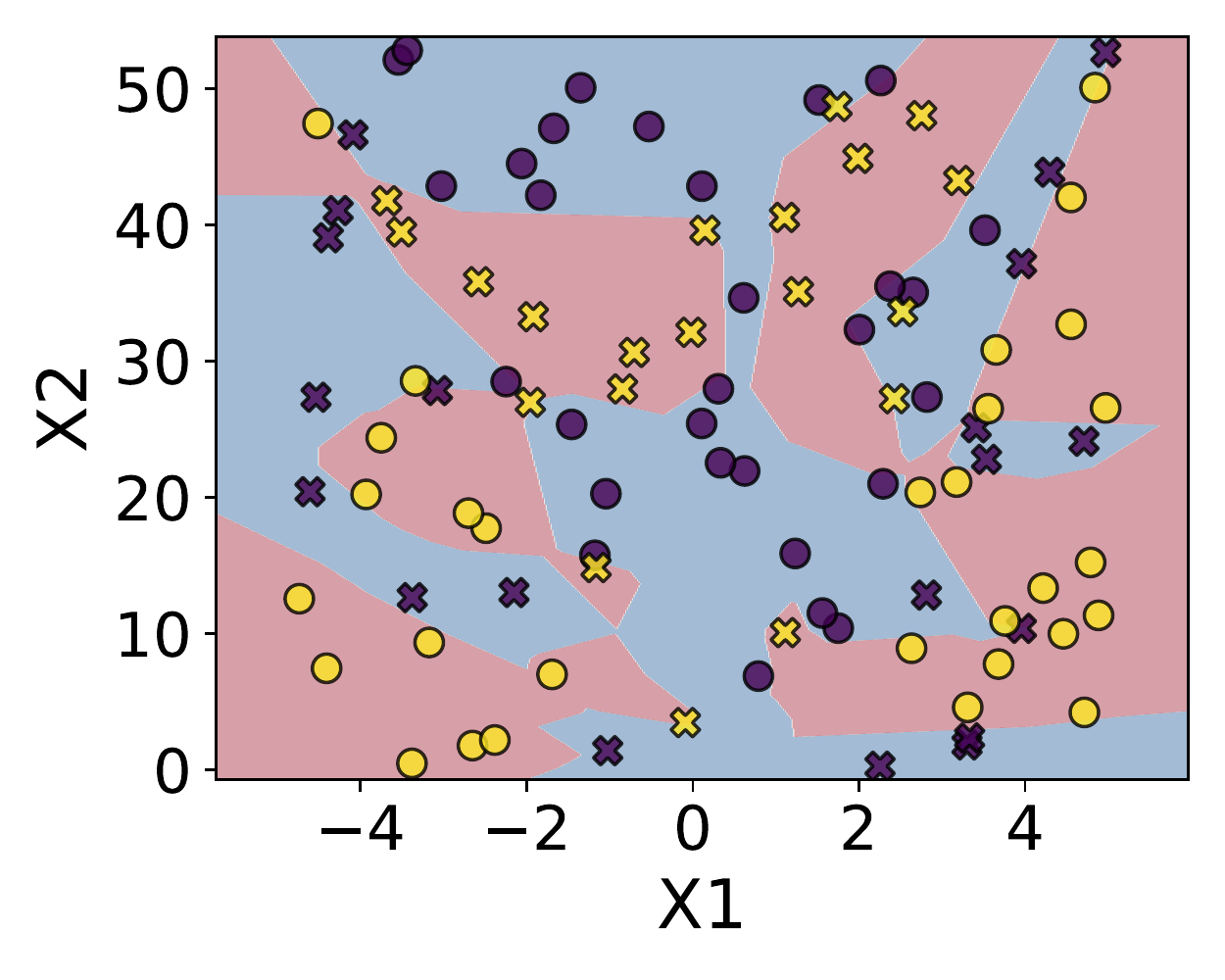}}
     \subfigure[Ours\small{Ours \small{(after 1st iter.)}}]
     {\includegraphics[width=0.24\linewidth]{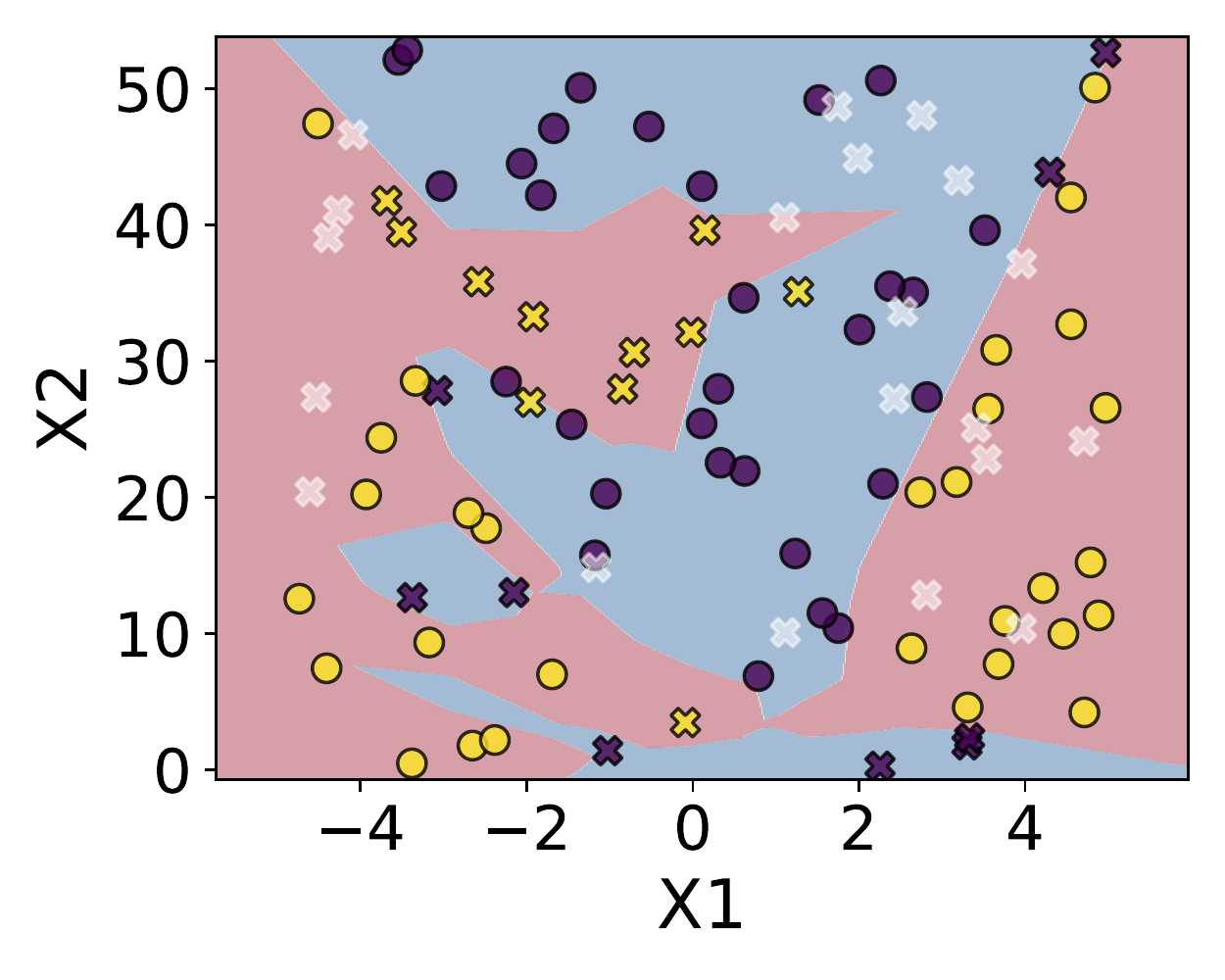}}
    \subfigure[Ours\small{Ours \small{(after 3rd iter.)}}]
     {\includegraphics[width=0.24\linewidth]{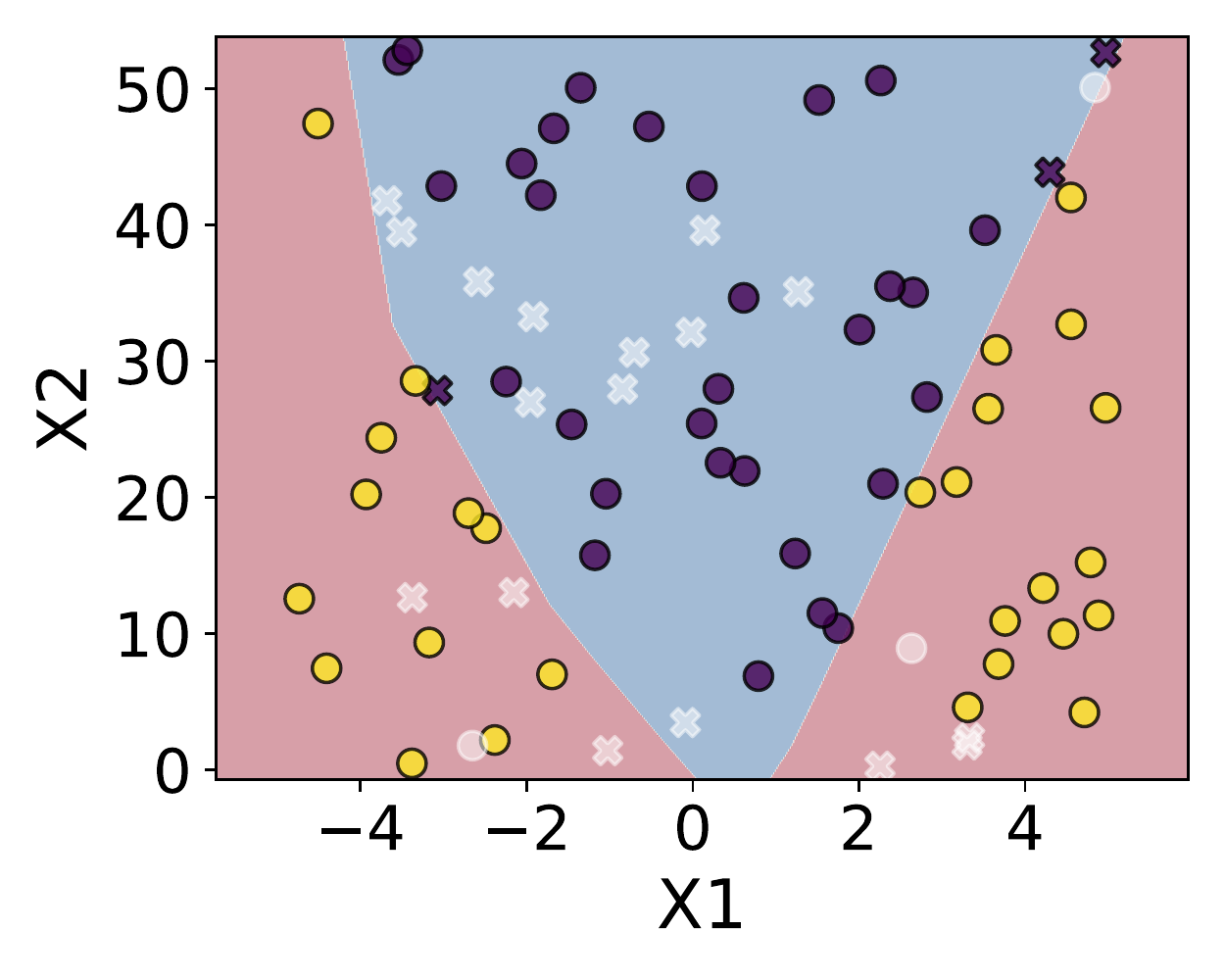}}
\vspace*{-0.1cm}
\caption{ \textbf{Change in decision boundary by Influential Rank.}
(a) DNN trained on clean data.
(b) DNN trained on noisy data (randomly chosen 40\% of labels are flipped).
(c) DNN after first iteration by Ours. 
(d) DNN after third iteration by Ours.
}
\label{fig:toyexample}
\end{figure*}
\vspace*{-0.2cm}

\begin{table*}[h]
\caption{
\textbf{Comparison on CIFAR with varying levels of symmetric label noises.}
The averaged test accuracy\,(\%) with LNL methods and their combination with RoG and \our. 
The mean accuracy is computed over three different noise realizations.
%where the values in parentheses are the relative performance gain (\%) compared with the pre-trained methods. 
%The best results are marked in bold.
}
\vspace{-0.cm}
\label{table:cifar-symm}
\centering
%\small
\footnotesize
\resizebox{1\textwidth}{!}
{
%\begin{tabular}{L{1.1cm} X|{1.1cm} X{1.1cm} X{1.1cm} X|{1.1cm} X{1.1cm} X{1.1cm} X|{1.1cm} X{1.1cm} X{1.1cm} X{1.1cm}}
\begin{tabular}{X{1.0cm}|l|X{1.15cm} X{1.3cm} X{1.2cm}|X{1.15cm} X{1.3cm} X{1.2cm}|X{1.15cm} X{1.3cm} X{1.2cm}}
\toprule
\multirow{2}{*}{\!\!Dataset\!\!}   & \multicolumn{1}{c|}{\multirow{2}{*}{Method}} & \multicolumn{3}{c|}{Symm-20}                                                                                                                                                         & \multicolumn{3}{c|}{Symm-50}                                                                                                                                                          & \multicolumn{3}{c}{Symm-70}                                                                                                                                                           \\ \cline{3-11} 
& \multicolumn{1}{c|}{}  & Original & +ROG \cite{ref:rog_2019_icml} & \!\!+Inf. Rank\!\!                                          & Original & +ROG \cite{ref:rog_2019_icml} & \!\!+Inf. Rank\!\!                                          & Original & +ROG \cite{ref:rog_2019_icml} & \!\!+Inf. Rank\!\!                                                             \\ \midrule 
\multirow{6}{*}{\rotatebox[origin=c]{90}{\makecell[l]{\hspace*{-1.75cm}CIFAR-10}}}  & CE                                           & \begin{tabular}[c]{@{}c@{}}80.46\\ (+0.0)\end{tabular} & \begin{tabular}[c]{@{}c@{}}86.97\\ (+6.51)\end{tabular} & \textbf{\begin{tabular}[c]{@{}c@{}}91.08\\ (+10.62)\end{tabular}} & \begin{tabular}[c]{@{}c@{}}48.84\\ (+0.0)\end{tabular} & \begin{tabular}[c]{@{}c@{}}62.59\\ (+13.76)\end{tabular} & \textbf{\begin{tabular}[c]{@{}c@{}}84.19\\ (+35.36)\end{tabular}} & \begin{tabular}[c]{@{}c@{}}28.42\\ (+0.0)\end{tabular} & \begin{tabular}[c]{@{}c@{}}44.92\\ (+16.50)\end{tabular} & \textbf{\begin{tabular}[c]{@{}c@{}}70.59\\ (+42.17)\end{tabular}} \\ \cline{2-11} 
                           & VolMinNet \cite{ref:volminnet_2021_icml}                                  & \begin{tabular}[c]{@{}c@{}}88.26\\ (+0.0)\end{tabular} & \begin{tabular}[c]{@{}c@{}}88.49\\ (+0.23)\end{tabular} & \textbf{\begin{tabular}[c]{@{}c@{}}91.89\\ (+3.63)\end{tabular}} & \begin{tabular}[c]{@{}c@{}}71.13\\ (+0.0)\end{tabular} & \begin{tabular}[c]{@{}c@{}}72.65\\ (+1.52)\end{tabular} & \textbf{\begin{tabular}[c]{@{}c@{}}83.63\\ (+12.50)\end{tabular}}  & \begin{tabular}[c]{@{}c@{}}33.69\\ (+0.0)\end{tabular} & \begin{tabular}[c]{@{}c@{}}42.08\\ (+8.40)\end{tabular} & \textbf{\begin{tabular}[c]{@{}c@{}}66.07\\ (+32.39)\end{tabular}}  \\ \cline{2-11} 
                           & Co-teaching \cite{ref:coteacing_nips_18}                                          & \begin{tabular}[c]{@{}c@{}}91.85\\ (+0.0)\end{tabular} & \begin{tabular}[c]{@{}c@{}}90.22\\ (-1.62)\end{tabular} & \textbf{\begin{tabular}[c]{@{}c@{}}93.10\\ (+1.25)\end{tabular}} & \begin{tabular}[c]{@{}c@{}}85.44\\ (+0.0)\end{tabular} & \begin{tabular}[c]{@{}c@{}}81.96\\ (-3.48)\end{tabular} & \textbf{\begin{tabular}[c]{@{}c@{}}87.30\\ (+1.86)\end{tabular}}  & \begin{tabular}[c]{@{}c@{}}52.63\\ (+0.0)\end{tabular} & \begin{tabular}[c]{@{}c@{}}53.93\\ (+1.30)\end{tabular} & \textbf{\begin{tabular}[c]{@{}c@{}}60.95\\ (+8.33)\end{tabular}}  \\  \cline{2-11} 
                           & ELR \cite{ref:elr_neurips_2020}                                          & \begin{tabular}[c]{@{}c@{}}91.88\\ (+0.0)\end{tabular} & \begin{tabular}[c]{@{}c@{}}91.50\\ (-0.39)\end{tabular} & \textbf{\begin{tabular}[c]{@{}c@{}}93.04\\ (+1.15)\end{tabular}} & \begin{tabular}[c]{@{}c@{}}88.48\\ (+0.0)\end{tabular} & \begin{tabular}[c]{@{}c@{}}87.62\\ (-0.86)\end{tabular} & \textbf{\begin{tabular}[c]{@{}c@{}}89.60\\ (+1.12)\end{tabular}}  & \begin{tabular}[c]{@{}c@{}}77.26\\ (+0.0)\end{tabular} & \begin{tabular}[c]{@{}c@{}}72.90\\ (-4.36)\end{tabular} & \textbf{\begin{tabular}[c]{@{}c@{}}80.13\\ (+2.86)\end{tabular}}  \\  \cline{2-11}
                           & ELR+ \cite{ref:elr_neurips_2020}                                   & \begin{tabular}[c]{@{}c@{}}93.75\\ (+0.0)\end{tabular} & \begin{tabular}[c]{@{}c@{}}93.00\\ (-0.75)\end{tabular} & \textbf{\begin{tabular}[c]{@{}c@{}}94.73\\ (+0.98)\end{tabular}} & \begin{tabular}[c]{@{}c@{}}92.05\\ (+0.0)\end{tabular} & \begin{tabular}[c]{@{}c@{}}91.11\\ (-0.94)\end{tabular} & \textbf{\begin{tabular}[c]{@{}c@{}}92.79\\ (+0.74)\end{tabular}}  & \begin{tabular}[c]{@{}c@{}}86.94\\ (+0.0)\end{tabular} & \begin{tabular}[c]{@{}c@{}}83.73\\ (-3.21)\end{tabular} & \textbf{\begin{tabular}[c]{@{}c@{}}88.21\\ (+1.27)\end{tabular}} \\ 
                           \cline{2-11}
                           & DivideMix \cite{ref:dividemix_iclr_2019}                                    & \begin{tabular}[c]{@{}c@{}}95.64\\ (+0.0)\end{tabular} & \begin{tabular}[c]{@{}c@{}}95.08\\ (-0.56)\end{tabular} & \textbf{\begin{tabular}[c]{@{}c@{}}96.13\\ (+0.49)\end{tabular}} & \begin{tabular}[c]{@{}c@{}}94.02\\ (+0.0)\end{tabular} & \begin{tabular}[c]{@{}c@{}}93.50\\ (-0.53)\end{tabular} & \textbf{\begin{tabular}[c]{@{}c@{}}94.83\\ (+0.80)\end{tabular}}  & \begin{tabular}[c]{@{}c@{}}91.27\\ (+0.0)\end{tabular} & \begin{tabular}[c]{@{}c@{}}88.69\\ (-2.58)\end{tabular} & \textbf{\begin{tabular}[c]{@{}c@{}}92.42\\ (+1.14)\end{tabular}} \\ 
                           \cline{2-11}
                           & UNICON \cite{ref:unicon_2022}                                    & \begin{tabular}[c]{@{}c@{}}91.95\\ (+0.0)\end{tabular} & \begin{tabular}[c]{@{}c@{}}91.27\\ (-0.68)\end{tabular} & \textbf{\begin{tabular}[c]{@{}c@{}}94.98\\ (+3.02)\end{tabular}} & \begin{tabular}[c]{@{}c@{}}93.59\\ (+0.0)\end{tabular} & \begin{tabular}[c]{@{}c@{}}92.38\\ (-1.22)\end{tabular} & \textbf{\begin{tabular}[c]{@{}c@{}}95.05\\ (+0.09)\end{tabular}}  & \begin{tabular}[c]{@{}c@{}}91.44\\ (+0.0)\end{tabular} & \begin{tabular}[c]{@{}c@{}}89.38\\ (-2.06)\end{tabular} & \textbf{\begin{tabular}[c]{@{}c@{}}93.12\\ (+1.68)\end{tabular}}  \\\hline \hline
\multirow{6}{*}{\rotatebox[origin=c]{90}{\makecell[l]{\hspace*{-1.75cm}CIFAR-100}}} 
    & CE    
    & \begin{tabular}[c]{@{}c@{}}64.35\\ (+0.0)\end{tabular} & \begin{tabular}[c]{@{}c@{}}68.21\\ (+3.86)\end{tabular} & \textbf{\begin{tabular}[c]{@{}c@{}}70.14\\ (+5.79)\end{tabular}} & \begin{tabular}[c]{@{}c@{}}39.43\\ (+0.0)\end{tabular} & \begin{tabular}[c]{@{}c@{}}56.94\\ (+17.51)\end{tabular} & \textbf{\begin{tabular}[c]{@{}c@{}}59.31\\ (+19.88)\end{tabular}} & \begin{tabular}[c]{@{}c@{}}15.50\\ (+0.0)\end{tabular} & \begin{tabular}[c]{@{}c@{}}39.03\\ (+23.53)\end{tabular} & \textbf{\begin{tabular}[c]{@{}c@{}}40.42\\ (+24.91)\end{tabular}} \\ 
    \cline{2-11}
   & VolMinNet \cite{ref:volminnet_2021_icml}                                 & \begin{tabular}[c]{@{}c@{}}65.11\\ (+0.0)\end{tabular} & \begin{tabular}[c]{@{}c@{}}64.93\\ (-0.18)\end{tabular} & \textbf{\begin{tabular}[c]{@{}c@{}}70.05\\ (+4.94)\end{tabular}} & \begin{tabular}[c]{@{}c@{}}48.77\\ (+0.0)\end{tabular} & \begin{tabular}[c]{@{}c@{}}53.91\\ (+5.14)\end{tabular} & \textbf{\begin{tabular}[c]{@{}c@{}}58.41\\ (+9.64)\end{tabular}}  & \begin{tabular}[c]{@{}c@{}}28.64\\ (+0.0)\end{tabular} & \begin{tabular}[c]{@{}c@{}}37.02\\ (+8.38)\end{tabular} & \textbf{\begin{tabular}[c]{@{}c@{}}40.48\\ (+11.84)\end{tabular}}  \\ \cline{2-11} 
   & Co-teaching \cite{ref:coteacing_nips_18}                                          & \begin{tabular}[c]{@{}c@{}}70.85\\ (+0.0)\end{tabular} & \begin{tabular}[c]{@{}c@{}}66.93\\ (-3.93)\end{tabular} & \textbf{\begin{tabular}[c]{@{}c@{}}72.73\\ (+1.87)\end{tabular}} & \begin{tabular}[c]{@{}c@{}}59.14\\ (+0.0)\end{tabular} & \begin{tabular}[c]{@{}c@{}}56.42\\ (-2.72)\end{tabular} & \textbf{\begin{tabular}[c]{@{}c@{}}61.29\\ (+2.16)\end{tabular}}  & \begin{tabular}[c]{@{}c@{}}35.40\\ (+0.0)\end{tabular} & \begin{tabular}[c]{@{}c@{}}35.97\\(+0.57)\end{tabular} & \textbf{\begin{tabular}[c]{@{}c@{}}38.29\\ (+2.89)\end{tabular}}  \\  \cline{2-11}
   & ELR \cite{ref:elr_neurips_2020}                                          & \begin{tabular}[c]{@{}c@{}}72.58\\ (+0.0)\end{tabular} & \begin{tabular}[c]{@{}c@{}}70.14\\ (-2.44)\end{tabular} & \textbf{\begin{tabular}[c]{@{}c@{}}74.23\\ (+1.66)\end{tabular}} & \begin{tabular}[c]{@{}c@{}}64.01\\ (+0.0)\end{tabular} & \begin{tabular}[c]{@{}c@{}}62.91\\ (-1.10)\end{tabular} & \textbf{\begin{tabular}[c]{@{}c@{}}64.43\\ (+0.42)\end{tabular}}  & \begin{tabular}[c]{@{}c@{}}38.78\\ (+0.0)\end{tabular} & \textbf{\begin{tabular}[c]{@{}c@{}}42.07\\ (+3.29)\end{tabular}} & \begin{tabular}[c]{@{}c@{}}40.07\\ (+1.29)\end{tabular}  \\  \cline{2-11}
   & ELR+ \cite{ref:elr_neurips_2020}                                   & \begin{tabular}[c]{@{}c@{}}74.15\\ (+0.0)\end{tabular} & \begin{tabular}[c]{@{}c@{}}70.29\\ (-3.86)\end{tabular} & \textbf{\begin{tabular}[c]{@{}c@{}}75.45\\ (+1.30)\end{tabular}} & \begin{tabular}[c]{@{}c@{}}65.66\\ (+0.0)\end{tabular} & \begin{tabular}[c]{@{}c@{}}65.65\\ (-0.01)\end{tabular} & \textbf{\begin{tabular}[c]{@{}c@{}}68.74\\ (+3.08)\end{tabular}}  & \begin{tabular}[c]{@{}c@{}}50.19\\ (+0.0)\end{tabular} & \begin{tabular}[c]{@{}c@{}}54.48\\ (+4.29)\end{tabular} & \textbf{\begin{tabular}[c]{@{}c@{}}56.53\\ (+6.34)\end{tabular}}  \\
   \cline{2-11}
   & DivideMix \cite{ref:dividemix_iclr_2019}                                    & \begin{tabular}[c]{@{}c@{}}76.57\\ (+0.0)\end{tabular} & \begin{tabular}[c]{@{}c@{}}72.29\\ (-4.28)\end{tabular} & \textbf{\begin{tabular}[c]{@{}c@{}}78.63\\ (+2.06)\end{tabular}} & \begin{tabular}[c]{@{}c@{}}72.29\\ (+0.0)\end{tabular} & \begin{tabular}[c]{@{}c@{}}68.88\\ (-3.41)\end{tabular} & \textbf{\begin{tabular}[c]{@{}c@{}}74.39\\ (+2.10)\end{tabular}}  & \begin{tabular}[c]{@{}c@{}}62.43\\ (+0.0)\end{tabular} & \begin{tabular}[c]{@{}c@{}}58.73\\ (-3.69)\end{tabular} & \textbf{\begin{tabular}[c]{@{}c@{}}65.41\\ (+2.98)\end{tabular}}  \\
   \cline{2-11} 
   & UNICON \cite{ref:unicon_2022}                                    & \begin{tabular}[c]{@{}c@{}}74.82\\ (+0.0)\end{tabular} & \begin{tabular}[c]{@{}c@{}}69.84\\ (-4.98)\end{tabular} & \textbf{\begin{tabular}[c]{@{}c@{}}79.61\\ (+4.79)\end{tabular}} & \begin{tabular}[c]{@{}c@{}}73.96\\ (+0.0)\end{tabular} & \begin{tabular}[c]{@{}c@{}}68.64\\ (-5.32)\end{tabular} & \textbf{\begin{tabular}[c]{@{}c@{}}75.70\\ (+1.74)\end{tabular}}  & \begin{tabular}[c]{@{}c@{}}68.61\\ (+0.0)\end{tabular} & \begin{tabular}[c]{@{}c@{}}63.22\\ (-5.39)\end{tabular} & \textbf{\begin{tabular}[c]{@{}c@{}}69.51\\ (+0.90)\end{tabular}}  \\\bottomrule
\end{tabular}
}
\vspace{-0.3cm}
\end{table*}

\vspace*{-0.1cm}
\subsection{Example: A Binary Classification} \label{sec:toyexample}

\vspace*{-0.1cm}
We present a toy example to verify and visualize our hypothesis and the efficacy of the proposed overfitting scores. 
%in Eq.\,\eqref{eq:OMscore} and \eqref{eq:ODscore}. 
Figure \ref{fig:toyexample} illustrates the toy example. For the two-dimensional binary classification problem, we first generate $100$ data points from the uniform distribution, where $x_1 \sim {\rm Unif}(-5, 5)$ and $x_2 \sim {\rm Unif}(0, 55)$, and their true labels $y$ are assigned following the binary rule depending on their $(x_1, x_2)$ values,
$y=1 \mbox{ if } ~x_2~ \geq ~3x_1^2$ and  $y= 0 \mbox{ if } ~{\rm otherwise.}$
As can be seen  in Figure \ref{fig:toyexample} (c) after the first iteration excluding 20 examples with  high overfitting scores and (d) after the third iteration excluding 20 more examples,  the post-trained decision boundary changes into that similar to the clean model. Therefore, this toy example illustrates the validity of \our for robust post-training. The details are presented in Appendix. % \ref{appendix:toyexample}.

%% To-do multi-round 실험 추가.
%% detection에 결과/ 얼마나, 정확하게 지웠는지?(precision,recall)
% dividemix selection한 것 빼고 추가한 실험 추가!

\vspace{-1.5mm}
\section{Experiments}
\label{sec:experiment}

%We present various experiments and analyses of \our in this section. 
%We describe our experimental settings. Then, we validate the superiority of \our against label noise, followed by ample in-depth comparison and ablation studies. Last, we show its potential as a detector for data cleaning and a universal regularizer.

%We first describe our experimental settings with implementation details in Section \ref{sub:exp_setting}.
% Next, we show the effectiveness of \our on three long-tailed classification benchmarks: CIFAR-100-LT, ImageNet-LT, and iNaturalist, where \our consistently boosts the performance of baselines with state-of-the-art level accuracy (Section~\ref{sub:exp:main_exp}). 
% We also present in-depth analyses of \our to study its inherent characteristics in Section~\ref{sub:exp_analysis}.
%\vspace*{0.15cm}
\subsection{Experimental Settings} \label{sub:exp_setting}
\noindent\textbf{Datasets.}
We conduct classification on multiple benchmark datasets, including synthetic noisy labels and real-world noisy labels as in Table~\ref{table:data}.
A detail of datasets is presented in Appendix. %~\ref{appendix:subsec:dataset}.

\begin{table}[h]
\centering\caption{\textbf{Summary of datasets.}}
%\vspace{-1mm}
\footnotesize
\label{table:data}
\begin{tabular}{l|c|c|c}
\hline
Dataset       & \# of training & noise ratio ($\varepsilon$) & noise type \\ \hline
CIFAR~\cite{ref:data_cifar}         & 50K            & 20, 50, 70  & synthetic  \\ \hline
CIFAR-N~\cite{ref:cifar-n_2021}       & 40K            & 9, 18, 40   & real-world \\ \hline
WebVision 1.0~\cite{ref:li2017webvision} & 2.4M           & 20          & real-world \\ \hline
Clothing1M~\cite{ref:clothing1m_cvpr_15}    & 1M             & 38          & real-world \\ \hline
\end{tabular}
\end{table}

\smallskip\smallskip
\noindent
\textbf{Compared methods.} 
We compare \our with the main baseline, RoG \cite{ref:rog_2019_icml}, which is the post-processing method using a robust generative classifier.
%Due to the merit of post-training, \our and RoG can be applied to any %pre-trained models using existing robust methods.
%\our{} and RoG both use 500 validation samples.
We combine the two post-training methods with a default method\,(CE)
%\footnote[2]{CE uses cross-entropy loss without any processing for label noise.}
and five state-of-the-art robust methods from different directions, \textit{i.e.}, a sample selection method Co-teaching\,\cite{ref:coteacing_nips_18}, a robust regularization method ELR\,\& ELR+\cite{ref:elr_neurips_2020}, a loss correction method VolMinNet\,\cite{ref:volminnet_2021_icml}, a semi-supervised learning (SSL) method DivideMix\,\cite{ref:dividemix_iclr_2019}, and a SSL \& contrastive learning method, UNICON~\cite{ref:unicon_2022} .
%
% Following the literature\,\cite{ref:elr_neurips_2020}, all the compared methods are trained using ResNet-34, Inception-ResNet V2, and ResNet-50 for CIFAR-10/100(N), WebVision datasets, and Clothing1M respectively. 
% We use SGD with a momentum of $0.9$ and dacay the initial learning rate twice during training.
% The networks are trained for 120 epochs for CIFAR-10(N), and 150 epochs for CIFAR-100(N).
%In our main experiment, the number of post-training epochs is set to be $20$, which is much smaller than the total training period of more than $100$ epochs.
The details of the compared models and experimental settings are presented in Appendix.% \ref{appendix:sec:implementation}. \looseness=-1 

\begin{table*}[t!]
\caption{
\textbf{Comparison on CIFAR-N with varying levels of real-world label noise.}
The averaged test accuracy\,(\%) with LNL methods and their combination with RoG and \our{}. %, where the values in parentheses are the performance improvement compared with CE. The best results are marked in bold.
%Relative performance gain values are shown in parentheses.
The mean accuracy is computed over three different noise realizations.
}
\vspace{-0.cm}
\label{table:cifar-n}
\centering
\footnotesize
%\small
\resizebox{1\textwidth}{!}
{
%\begin{tabular}{l|X{0.65cm} X{1.0cm} X{1.15cm} X{0.65cm} X{1.0cm} X{1.15cm} X{0.65cm} X{1.0cm} X{1.2cm} |X{0.65cm} X{1.0cm} X{1.15cm}}
\begin{tabular}{l|X{0.85cm} X{1.15cm} X{1.15cm} X{0.85cm} X{1.15cm} X{1.15cm} X{0.85cm} X{1.15cm} X{1.2cm} |X{0.85cm} X{1.15cm} X{1.15cm}}
\toprule
& \multicolumn{9}{c|}{CIFAR-10N}  & \multicolumn{3}{c}{CIFAR-100N} \\ \hline
\multirow{2}{*}{Method} & \multicolumn{3}{c|}{Aggregate ($\varepsilon \approx 9\%$)}                     & \multicolumn{3}{c|}{Random1 ($\varepsilon \approx 18\%$)}                                              & \multicolumn{3}{c|}{Worst ($\varepsilon \approx 40\%$)}                                                & \multicolumn{3}{c}{Noisy ($\varepsilon \approx 40\%$)}                                                 \\ \cline{2-13} 
&   & \!\!\!\!+ROG \cite{ref:rog_2019_icml}\!\!\!\! & \multicolumn{1}{c|}{\!{+Inf. Rank}\!} 
&   & \!\!\!\!+ROG \cite{ref:rog_2019_icml}\!\!\!\! & \multicolumn{1}{c|}{\!\!\!{+Inf. Rank}\!\!}
&   & \!\!\!\!+ROG \cite{ref:rog_2019_icml}\!\!\!\! & \!\!\!{+Inf. Rank}\!\!\!
&   & \!\!\!\!+ROG \cite{ref:rog_2019_icml}\!\!\!\! & \!\!\!{+Inf. Rank}\!\!\! \\ \midrule
CE      & \begin{tabular}[c]{@{}c@{}}89.81\\ (+0.0)\end{tabular} 
        & \begin{tabular}[c]{@{}c@{}}90.19\\ (+0.38)\end{tabular}  
        & \multicolumn{1}{c|}{\textbf{\begin{tabular}[c]{@{}c@{}}91.85\\(+2.05)\end{tabular}}} 
        & \begin{tabular}[c]{@{}c@{}}83.80\\ (+0.0)\end{tabular} 
        & \begin{tabular}[c]{@{}c@{}}85.10\\ (+1.30)\end{tabular} 
        & \multicolumn{1}{c|}{\textbf{\begin{tabular}[c]{@{}c@{}}90.05\\ (+6.25)\end{tabular}}} 
        & \begin{tabular}[c]{@{}c@{}}64.86\\ (+0.0)\end{tabular} 
        & \begin{tabular}[c]{@{}c@{}}69.61\\ (+4.76)\end{tabular} 
        & \textbf{\begin{tabular}[c]{@{}c@{}}83.73\\ (+18.87)\end{tabular}} 
        & \begin{tabular}[c]{@{}c@{}}54.71\\ (+0.0)\end{tabular} 
        & \begin{tabular}[c]{@{}c@{}}59.64\\ (+4.93)\end{tabular} 
        & \textbf{\begin{tabular}[c]{@{}c@{}}62.32\\ (+7.61)\end{tabular}}  \\ \hline
VolMinNet & \begin{tabular}[c]{@{}c@{}}88.59\\ (+0.0)\end{tabular} 
        & \begin{tabular}[c]{@{}c@{}}88.93\\ (+0.35)\end{tabular}  
        & \multicolumn{1}{c|}{\textbf{\begin{tabular}[c]{@{}c@{}}91.61\\ (+3.02)\end{tabular}}} 
        & \begin{tabular}[c]{@{}c@{}}85.37\\ (+0.0)\end{tabular} 
        & \begin{tabular}[c]{@{}c@{}}85.94\\ (+0.57)\end{tabular} 
        & \multicolumn{1}{c|}{\textbf{\begin{tabular}[c]{@{}c@{}}90.42\\ (+5.05)\end{tabular}}} 
        & \begin{tabular}[c]{@{}c@{}}72.35\\ (+0.0)\end{tabular} 
        & \begin{tabular}[c]{@{}c@{}}73.88\\ (+1.53)\end{tabular} 
        & \textbf{\begin{tabular}[c]{@{}c@{}}81.51\\ (+9.16)\end{tabular}}   
        & \begin{tabular}[c]{@{}c@{}}54.32\\ (+0.0)\end{tabular} 
        & \begin{tabular}[c]{@{}c@{}}56.94\\ (+2.62)\end{tabular} 
        & \textbf{\begin{tabular}[c]{@{}c@{}}59.55\\ (+5.23 )\end{tabular}}  \\ \hline
Coteaching & \begin{tabular}[c]{@{}c@{}}92.79\\ (+0.0)\end{tabular} 
        & \begin{tabular}[c]{@{}c@{}}91.64\\ (-1.16)\end{tabular}  
        & \multicolumn{1}{c|}{\textbf{\begin{tabular}[c]{@{}c@{}}93.48\\ (+0.69)\end{tabular}}}  
        & \begin{tabular}[c]{@{}c@{}}91.59\\ (+0.0)\end{tabular} 
        & \begin{tabular}[c]{@{}c@{}}90.41\\ (-1.18)\end{tabular} 
        & \multicolumn{1}{c|}{\textbf{\begin{tabular}[c]{@{}c@{}}92.54\\ (+0.95)\end{tabular}}} 
        & \begin{tabular}[c]{@{}c@{}}84.30\\ (+0.0)\end{tabular} 
        & \begin{tabular}[c]{@{}c@{}}83.10\\ (-1.20)\end{tabular} 
        & \textbf{\begin{tabular}[c]{@{}c@{}}86.24\\ (+1.93)\end{tabular}}   
        & \begin{tabular}[c]{@{}c@{}}61.07\\ (+0.0)\end{tabular} 
        & \begin{tabular}[c]{@{}c@{}}58.20\\ (-2.87)\end{tabular} 
        & \textbf{\begin{tabular}[c]{@{}c@{}}62.75\\ (+1.68)\end{tabular}} \\ \hline
ELR     & \begin{tabular}[c]{@{}c@{}}92.09\\ (+0.0)\end{tabular} 
        & \begin{tabular}[c]{@{}c@{}}91.66\\ (-0.43)\end{tabular}  
        & \multicolumn{1}{c|}{\textbf{\begin{tabular}[c]{@{}c@{}}93.03\\ (+0.94)\end{tabular}}}  
        & \begin{tabular}[c]{@{}c@{}}91.59\\ (+0.0)\end{tabular} 
        & \begin{tabular}[c]{@{}c@{}}90.97\\ (-0.62)\end{tabular} 
        & \multicolumn{1}{c|}{\textbf{\begin{tabular}[c]{@{}c@{}}92.41\\ (+0.82)\end{tabular}}} 
        & \begin{tabular}[c]{@{}c@{}}86.07\\ (+0.0)\end{tabular} 
        & \begin{tabular}[c]{@{}c@{}}85.48\\ (-0.60)\end{tabular} 
        & \textbf{\begin{tabular}[c]{@{}c@{}}87.42\\ (+1.34)\end{tabular}}   
        & \begin{tabular}[c]{@{}c@{}}62.72\\ (+0.0)\end{tabular} 
        & \begin{tabular}[c]{@{}c@{}}62.56\\ (-0.16)\end{tabular} 
        & \textbf{\begin{tabular}[c]{@{}c@{}}64.65\\ (+1.94)\end{tabular}} \\ \hline
ELR+    & \begin{tabular}[c]{@{}c@{}}94.36\\ (+0.0)\end{tabular} 
        & \begin{tabular}[c]{@{}c@{}}93.35\\ (-1.02)\end{tabular} 
        & \multicolumn{1}{c|}{\textbf{\begin{tabular}[c]{@{}c@{}}94.61\\ (+0.24)\end{tabular}}} 
        & \begin{tabular}[c]{@{}c@{}}93.60\\ (+0.0)\end{tabular} 
        & \begin{tabular}[c]{@{}c@{}}92.53\\ (-1.07)\end{tabular} 
        & \multicolumn{1}{c|}{\textbf{\begin{tabular}[c]{@{}c@{}}94.26\\ (+0.66)\end{tabular}}}  
        & \begin{tabular}[c]{@{}c@{}}89.74\\ (+0.0)\end{tabular} 
        & \begin{tabular}[c]{@{}c@{}}88.59\\ (-1.15)\end{tabular} 
        & \textbf{\begin{tabular}[c]{@{}c@{}}90.54\\ (+0.80)\end{tabular}}  
        & \begin{tabular}[c]{@{}c@{}}63.20\\ (+0.0)\end{tabular} 
        & \begin{tabular}[c]{@{}c@{}}63.26\\ (+0.06)\end{tabular} 
        & \textbf{\begin{tabular}[c]{@{}c@{}}64.89\\ (+1.69)\end{tabular}}  \\ \hline
DivideMix & \begin{tabular}[c]{@{}c@{}}94.99\\ (+0.0)\end{tabular} 
        & \begin{tabular}[c]{@{}c@{}}94.34\\ (-0.66)\end{tabular}  
        & \multicolumn{1}{c|}{\textbf{\begin{tabular}[c]{@{}c@{}}95.46\\ (+0.46)\end{tabular}}} 
        & \begin{tabular}[c]{@{}c@{}}94.90\\ (+0.0)\end{tabular} 
        & \begin{tabular}[c]{@{}c@{}}94.05\\ (-0.84)\end{tabular}  
        & \multicolumn{1}{c|}{\textbf{\begin{tabular}[c]{@{}c@{}}95.52\\ (+0.63)\end{tabular}}} 
        & \begin{tabular}[c]{@{}c@{}}92.24\\ (+0.0)\end{tabular} 
        & \begin{tabular}[c]{@{}c@{}}90.14\\ (-2.09)\end{tabular} 
        & \textbf{\begin{tabular}[c]{@{}c@{}}93.47\\ (+1.23)\end{tabular}}  
        & \begin{tabular}[c]{@{}c@{}}69.29\\ (+0.0)\end{tabular} 
        & \begin{tabular}[c]{@{}c@{}}65.39\\ (-3.90)\end{tabular} 
        & \textbf{\begin{tabular}[c]{@{}c@{}}70.86\\ (+1.57)\end{tabular}}  \\ \hline
UNICON & \begin{tabular}[c]{@{}c@{}}90.82\\ (+0.0)\end{tabular} 
        & \begin{tabular}[c]{@{}c@{}}90.10\\ (-0.72)\end{tabular}  
        & \multicolumn{1}{c|}{\textbf{\begin{tabular}[c]{@{}c@{}}93.90\\ (+3.08)\end{tabular}}} 
        & \begin{tabular}[c]{@{}c@{}}91.87\\ (+0.0)\end{tabular} 
        & \begin{tabular}[c]{@{}c@{}}90.71\\ (-1.15)\end{tabular}  
        & \multicolumn{1}{c|}{\textbf{\begin{tabular}[c]{@{}c@{}}94.22\\ (+2.35)\end{tabular}}} 
        & \begin{tabular}[c]{@{}c@{}}92.33\\ (+0.0)\end{tabular} 
        & \begin{tabular}[c]{@{}c@{}}90.61\\ (-1.71)\end{tabular} 
        & \textbf{\begin{tabular}[c]{@{}c@{}}93.96\\ (+1.63)\end{tabular}}  
        & \begin{tabular}[c]{@{}c@{}}68.33\\ (+0.0)\end{tabular} 
        & \begin{tabular}[c]{@{}c@{}}63.47\\ (-4.87)\end{tabular} 
        & \textbf{\begin{tabular}[c]{@{}c@{}}71.04\\ (+2.70)\end{tabular}} \\ \bottomrule
\end{tabular}
}
\vspace{-0.3cm}
\end{table*}

\subsection{Robustness Comparison}
\label{exp:main_exp}

\subsubsection{Synthetic Label Noise}
\label{subsub:exp:cifar-symm}
We conduct experiments on CIFAR dataset with different levels of symmetric noise, $\varepsilon \in \{20\%, 50\%,70\%\}$.
The overall classification (test) accuracies are provided in Table~\ref{table:cifar-symm}.
The results show that \our{} consistently improves the performance of all LNL methods when combined.
Also, it is noticeable that applying to a standard cross-entropy (CE) method shows the performance better than or comparable to VolMinNet. These results demonstrate that our post-processing of removing influential examples is effective under varying levels of label noise. 
Meanwhile, RoG shows inconsistent gains and fails to improve performance of some baselines like DivideMix and UNICON, which is attributed to the assumption of multivariate Gaussian distribution in feature representations. While we terminate the algorithm after the 2nd round, we show the results on more multiple rounds, and the noisy label detection results in Appendix.

%in Appendix \ref{appendix:sec:exp_multi}.  
% We also present the noisy label detection results in Appendix~\ref{appendix:sec:noise_detection}, and the accuracy after one round of \our in Appendix~\ref{appendix:sec:oneround}.
\vspace{-2mm}

\subsubsection{Real-world Label Noise}

\noindent\textbf{CIFAR-10/100N.}
We further conduct experiment on real-world noisy CIFAR-N in Table~\ref{table:cifar-n}. 
Although real-world noise is more challenging than a synthetic one, a similar trend in synthetic noisy CIFAR has been observed in real-world noisy CIFAR; the performance gain from \our is prone to increase with the increase in the noise ratio, while RoG rather decreases test accuracy in many cases.

\smallskip
\noindent\textbf{Webvision.}
From Table~\ref{table:webvision}, when combining \our with the state-of-the-art robust approach, DivideMix, it achieves the best performance. The top-1 accuracy of $76.24\%$ of DivideMix is further increased to $77.88\%$. 
In addition, it is noteworthy that our post-processing with the basic method CE shows superior performance to other complex LNL methods, such as Co-teaching and Iterative-CV.

\begin{table}[t!]
\caption{
\textbf{Comparison on WebVision with real-world label noise of $20\%$.}
The top-1 top-5 test accuracy. The results are taken from~\cite{ref:dividemix_iclr_2019} and~\cite{ref:elr_neurips_2020}.
$\ast$ is re-trained in our experimental setup using the official code for post-training.
%since there is no pre-trained model publicly available.
%, and $\dagger$ refers to the result obtained from running the official code. 
%The best and the second best results are marked in bold and underlined, respectively.
}
%\vspace{0.1cm}
\label{table:webvision}
\centering
\small
\resizebox{1\linewidth}{!}
{
\begin{tabular}{l|cc|cc}
%\begin{tabular}{L{3.3cm} X{1.cm} X{1.cm}X{1.cm}X{1.cm}}
\toprule
\multirow{2}{*}{Method} & \multicolumn{2}{c|}{WebVision}                 & \multicolumn{2}{c}{ILSVRC12}                  \\
                        & Top-1                 & Top-5                  & Top-1                 & Top-5                 \\ \hline
%F-correction~\cite{ref:patrini_qu_cvpr2017}  & 61.12 & 82.68 & 57.36 & 82.36 \\
%Decoupling~\cite{ref:decouple_2017_nips}\!\!    & 62.54 & 84.74 & 58.26 & 82.26\\
%D2L~\cite{ref:d2l_2018_icml}           & 62.68 & 84.00 & 57.80 & 81.36 \\
MentorNet~\cite{ref:mentornet_icml_18}     & 63.00 & 81.40 &57.80 & 79.92 \\
Co-teaching~\cite{ref:coteacing_nips_18}   & 63.58 & 85.20 &61.48 & 84.70 \\
Iterative-CV~\cite{ref:chen2019understanding}  & 65.24 & 85.34 &61.60 & 84.98 \\
ELR~\cite{ref:elr_neurips_2020}           & 76.26 & 91.26 & 68.71 & 87.84 \\
ELR+~\cite{ref:elr_neurips_2020}           & {77.78} & \underline{91.68} & 70.29 & 89.76 \\
DivideMix~\cite{ref:dividemix_iclr_2019} (reported)     & 77.32 & 91.64 & 75.20 & 90.84 \\
DivideMix~\cite{ref:dividemix_iclr_2019}$^\ast$ (reproduced)     & 76.24 & 91.40 & 73.44 & 91.60 \\
UNICON~\cite{ref:unicon_2022}      & \underline{77.60} & \textbf{93.44} & \textbf{75.29} & \textbf{93.72} \\
\midrule
CE + \our & \multicolumn{1}{l}{72.64} & \multicolumn{1}{l|}{89.20} & \multicolumn{1}{l}{69.40} & \multicolumn{1}{l}{90.60} \\
DivideMix$^\ast$ + \our & \multicolumn{1}{l}{\textbf{77.88}} & \multicolumn{1}{l|}{{91.56}} & \multicolumn{1}{l}{\underline{75.28}} & \multicolumn{1}{l}{\underline{92.52}} \\ \hline
\end{tabular}
\vspace{-0.2cm}
}
\end{table}

\begin{table}[]
\vspace{-0.1cm}
\caption{
\textbf{Comparison with state-of-the-art methods in test accuracy(\%) on Clothing1M}
Results for baselines are copied from original papers, and $\ast$ are reproduced by the official code.
%The best results are marked in bold.
}
%\vspace{0cm}
\label{table:clothing1m}
\centering
\footnotesize
%\resizebox{0.85\textwidth}{!}
{
\begin{tabular}{lc}
\toprule
Method        & Test Accuracy \\ \midrule
Cross-Entropy & 69.21         \\
%F-correction~\cite{ref:patrini_qu_cvpr2017}  & 69.84         \\
%M-correction~\cite{ref:mcorrection_2019_icml}  & 71.00         \\
Joint-Optim~\cite{ref:joint_optim_2018_cvpr}   & 72.16         \\
VolMinNet~\cite{ref:volminnet_2021_icml}     & 72.42         \\
Meta-Cleaner~\cite{ref:metacleaner_2019_cvpr}  & 72.50         \\
ELR~\cite{ref:elr_neurips_2020}           & 72.87         \\
ELR+~\cite{ref:elr_neurips_2020}           & 74.81         \\
Meta-Learning~\cite{ref:Li_2019_CVPR} & 73.47         \\
P-correction~\cite{ref:yi_pencil_cvpr2019}  & 73.49         \\
DivideMix (reported)~\cite{ref:dividemix_iclr_2019}     & 74.76         \\
DivideMix$^\ast$ (reproduced)~\cite{ref:dividemix_iclr_2019}     & 74.23      \\
DivideMix$^\ast$ (longer)~\cite{ref:dividemix_iclr_2019}  & 74.42  \\
UNICON~\cite{ref:unicon_2022}     & \textbf{74.98}      \\
\hline
CE + Ours          &   72.80            \\
DivideMix$^\ast$ + Ours & \underline{74.90} \\ \bottomrule
\end{tabular}
\vspace{-0.1cm}
}
\end{table}

\smallskip
\noindent\textbf{Clothing1M.}
In Table~\ref{table:clothing1m}, we compare the classification accuracy of \our with various state-of-the-art methods.
Post-processing with \our to the basic training with CE loss improves the performance with a significant gap, outperforming many recent baselines.
Also, applying \our to DivideMix outperforms the state-of-the-art methods. 
It is noteworthy that just increasing the number of training epochs cannot bring the meaningful improvement (\emph{i.e.} DivideMix$^\ast$ (longer)). 
While UNICON shows the superior performance, they train much longer hours with 350 epochs. 
Also, we believe that further performance improvement can be obtained if \our is applied for multiple rounds.
%While \our is applied for one round for fair comparison, 

\subsection{Empirical Analysis}
\label{sec:detailed comparison}
\vspace{-1mm}

\subsubsection{Comparison with Small-loss Removal}
\vspace{-2mm}
In this section, we show that \our can be more effective for post-training the pre-trained model than using `small loss' tricks, which existing methods rely on.
%We first qualitatively compare the performance of identifying mislabeled examples for robust training. 
%As a qualitative result, Figure~\ref{fig:mislabeledexamples} shows some mislabeled example where the small-loss trick failed to detect, but \our succeeded; that is, the examples with small losses but with high influence scores. 
%These examples are mostly very `confusing' examples to classify their clean labels, thus exhibiting small losses similar to the clean examples. 
%After training is over, it is difficult to select them as the mislabeled ones using the small-loss trick. However, \our successfully identifies them to rule out from training data, in that they exhibit high influence scores during post-training. 
%Therefore, \our can synergize when combined with other small-loss tricks, such as DivideMix, by providing a new perspective to find such mislabeled examples, which are not detectable by themselves.
%
%During training, some samples have small losses and high confidence score that small-loss tricks (i.e., DivideMix) fail to detect them.
%However, \our can detect these samples since they have high influence scores. Therefore, \our is complementary to the existing methods.
%
%\subsection{Ablation Study}
%\label{sec:ablation}
%\subsubsection{Example with High Influence Score}
%\label{subsec:examples}
%\ref{fig:examples}

First, we quantitatively show our overfitting scores are superior to the small-loss trick for post-training. 
Specifically, the loss of each example is used instead of the overfitting scores in Eqs.\,\eqref{eq:OMscore} and \eqref{eq:ODscore} for removing mislabeled examples.
Hence, we design a modified version we call `CE + Small-loss', which excludes high-loss examples following our proposed post-training pipeline. %, where DivideMix is used for pre-training. 
Table \ref{table:our_loss} compares \our with the modified version of robust post-training on CIFAR-10 with synthetic and real-world label noise. It is observed that the  \our{} provides a much larger improvement compared to loss-based removal.
% We argue that this is because since robust methods (\emph{i.e.}, DivideMix) use small-loss tricks during training, using the same criterion in post-training could not be beneficial. 
% However, our overfitting scores can provide a new perspective to identify `confusing' examples with incorrect labels.
%Figure~\ref{fig:mislabeledexamples} shows some mislabeled examples where the small-loss trick failed to detect, but \our succeeded. 

\begin{figure}[h]
\vspace*{-0.2cm}
\centering
\subfigure[Training Loss.]{
\includegraphics[width=0.45\columnwidth]{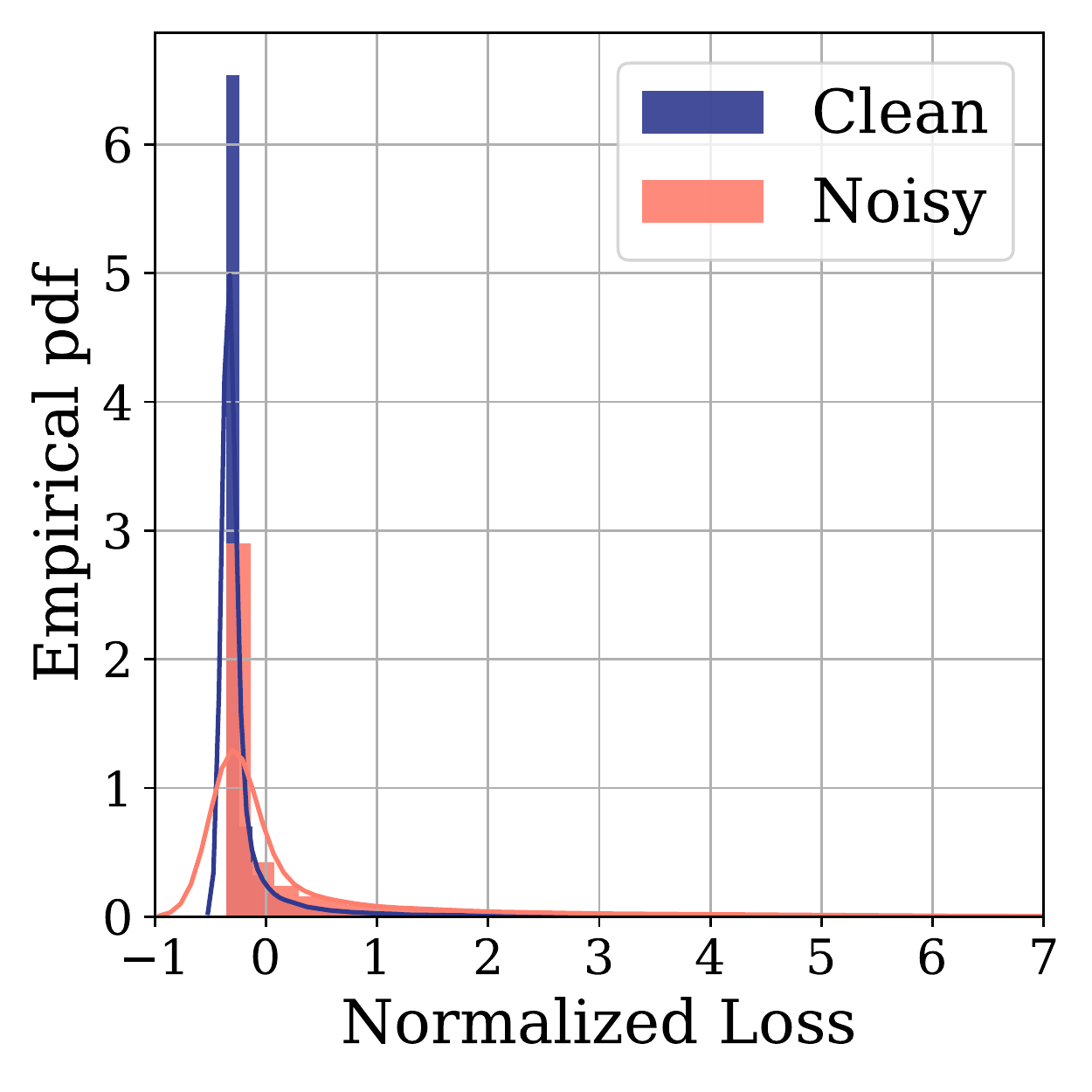}
}
\subfigure[OSM.]{
\includegraphics[width=0.45\columnwidth]{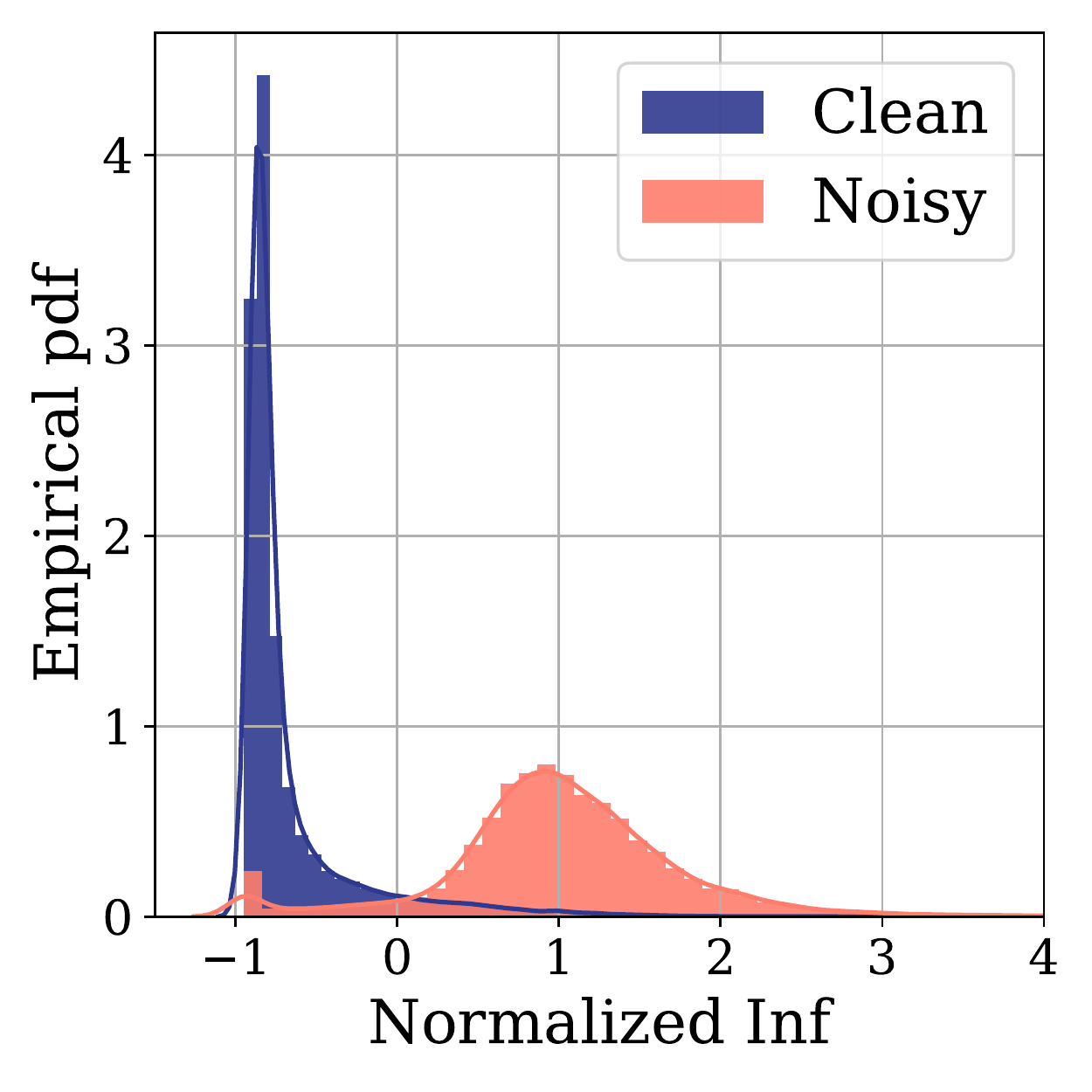}
}
%\subfigure{\includegraphics[width=0.48\columnwidth]{fig/loss.pdf}}
%\subfigure{\includegraphics[width=0.48\columnwidth]{fig/inf.pdf}} 
% \hspace*{1.1cm}  {\small (a) Training Loss.} \hspace*{1.5cm} {\small (b) OSM.}
%\vspace*{0.1cm}
\caption{ \textbf{Loss and OSM distribution for all noisy training examples} after training CIFAR-10 with symmetric noise of $40\%$.
%\vspace*{-0.4cm}
}
\label{fig:gmm}
\end{figure}

Next, Figure~\ref{fig:gmm} compares the distribution of the normalized loss and OSM of training samples on the pretrained model with DivideMix. 
Since training losses are distributed close to 0, it is difficult to classify clean and mislabeled samples with losses after training is done.
However, we argue that OSM can provide a new perspective to identify `confusing' examples with incorrect labels.

\begin{table}[h]
\caption{
{\textbf{Comparison with post-training using the small-loss trick on CIFAR-10 with synthetic and real-world noise.} We report the best test accuracy (\%).}
}
\label{table:loss_selection}
\centering
\vspace*{-0.0cm}
%tabcolsep=0.1cm
\small
\resizebox{0.9\linewidth}{!}
{
\begin{tabular}{lcc}
\toprule
Method           & \!CIFAR-10\,(Symm-70) & CIFAR-10N\,(Worst) \\ \hline
CE        & 29.91             & 63.94             \\
CE + Small-loss   & 53.43              & 76.16              \\
CE + Inf. Rank & \textbf{75.98}    & \textbf{84.27}             \\ \hline
\end{tabular}
}
\vspace*{-0.4cm}
\label{table:our_loss}
\end{table}

\begin{table}[t!]
\caption{
\textbf{Mean test accuracy of training with longer epochs (`+ Longer')} on CIFAR-10 with synthetic and real-world label noise. 
}
\vspace*{-0.05cm}
\label{table:more_training}
\centering
\small
\resizebox{1\columnwidth}{!}{
\begin{tabular}{l|ccc|ccc}
\toprule
& \multicolumn{3}{c|}{CIFAR-10 (Symm-70)} & \multicolumn{3}{c}{CIFAR-10N (Worst)} \\ \cline{2-7} 
    & Original     & +Longer      & +Inf.Rank      
    &  Original    & +Longer     & +Inf.Rank     \\ \midrule
CE      & 28.42      & 29.60      & \textbf{70.59}         
        & 64.86      & 66.92     & \textbf{83.73}        \\
VolMinNet~\cite{ref:volminnet_2021_icml} & 33.69      & 35.09      & \textbf{66.07}         & 72.35      & 72.81     & \textbf{81.51}        \\
Coteaching~\cite{ref:coteacing_nips_18}       & 52.63      & 53.51      & \textbf{60.95}         
        & 84.30      & 84.83     & \textbf{86.24}        \\
ELR~\cite{ref:elr_neurips_2020}       & 77.26      & 77.83      & \textbf{80.13}         & 86.07      & 86.18     & \textbf{87.42}        \\
ELR+~\cite{ref:elr_neurips_2020}       & 86.94      & 87.59      & \textbf{88.21}         & 89.74      & 00.00     & \textbf{90.54}        \\
DivideMix~\cite{ref:dividemix_iclr_2019} & 91.27       &   92.00     & \textbf{92.42}          & 92.24       &    92.46     & \textbf{93.47}        \\
UNICON~\cite{ref:unicon_2022} & 91.44       &   92.28     & \textbf{93.12}          & 92.33       &    93.18     & \textbf{93.96}         \\ \bottomrule
\end{tabular}
}
\vspace*{-0.05cm}
\end{table}

\setlength{\textfloatsep}{7pt}
\begin{figure}[h]
\hspace*{-0.3in}
\centering
    \includegraphics[width=0.8\linewidth]{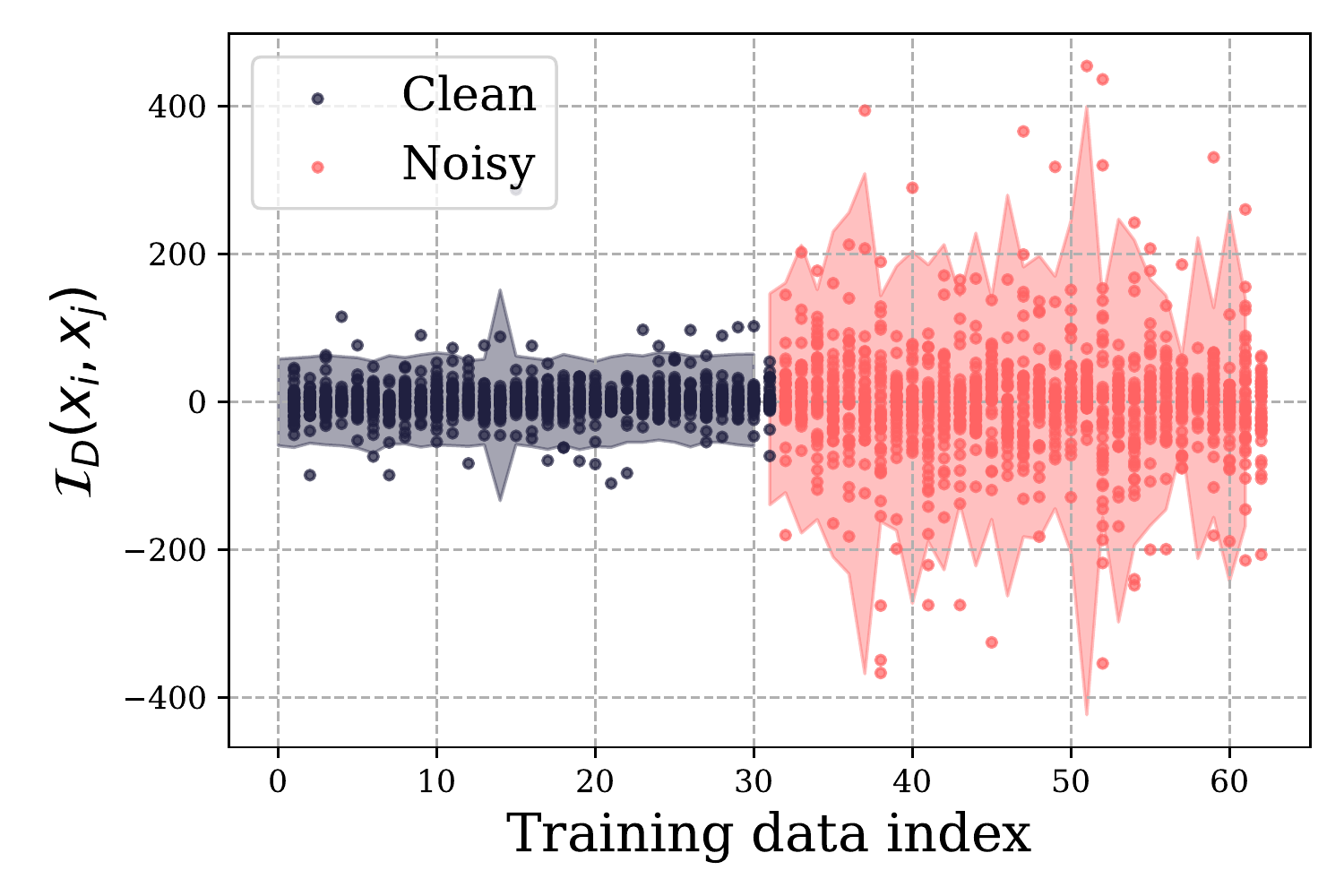}
    \vspace*{-3mm}
    \caption{\textbf{OSD distribution  of training samples on validation samples.} Shaded areas show the variance of ${\cal I}_D$s of each training sample. The difference in variance between the clean and noisy sets is clearly distinguished.}
  \label{fig:inf_by_category}
\vspace{-2mm}
\end{figure}

\begin{figure*}[t!]
%     \subfigure[Mislabeled as `Swing baseball'.]
%     \includegraphics[width=0.67\columnwidth]{fig/mislabeled_hmdb/swb.png}
% \hspace*{0.05cm}
%     \subfigure[Mislabeled as `Jump'.]
%     \includegraphics[width=0.67\columnwidth]{fig/supp_mislabel/jump1.png}
% \subfigure[Mislabeled as `Run'.]
% \includegraphics[width=0.67\columnwidth]{fig/supp_mislabel/run1.png}
\subfigure[Mislabeled as `Swing baseball'.]{\includegraphics[width=0.67\columnwidth]{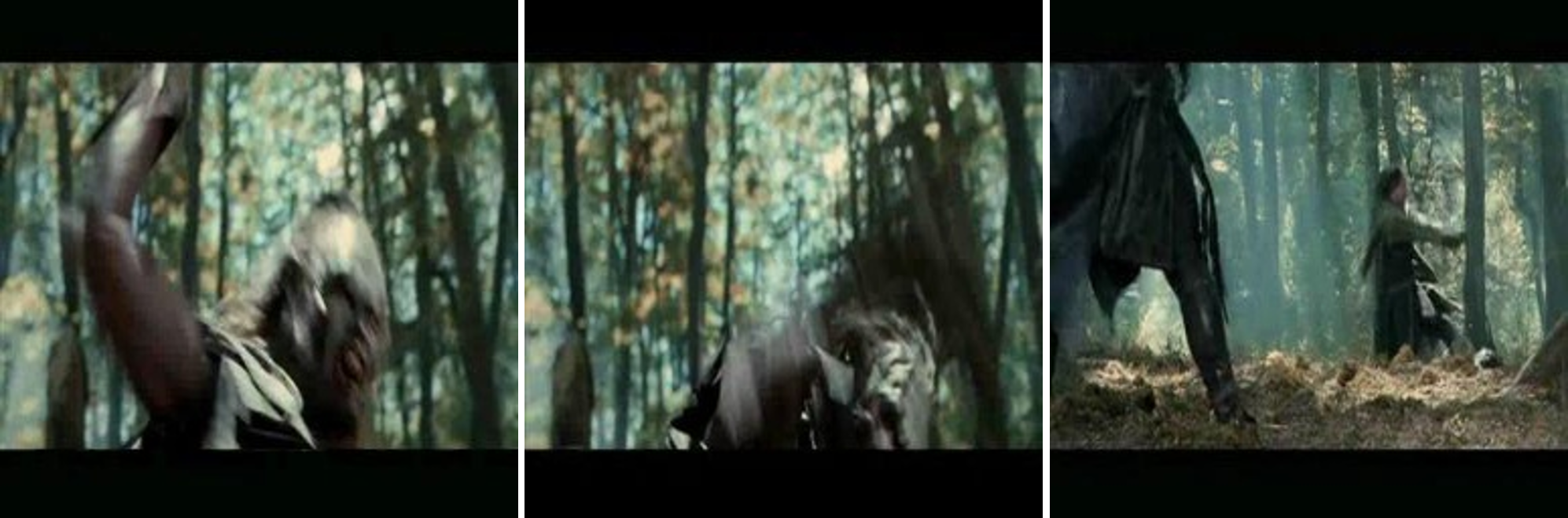}}\hspace*{0.05cm}
\subfigure[Mislabeled as `Jump'.]{\includegraphics[width=0.67\columnwidth]{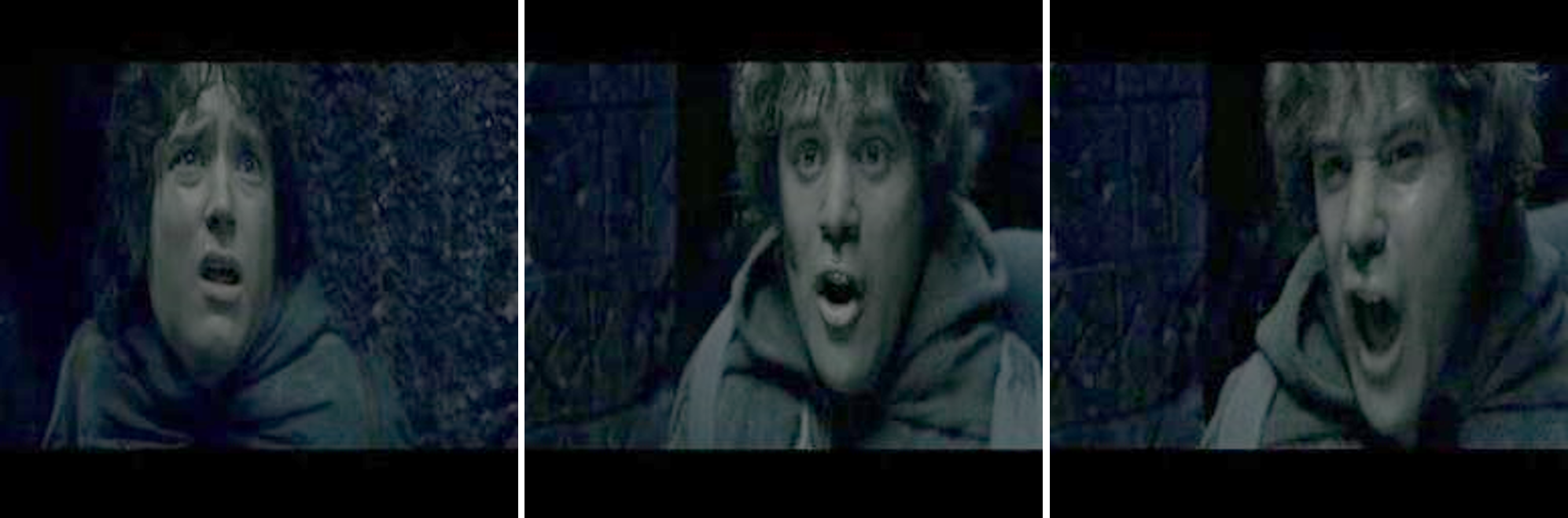}}
\subfigure[Mislabeled as `Run'. ]{%(Climb stairs?)
\vspace*{-0.3cm}
\includegraphics[width=0.67\columnwidth]{ 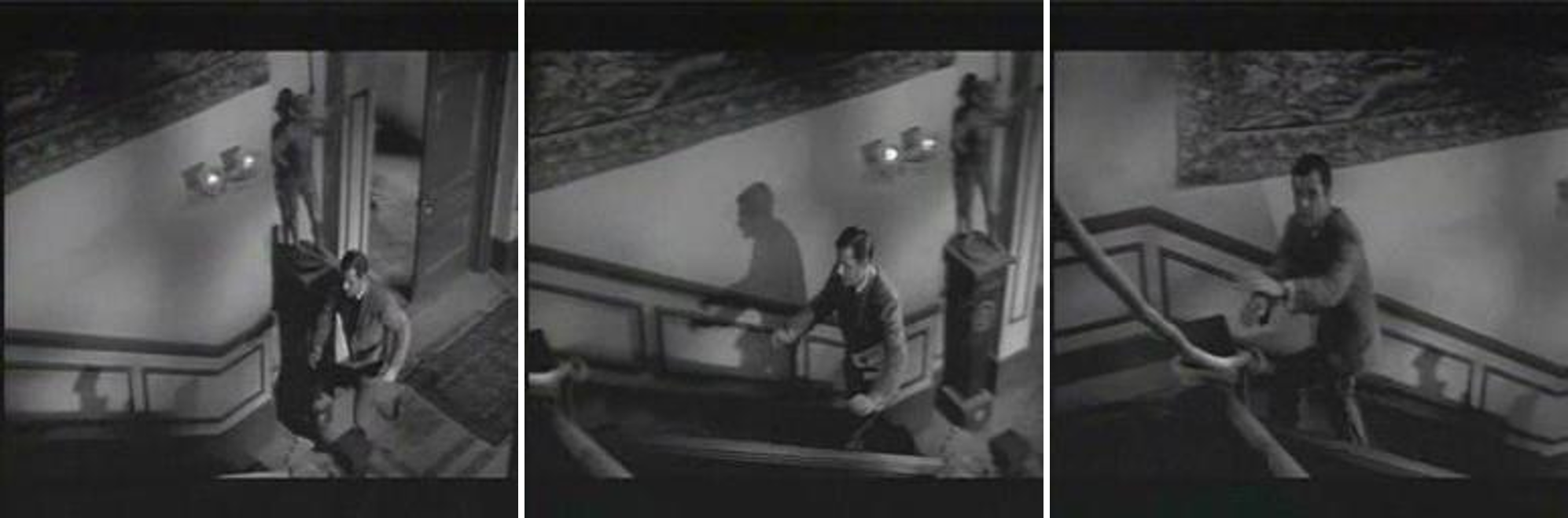}}
\centering \caption{\textbf{Examples with high OSM of training data.} In the video clips, the labeled actions did not appear. Surprisingly, a considerable number of video clips with incorrect labels are detected, which can be found in Appendix. }%~\ref{appendix:sec:video}.}
\label{fig:noisy_video}
\vspace*{-0.45cm}
\end{figure*}

\vspace*{-0.3cm}
\subsubsection{Training with Longer Epochs}
It is of interest to see whether or not the performance improvement comes from additional training epochs used for post-training, though it is reasonably shorter than the total number of epochs used for pre-training.
Table \ref{table:more_training} shows the performance of the existing state-of-the-art robust methods when training the model with longer epochs, where the number of post-training epochs (\emph{i.e.}, 40) is added to the original epochs (see the columns marked with `+Longer').
%To verify whether the performance improvement comes from more training epochs, we compare the results of \our with more training epochs.
%Without removing influential samples, we train the baseline methods for additional 20 epochs.
%As shown in Table~\ref{table:more_training}, naively training with more epochs barely leads to performance improvement.
In general, the performance of the robust methods remains similarly even with longer training epochs. 
%In some cases, it rather degrades their performance because the risk of overfitting to noisy labels becomes more severe with the increase in training epochs.
%
Therefore, our post-training approach is more desirable than simply increasing the training epochs.

%Rather, it can degrade the performance since the model can memorize false labels as training progresses~\cite{ref:elr_neurips_2020}.
%Thus, removing influential samples with ~\our{} is critical for post-training.

%\subsubsection{Comparison with sample selection method}
% 1. sample selection (dividemix) 것 사용했을 때 성능 향상 x
% 2. 실제 loss는 별로 안높지만 influence 높은 데이터 중 헷갈리는 것 샘플 뽑기.
%To justify selecting samples with high influence in post-processing, we compare \our and a loss-based sample selection (LS) method.
%Since DivideMix~\cite{ref:dividemix_iclr_2019} uses small-loss tricks to select clean samples, we select noisy samples having small losses for removal.
%Then, we post-train the network following the same experimental setup in ~\ref{sub:exp_setting}.

%As shown in Table~\ref{table:loss_selection}, the loss-based selection provides limited performance gain compared to \our.
%We suspect that this this is because there is little additional information from large-loss samples which already have been fully utilized during training.
%On the other hand, since our method utilizes samples obtained from a different point of view (i.e., the influence score), complementary new information can be used. 
%For example, samples that have been previously excluded in the early-training stage can be newly used in post-processing, which can handle confirmation bias.

\vspace{-3mm}
\subsubsection{Validity of OSD.}\label{subsec:validity_od}
\vspace{-1mm}
%The improvement in noisy-label detection is attributed to both using Overfitting score on data and iterative algorithm. 
To show the validity of OSD, we investigate the distribution of the ${\mathcal I_D}(x_i, x_t^v;\hat{\theta})$ on real-world images.
We use 1,000 `dog' and `fish' images from ImageNet~\cite{ref:data_imagenet}, where 20\% labels are randomly flipped.
After training the model on this noisy dataset, we calculate OSD using 80 clean validation samples. 
The OSD distribution is illustrated in Figure \ref{fig:inf_by_category}
%, which has been obtained from the dog category in the Dog vs. Fish dataset. 
%of influence of  variance of a training instance on validation samples can be a meaningful indicator.
%Figure \ref{fig:inf_by_category} shows the distribution of the influences of training samples on validation samples.
%In order to remove the difference caused by distinct categories, the training data and the validation data are chosen for the dog category.
The horizontal axis is the index of the training data, and the vertical axis is OSD of a training sample $x_i$ on a validation sample $x_t^v$, i.e., ${\mathcal I_D} (x_i, x_t^v;\hat{\theta})$.
We measure OSD on 40 validation samples for each training sample. 
As illustrated in Figure \ref{fig:inf_by_category}, the variation of the influence of a noisy training sample is much larger than a clean training sample.
It verifies our intuition that the mislabeled samples exert much more inconsistent influences on validation data than the clean samples do.
Therefore, 
%on clean validation samples data is significantly volatile than the influences of clean training data vary.
%Because most influence values are distributed around zero,
the variance of influences, $\sigma_{k}({\cal I}_D(x_i, x^v; \hat{\theta}))$ in Eq. (\ref{eq:ODscore}) can be used to find the mislabeled samples.
% than the average of the influences. % can attenuate these effects while variance clearly indicates the inconsistency in influences.
This distribution appears consistently in other categories. 
In addition, we show that clean and noisy labels can be detected by fitting GMM model on OSD in Appendix. %~\ref{appendix:sec:osd_dist}. 
\vspace{-4mm}

\subsubsection{Effects of hyperparameter.}
To analyze the effects of the hyperparameter $\gamma$, we experiment with different values of $\gamma$ on CIFAR-10 trained with DivideMix.
The higher gamma leads to the higher precision, but less data is erased.
Therefore, choosing $\gamma$ is a tradeoff between the more accurate detection and the faster cleansing.
The details are included in Appendix. %~\ref{appendix:sec:gamma}.
\vspace{-1mm}

\subsection{Detector for Video Data Cleaning}
%\subsubsection{Compatibility on real-world video data}
In this section, we show that the proposed overfitting score can be expanded to detecting mislabeled videos. 
Data cleaning for real-world video data is gaining significant attention due to the growth in the popularity of video-based tasks\,\cite{ref:video_2014_cvpr,ref:video_mao2018hierarchical, ref:video_Wu_2021_CVPR}.
However, detecting video clips with incorrect labels are time-consuming for human annotators more than exploring images because it requires to play and watch the video clip one by one; thus, automatic cleaning of video data can help reduce extreme labeling costs.
%the work of noisy-label detection could benefit this area.
Therefore, we extend our work to video action recognition for data cleaning. \looseness=-1
%and demonstrate that our algorithm is easily applicable to various domains. 

We first train the TSN architecture\,\cite{ref:Xiong_Gool_2016} on HMDB-51 data~\cite{ref:hmdb_data} for action recognition. 
Since each video clip has multiple scenes, the overfitting score of the clip is computed by averaging the score for randomly sampled scenes in the clip. 
Then, we filter out mislabeled video clips based on the proposed OSD. 
Figure \ref{fig:noisy_video} shows some examples of detected mislabeled video clips by \our. 
While HMDB-51 has been known to be clean, surprisingly, we observe that some videos are incorrectly labeled and do not contain any scene corresponding to the label.
We include more detected examples and details of implementation in Appendix.
%See Appendix \ref{appendix:sec:video} for 

\subsection{Regularizer for Performance Boosting}\label{sub:reg}
\vspace{-2mm}
%\subsubsection{Regularization to avoid Overfitting}\label{sub:reg}
%In this section, we show that o
As another use case, \our{} can be considered as a regularizer to avoid overfitting, when there is no apparent label noise in training data. Recently, many regularization techniques have been proposed to reduce the generalization gap of DNNs\,\cite{ref:reg_dropout_jmlr_2014, ref:zhang2018mixup}.
%Since DNNs are vulnerable to overfitting even when there is no apparent label noise in data, many regularization methods have been proposed
Our method post-processes the overfitted decision boundary by squeezing out the negative impact of highly influential examples. Thus, it has the potential to be used as regularization to smooth decision boundaries.

\begin{figure}[t!]
\centering
% \subfigure[Original.]
%     {\includegraphics[width=0.45\columnwidth]{fig/res/reg_origin.png}}
% \hspace{0.1cm}
% \subfigure[W. \our.]
%     {\includegraphics[width=0.45\columnwidth]{fig/res/reg_after.png}}
    \subfigure[]{\includegraphics[width=0.49\columnwidth]{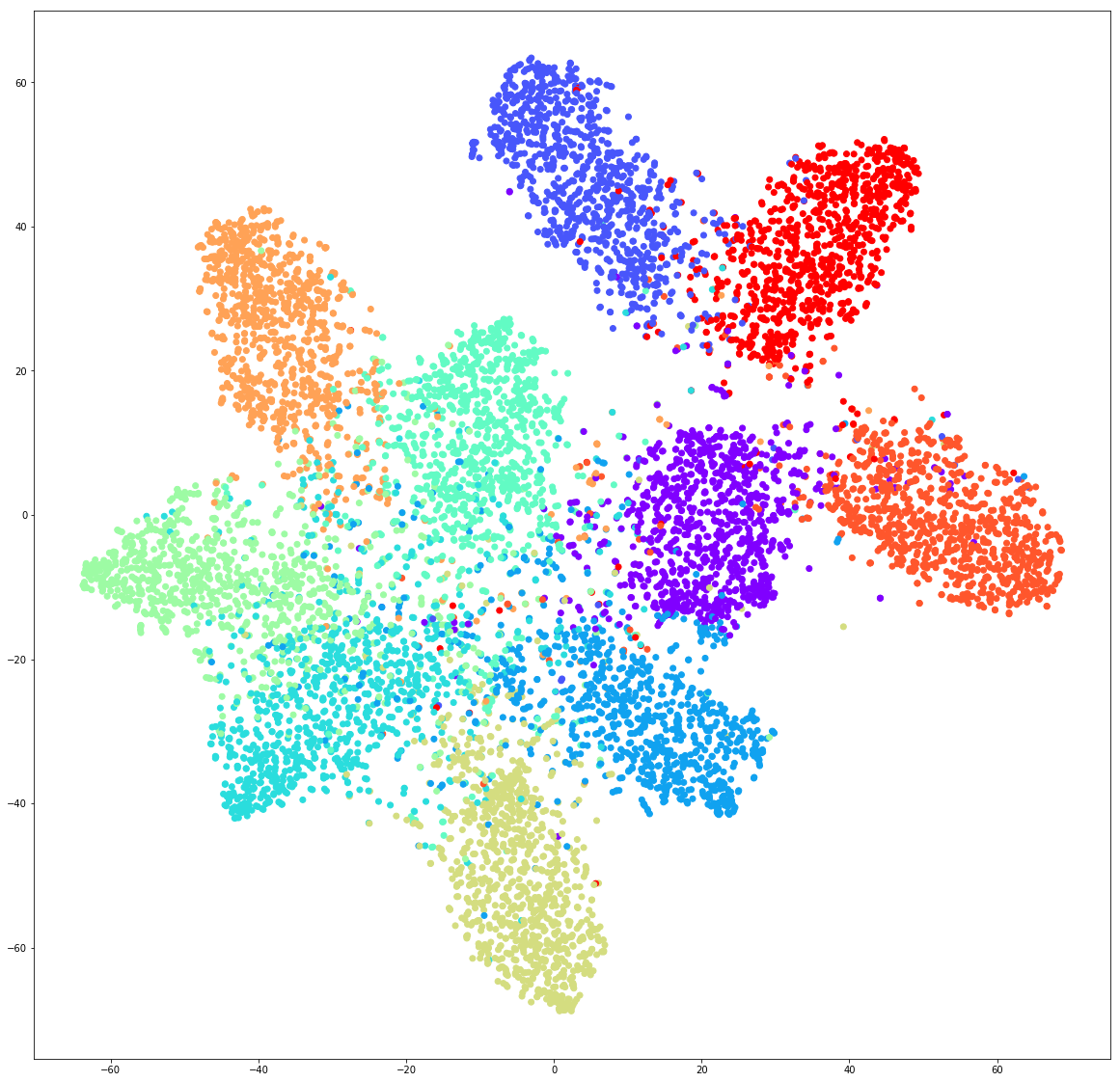}}
    \subfigure[W. \our.]{\includegraphics[width=0.49\columnwidth]{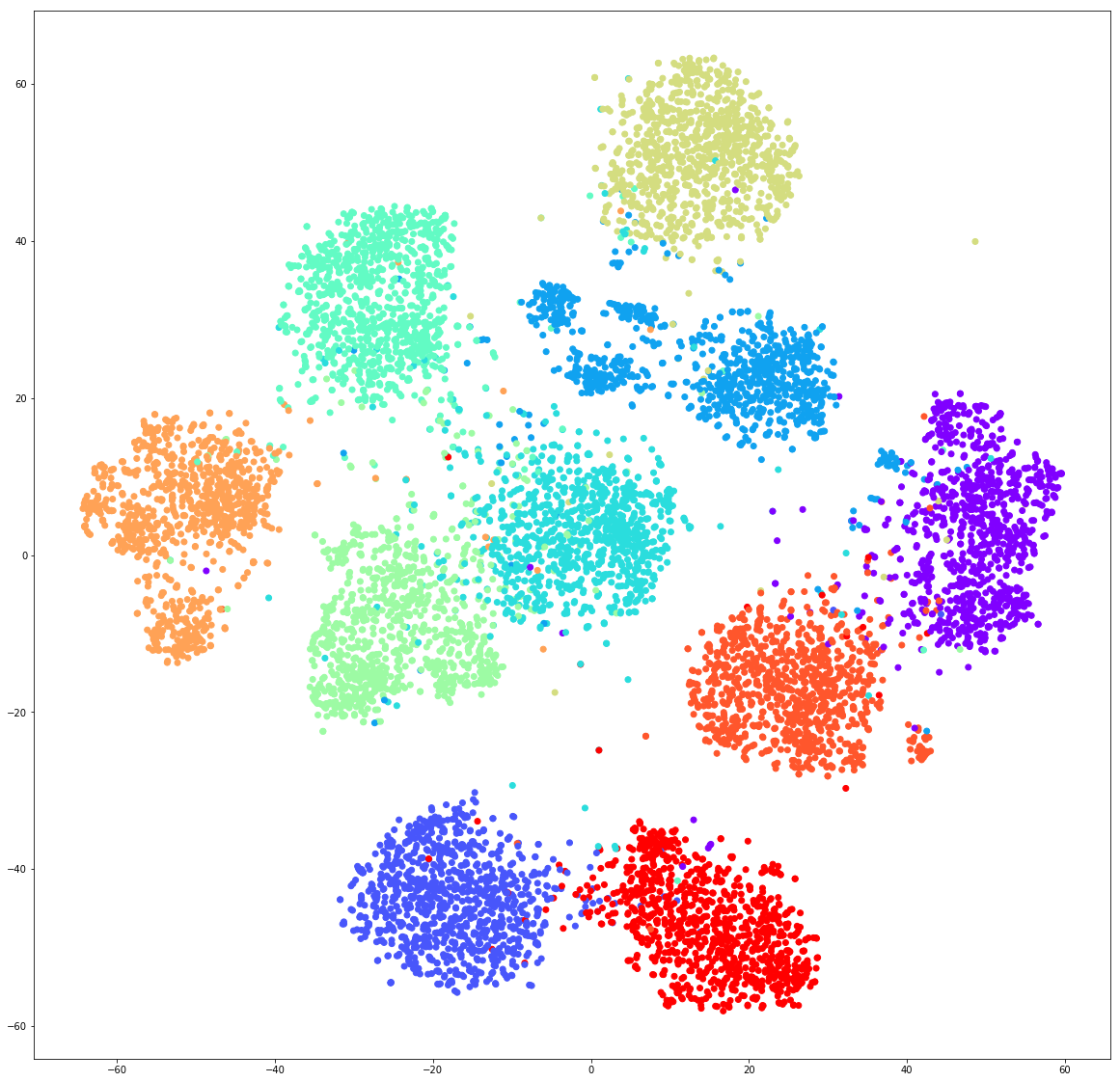}}
    %\vspace*{-0.4cm}
\caption{\textbf{t-SNE visualization for the learned representation of the trained models.}
%We apply \our algorithm to the default model (a). After post-training on the dataset after removing some fraction of clean examples causing over-fitting, the performance (b) improves significantly.  
}
\label{fig:reg_tsne}
\vspace*{-0.1cm}
\end{figure}

As a case study, we conduct an experiment on clean CIFAR-10 using the same experimental configuration. 
Table~\ref{table:regularizer} and Figure~\ref{fig:reg_tsne} shows that \our can also improve the model trained on clean dataset.
%By the post-training, $1,011$ training examples are excluded from $50,000$ training data. 
%When we post-train the model with the rest $48,989$ examples, the test accuracy improves remarkably to \textbf{96.6\%} from 94.2\%. Figure \ref{fig:reg_tsne} shows the t-SNE results for test data using the models trained with and without \our. 
%
%While the default model does not separate classes well, \our helps clearly separate classes as shown in Figure \ref{fig:reg_tsne}(b). %we can see clearly well-represented differences with \our.
We conjecture that this is because \our removes spurious or isolated data points leading the decision boundary astray, and get a well-generalized decision boundary.
%via post-training with \our. 

%As a result, a total of $1,011$ training samples are removed. 
%When trained on a new training data after removal, the performance remarkably improves to 96.6\% from 90.8\%. 
%In addition, we visualize the features on the test dataset in Figure \ref{fig:reg_tsne}.
%While the default model does not separate classes well, the proposed \our can clearly separate classes as shown in the figure. %we can see clearly well-represented differences with \our.
%We conjecture this is because we remove spurious or isolated datapoints leading the decision boundary astray, and get a well-generalized decision boundary. 

\vspace{-0mm}
\begin{table}[t]
\centering\caption{\textbf{Result of \our on clean CIFAR-10.}}
\vspace{-0mm}
\footnotesize
\label{table:regularizer}
\begin{tabular}{l|cc}
\toprule
           & \# of training & Accuracy \\ \midrule
Original   & 50,000         & 94.2     \\
+Inf. Rank & 48,989         & \textbf{96.6}     \\ \bottomrule
\end{tabular}
\end{table}

%, we evaluate our method on a video action recognition dataset, HMDB-51, with the TSN architecture \cite{ref:Xiong_Gool_2016}.
% We used a test set as a validation set and computed ${\cal O}_D^k (x_i^t;\hat{\theta})$ for 3,750 training samples. 
%The data are checked in the order of the highest overfitting score, and we unexpectedly discovered many noisy labels that were supposed to be clean.  
%Figure \ref{fig:noisy_video} presents examples of detected noisy labels in HMDB-51. 
%See Supplementary Material for more examples and details of implementation. 
%According to more examples in the Supplementary Material, some videos are incorrectly labeled and do not contain any scene corresponding to the label, whereas some videos are partly noisy and include scenes corresponding to other labels that seem more suitable.

% \subsection{Further Analysis}
% To further justify our proposed method, the following experiments are included in the Appendix: Effect of hyperparameter $\gamma$, and the components (OSM, OSD).

% 본문에 언급되지 않은 Supplementary의 Analysis에 대한 summary 및 참조 안내 문구 작성.
% Regularizer만 공간 되는 양많큼 넣어도 좋을 듯 ...
% \\

% \noindent
% \textbf{Regularizer for Performance Boosting}.
% Further experiment was conducted to show \our{} can be considered as a regularizer to avoid overfitting, even when there is no apparent label noise in training data. The detailed results are presented in Appendix
% \ref{sub:reg}. \\

% \noindent
% \textbf{ Noise Ratio Reduction after Influential
% Rank}. fmlgvoieiervrgibvmori;sgmbvsdrigmb....\\

% \noindent
% \textbf{ Calculation of Hessian}. mgvdfgvdsgngn fjunuh..
\vspace*{-0.2cm}
\section{Conclusion}
\vspace*{-0.1cm}

%excluding examples with high influence score to 
We have proposed a post-training method named Influential Rank, which sways the overfitted decision boundary to be correct, in the presence of noisy labels.
%In this paper, we proposed a simple but effective post-processing algorithm for fixing overfitted decision boundary in noisy-label circumstances.
%Unlike the existing methods that aim to prevent overfitting during training, 
Unlike the existing methods, \our starts from an overfitted model and makes the model more robust against noisy labels progressively.
%interpret the property of an overfitted model from a new perspective. 
%The proposed influential score eliminates mislabeled examples from noisy data with high precision.
%
We have conducted extensive experiments on real-world and synthetic noisy benchmark datasets. The results demonstrate that \our{} consistently provides performance gain when combined with multiple state-of-the-art robust learning methods. In addition, we have shown that
%the overfitted decision boundary on multiple benchmark datasets.
%Finally, we have shown that 
\our performs as a detector for video data cleaning or a regularizer to smooth the decision boundary.
%As a future study, we will investigate how \our can bring the regularization effect. 
%We include limitations and potential negative societal impact in Appendix~\ref{appendix:sec:limitation}.
% \noindent\textbf{Limitations.}

% \noindent\textbf{Potential negative societal impact.}
\appendix
\twocolumn
\setcounter{figure}{7}
\setcounter{table}{8}
%\numberwithin{equation}{section}
\renewcommand*{\thesection}{\Alph{section}}
\renewcommand{\theequation}{\thesection.\arabic{equation}}

\begin{center}
%\bf {\LARGE Over-Fit: Noisy Label Detection by Novel Criteria based on  Overfitted Model Property}
%Anonymous Author(s)
\bf {\LARGE Supplementary Material}
\end{center}
\begin{figure*}[t!]
\centering
    \subfigure{
    \includegraphics[width=0.23\linewidth]{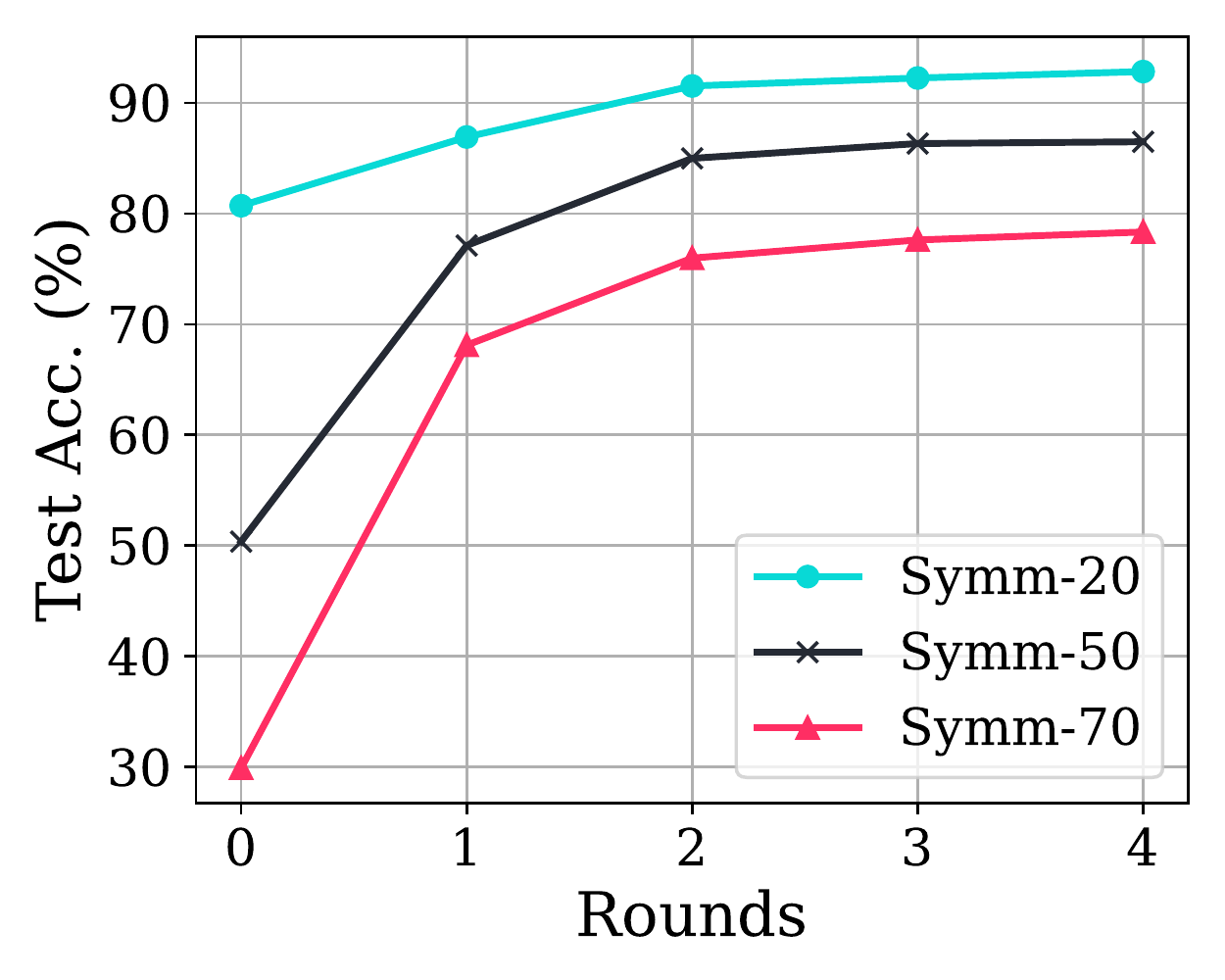}
    }
    \subfigure{
    \includegraphics[width=0.23\linewidth]{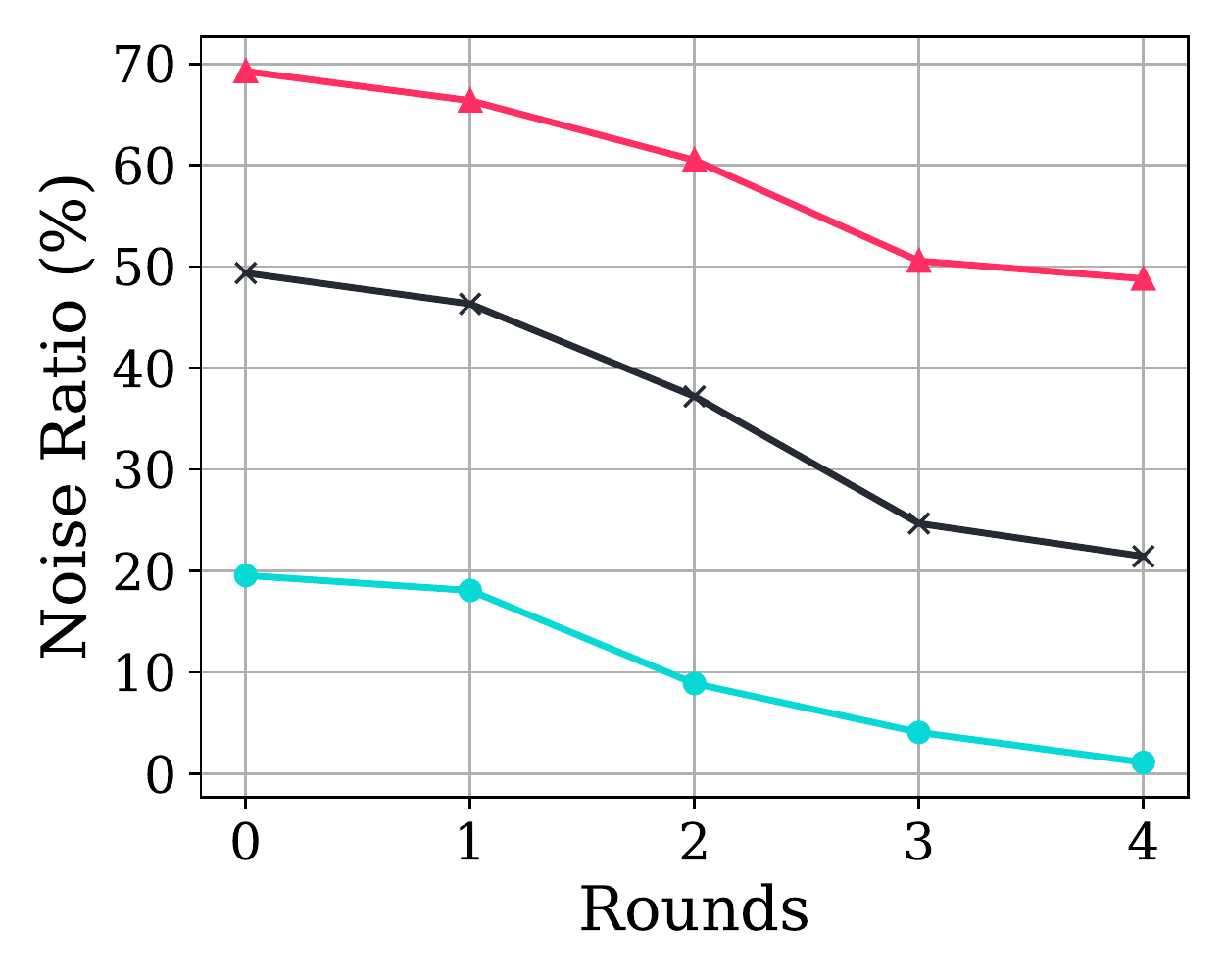}
    }
    \subfigure{
    \includegraphics[width=0.23\linewidth]{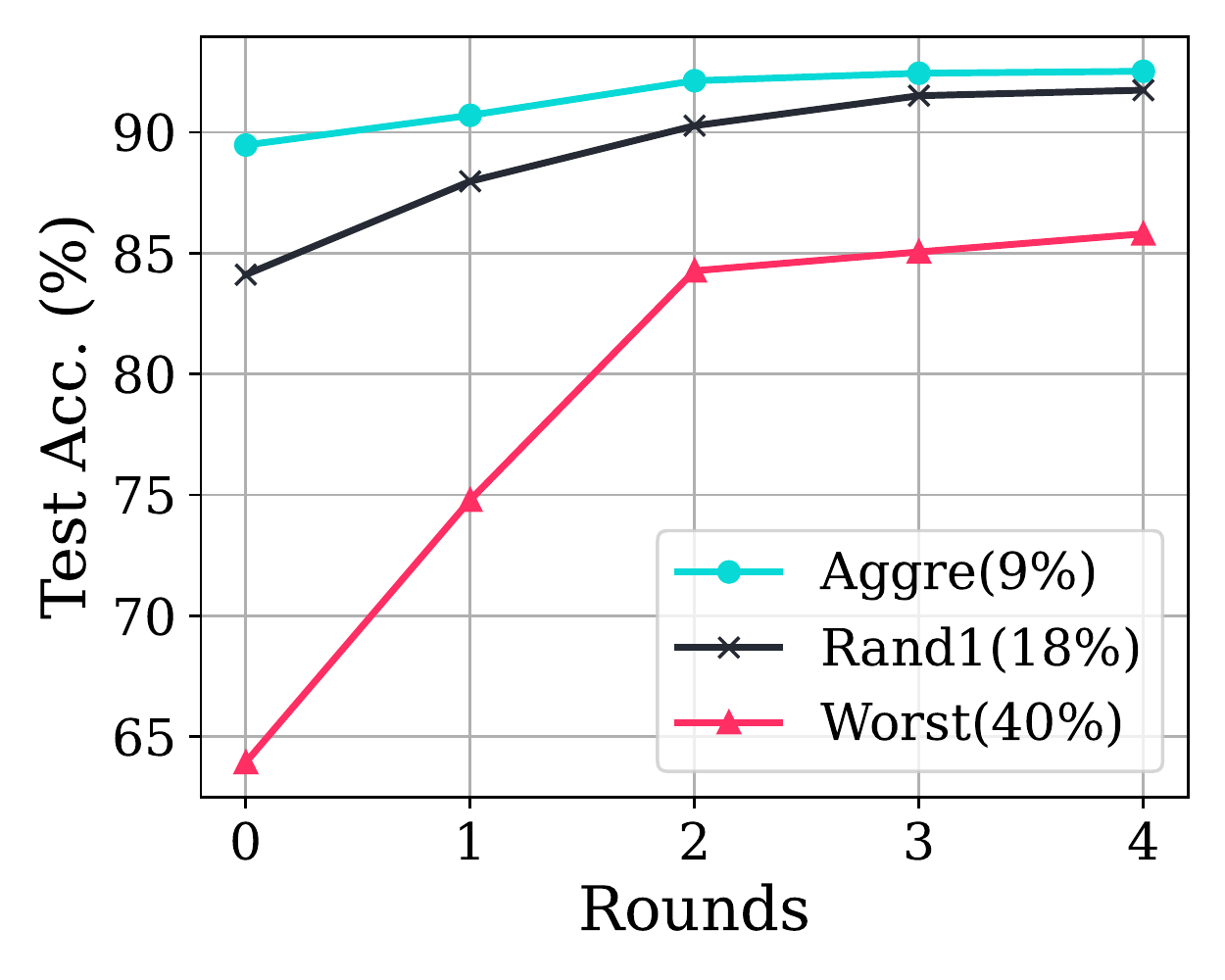}
    }
    \subfigure{
    \includegraphics[width=0.23\linewidth]{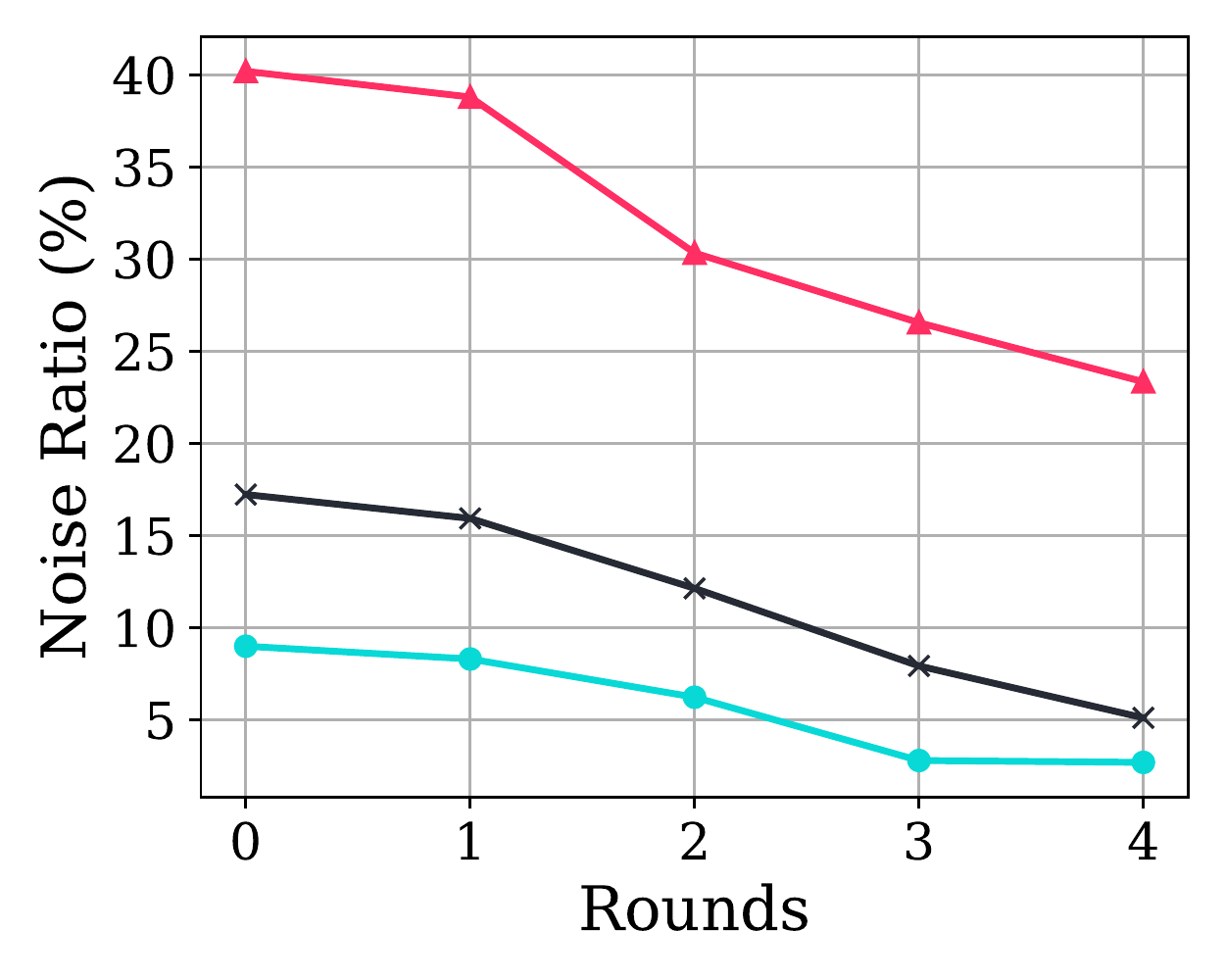}
    }
\vspace*{-0.1cm}
\hspace*{0.2cm} {\small (a) CIFAR-10 with Synthetic Noise.} \hspace*{3.3cm} {\small (b) CIFAR-10N with Real-world Noise.}
%\vspace*{-0.2cm}
\caption{\textbf{Effect of multi-round post-training on CIFAR-10 with synthetic label noise and real-world.}  (Left: Test accuracy over rounds by \our over rounds, Right: Noise ratio of the refined data.)}
\label{fig:acc_error_by_cycle}
\vspace*{-0.2cm}
\end{figure*}

\section{Example of \our: A Binary Classification} \label{appendix:toyexample}
In Figure {\color{red} 3}, yellow and purple circles represent examples of two different classes, and blue and pink shades indicate their decision surfaces.
Next, for the label noise scenario, $40\%$ of the true labels are randomly corrupted in data, \emph{i.e.}, $\times$ marks in Figure~{\color{red} 3}(b). Then, we fit a two-layer feedforward neural network with 50 hidden neurons. 

Figure~{\color{red} 3}(a) shows that the decision boundary trained on clean data is well-formed close to the ground truth.
%Then, we corrupt data by randomly flipping 20\% of the labels ($\times$ mark in Figure~{\color{red} 3}(b)).
However, when trained with noisy labels shown in Figure~{\color{red} 3}(b), we observe that the trained model overfits to mislabeled examples, and forms a complex decision boundary such that many mislabeled examples locate near the overfitted decision boundary.
When we post-train the model after excluding 20 examples with high overfitting scores (\emph{i.e.}, white examples), the overfitted decision boundary begins to recover in Figure {\color{red} 3}(c).
Again, after excluding total 20 more high influential examples after the third iteration in Figure {\color{red} 3}(d), the decision boundary becomes almost similar to that of the clean model. Therefore, this toy example justifies our proposed \our.
%validates the potential of our influence score for robust post-training. %as a new criterion for sample selection.

% \section{Summary of datasets}~\label{appendix:summary_dataset}
% \noindent Table~\ref{table:data_summary} summarizes datasets used in this paper.

% \begin{table}[h]
% \centering\caption{Summary of datasets.}
% \footnotesize
% \resizebox{\linewidth}{!}{
% \label{table:data_summary}
% \begin{tabular}{c|c|c|c|c}
% \toprule
%             & \# of training & \# of testing & \# of class      & type  \\ \hline
% CIFAR-10(N)     & 50,000         & 10,000        & 10           & Image \\ \hline
% CIFAR-100(N)    & 50,000         & 10,000        & 100          & Image \\ \hline
% WebVision 1.0 & 2,400,000          & 50,000           & 1,000           & Image \\ \hline
% Clothing1M   & 1M             & 10,526        & 14          & Image \\ \hline
% HMDB-51       & 3,750          & 1,530         & 51        & Video \\
% \bottomrule
% \end{tabular}
% }
% \vspace*{0.0cm}
% \end{table}

\section{Experimental Setting}\label{appendix:sec:implementation}

\subsection{Datasets}\label{appendix:subsec:dataset}
\footnotetext[1]{There are `random1', `random2', and `random3', but we use `random1' since they have the same noise rate of $18\%$.}
For CIFAR-10 and CIFAR-100, noisy labels are injected using the symmetric noise\,\cite{ref:coteacing_nips_18} of flipping true labels into other labels with equal probability $\varepsilon$, \emph{i.e.}, the noise ratio.
Regarding the real-world noisy data, CIFAR-N\,\cite{ref:cifar-n_2021} has various versions of human noise level.
`aggregate'\,($9\%)$, `random'\,($18\%)$\footnotemark[1], and `worst'\,($40\%)$, while CIFAR-100N has only a single version, `noisy'\,($40\%$). 
Clothing1M includes about $38\%$ real noisy labels, and WebVision 1.0 contains about $20\%$ real-world noisy labels\,\cite{ref:song2020learning}. 
Following the previous work\,\cite{ref:chen2019understanding}, we only use the first 50 classes of the Google image subset in WebVision.
Lastly, we use a video stream data, HMDB-51\,\cite{ref:Simonyan_Andrew_2014}, to verify that our method can be effective as a detector for data cleaning.

To illustrate the applicability of our algorithm to video streams, we experiment on HMDB-51, a popular dataset frequently used in video action recognition  \cite{ref:Simonyan_Andrew_2014}. %CIFAR-N is a real-world human annotated noisy labels (i.e., instance-dependent).
Clothing1M and WebVision 1.0 are large-scale real-world datasets.
Clothing1M includes about $38\%$ real noisy labels and WebVision 1.0 contains about $20\%$ noisy labels.
Following previous work~\cite{ref:chen2019understanding}, we compare baseline methods on the first 50 classes of the Google image subset.
Furthermore, to illustrate the applicability of our algorithm to video streams, we experiment on HMDB-51, a popular dataset frequently used in video action recognition  \cite{ref:Simonyan_Andrew_2014}.

\subsection{Implementation Details}\label{appendix:subsec:dataset}
In this section, we describe more implementation details which are not included in Section~{\color{red} 4.1}.
Following the prior literature\,\cite{ref:elr_neurips_2020}, all the compared methods are trained using ResNet-34, Inception-ResNet V2, and ResNet-50 for CIFAR, WebVision datasets, and Clothing1M respectively. 
For all experiments, the last fully connected (FC) layers in the networks are used as the overfitted classifiers.
In addition, to reduce the number of the classifier parameters, we add a penultimate FC layer with 50, 100, 100 neurons, for CIFAR-100, WebVision 1.0, and HMDB-51, respectively. 
This allows to save the computational cost of hessian computation.
Lastly, for label refinement, we set the threshold $S$ to 0.8. 
We will make our code publicly available after publication.

\subsubsection{CIFAR and CIFAR-N}
All networks are trained for 120 epochs for CIFAR-10(N), and 150 epochs for CIFAR-100(N) with Stochastic Gradient Descent (SGD) (momentum=0.9).
Regarding to training with CE, we set the initial learning rate as 0.1, and reduce it by a factor of 10 after 40 and 80 epochs for CIFAR-10(N). 
For CIFAR-100(N), the initial learning rate is decayed at 60th and 100th epoch by 0.1.
To implement LNL baselines, we set the hyperparameters and training scheme for the baselines as reported in their original papers~\cite{ref:coteacing_nips_18, ref:elr_neurips_2020, ref:dividemix_iclr_2019, ref:volminnet_2021_icml}.
In all experiments, we use the standard data augmentation of horizontal random flipping and 32 × 32 random cropping after padding 4 pixels around images.
Following the recent works, we also adopt the augmentation policy from~\cite{ref:autoaugment_cvpr2019}. 

For the results in Table~{\color{red} 1} and~{\color{red} 3}, the algorithm is applied for 2 rounds with 20 epochs each.
For post-training iteration, we set the initial learning rate as same as the one used in earlier pre-training, and drop it after 5 epochs. 
For cross-entropy (CE) loss, the learning rate at start is set to 0.1 and is decreased by a factor of 0.1 after the 5th and 15th epoch.
By increasing the learning rate high at the first epoch in each retraining iteration, we can encourage the network to explore a newly updated dataset and form a new classifier.
We apply RoG and \our to the models from the last epoch.
\our{} and RoG both use 500 validation samples.
Experiments are conducted with three different noise realizations and the averaged test accuracies are reported.

\subsubsection{WebVision}
For WebVision 1.0, we use inception-resnet v2~\cite{ref:inception_resnet_2017} following~\cite{ref:chen2019understanding}.
For fair comparison with other baselines, both networks are trained for 80 epochs first, and then post-trained with \our for 20 epochs. 
We train the network with CE loss for 80 epochs using the SGD optimizer (momentum=0.9) with an initial learning rate 0.01, which is divided by 10 after 50 epochs.
When training with DivideMix~\cite{ref:dividemix_iclr_2019}, we follow the setting in their original paper. 
After 1 round of \our, 5K and 6K highly influential examples are removed in CE and DivideMix, respectively. 
When post-training, both networks are trained for 20 epochs with a learning rate 0.01, and the learning rate is dropped to 0.01 after 10 epochs.

\subsubsection{Clothing1M}
For Clothing1M, the network is initially trained for 80 epochs with learning rate $0.002$ which is decreased by a factor of 0.1 after 40 epochs. 
We set a batch size to 64, and train the network using SGD optimizer (momentum=0.9) with CE.
When training with DivideMix~\cite{ref:dividemix_iclr_2019}, we follow the setting in their original paper. 
After 1 round of \our, 140K and 230K highly influential examples are removed in CE and DivideMix, respectively. 
For post-training with CE, the model is trained with a learning rate of 0.002 for 10 epochs and then the learning rate is dropped to 0.0002.
For post-training with DivideMix, the model is trained with a learning rate of 0.0002 for 10 epochs and then the learning rate is dropped to 0.00002.

 \begin{figure*}[t]
 \vspace*{-0.33cm}
 \begin{center}
 \subfigure[Precision.]{
 \includegraphics[width=0.85\columnwidth]{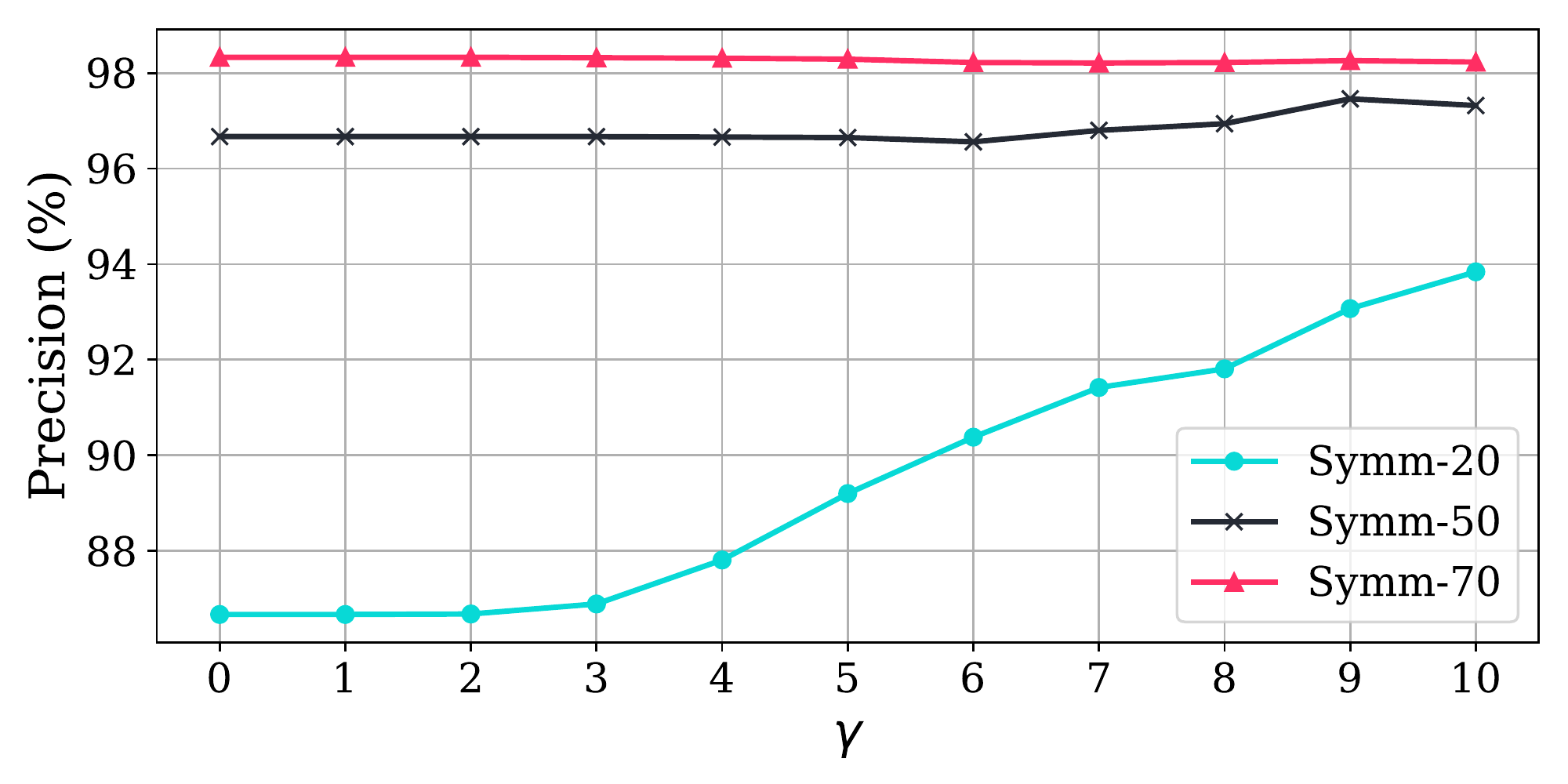}
    }
 \hspace{5mm}
  \subfigure[Remaining Noise.]{
 \includegraphics[width=0.85\columnwidth]{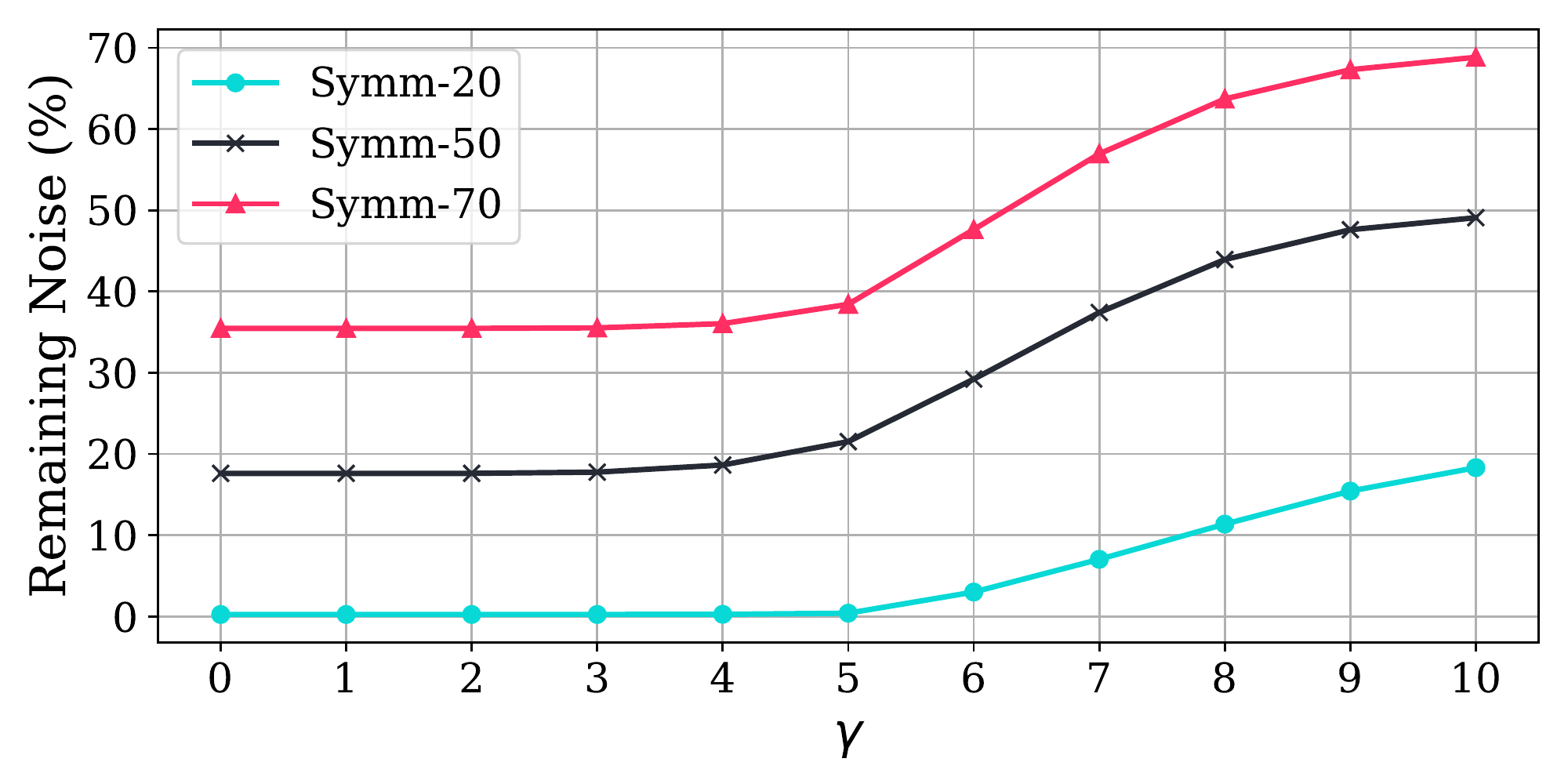}
    }
 \end{center}
% % \vspace*{0.1cm}
 \caption{ \textbf{Effects of $\gamma$.} \our is applied to the model trained on CIFAR-10 with DivideMix.}
 \label{appendix:fig:gamma}
 \end{figure*}

\subsection{Calculation of Hessian}\label{appendix:sec:hessian}
%We implement some tricks for more pragmatic approach.
We calculate the Hessian matrix using only sampled $n$ ($\ll N$) data to reduce the computation cost, which is a reasonable approximation by the law of large numbers when the volume of training data is large. 
For deep neural networks (DNNs), the Hessian matrix could not be positive definite, so we added a positive constant 0.01 to the diagonal following~\cite{ref:Koh_Liang_2017}.
To efficiently calculate the inverse of the Hessian matrix, we also adopt the conjugate gradient method from optimization theory.
The conjugate methods do not require explicitly computing the inverse of the hessian, thus computational complexity is only $O(np)$, where $p$ is the number of parameters of the last fully connected layer.
In most cases, we simply use open library to calculate the inverse of the Hessian because the number of parameters is sufficiently reduced and many open libraries, (e.g., NumPy), provide optimized solutions.

\section{Further Analysis}
\subsection{Effects of Multi-round Post-training}
\label{appendix:sec:exp_multi}
To verify the potential benefit of using multi-round post-training, we set the number of total rounds to $4$, and post-train the network, which is pre-trained using the plain CE. 
Figure \ref{fig:acc_error_by_cycle} depicts the effect of the multi-round post-training on CIFAR-10 and CIFAR-10N, where the round $0$ means the model before any post-training.  
Overall, the noise ratio of the refined data by \our reduces gradually as the round goes up. 
In CIFAR-10 of Figure \ref{fig:acc_error_by_cycle}(a), the test accuracy is largely improved to 92.83\%, 86.49\%, and 78.34\% from the initial accuracy of 80.71\%, 50.37\%, 29.91\%, respectively. 
In addition, the initial noise ratios of 20\%, 50\%, and 70\% become 1.12\%, 21.43\%, and 48.81\% at the final round of post-training. 
Consistently, this improvement trend is exactly the same in CIFAR-10N with real-world noise in Figure \ref{fig:acc_error_by_cycle}(b). Particularly, the improvement in noise ratio and test error becomes larger when data is corrupted with heavier noise.
While performance increase can be expected with multi-rounds, we discover that setting only 2-3 rounds can be sufficiently beneficial in terms of increasing computational burdens. 

% In addition to CE loss, we apply more rounds to the state-of-the-art method DivideMix in Table \ref{table:more_round_dm}.
% We set the maximum number of rounds to $2$ due to the high computational cost of DivideMix. 
% As shown in the table, the improvement trend also remains consistently. 
% For the complete results, we present the improvement in noise ratio for our main experiments \emph{i.e.}, Tables \ref{table:cifar-symm} and \ref{table:cifar-n}, in Appendix \ref{sec:complete_noise}.

 \begin{figure}[h]
 \vspace*{-0.3cm}
 \begin{center}
 \subfigure[CE]{
 \includegraphics[width=0.45\columnwidth]{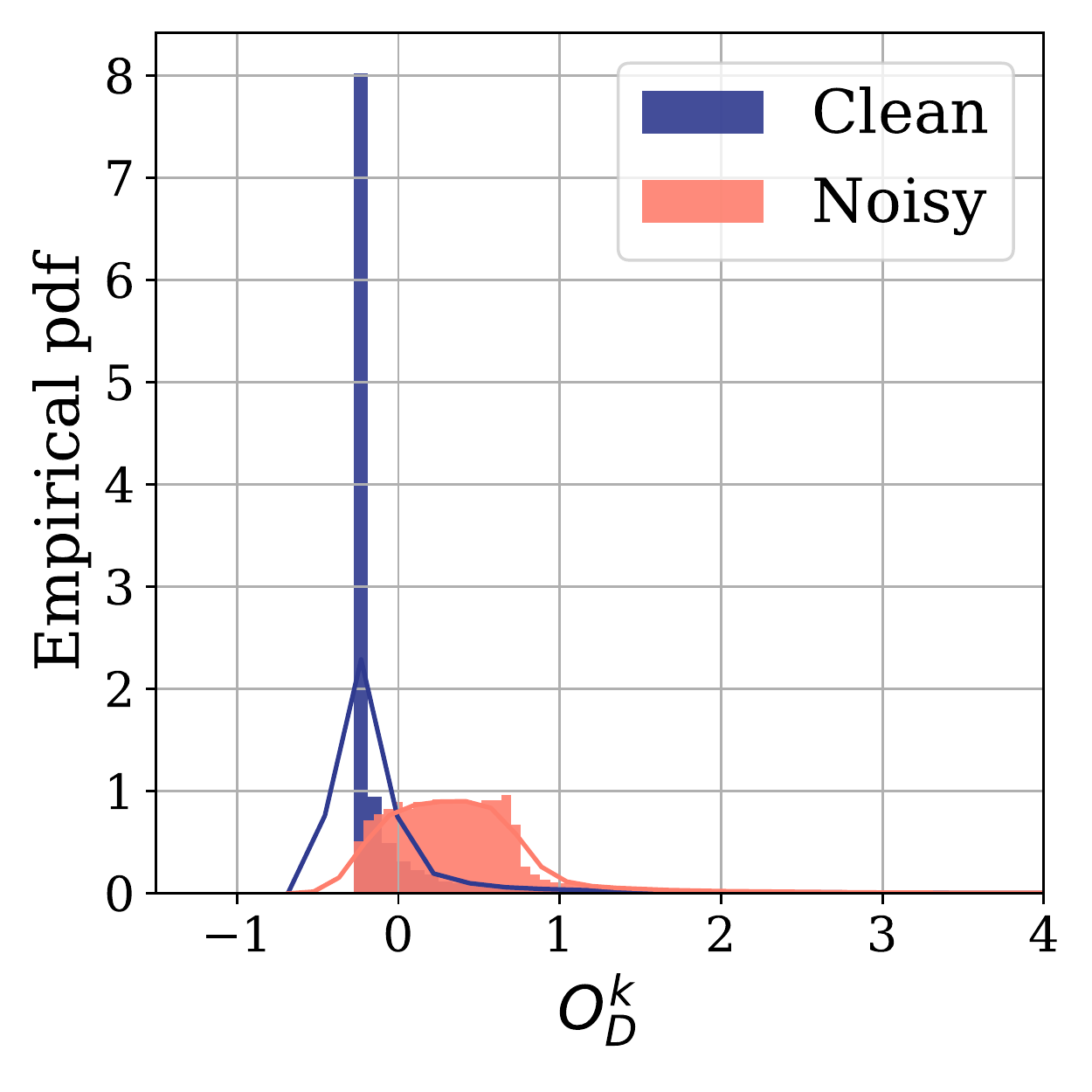}
}
 \subfigure[DivideMix]{
 \includegraphics[width=0.45\columnwidth]{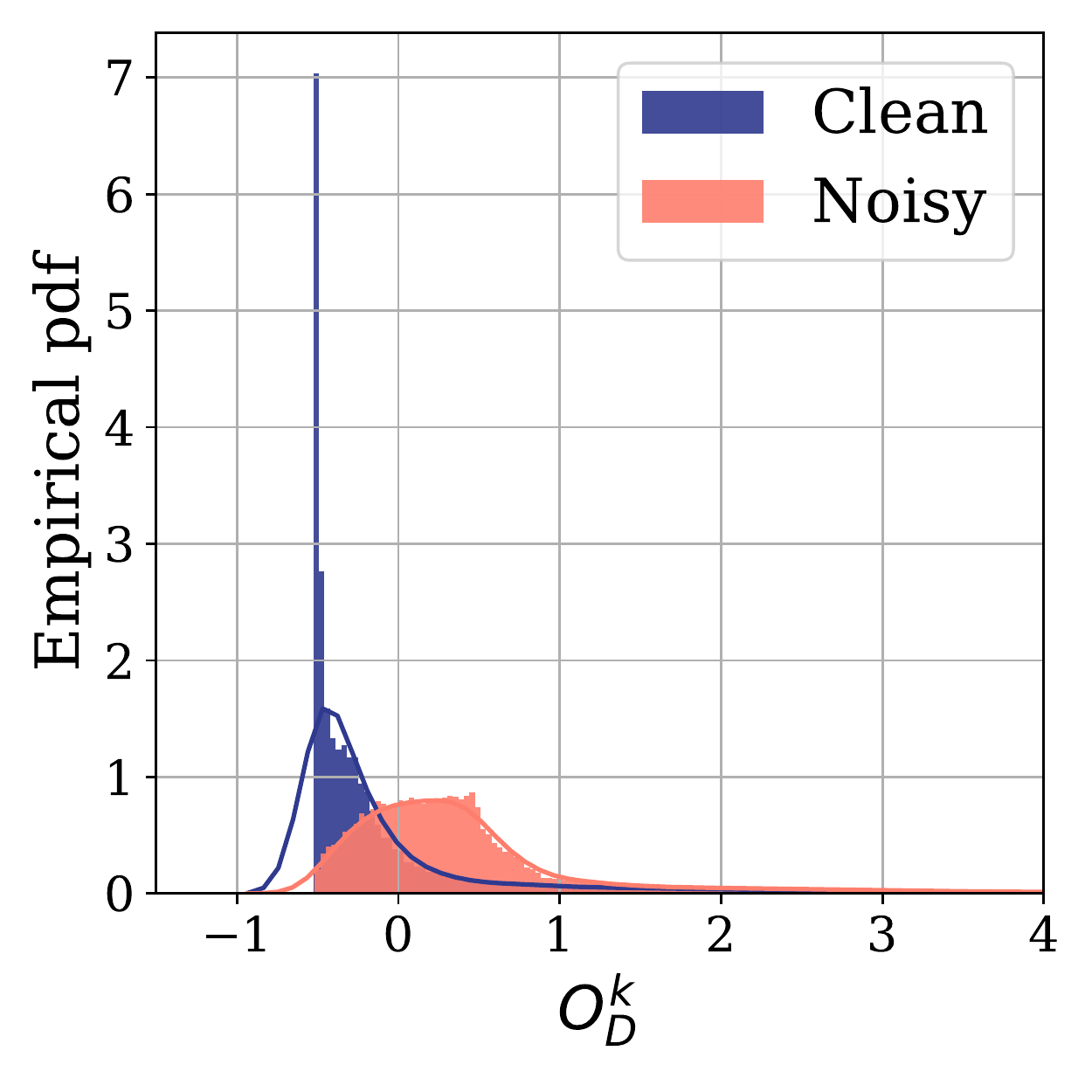}
}
 \end{center}
 \vspace*{-0.5cm}
%  \vspace*{-0.5cm}
%  \hspace*{1.1cm}  {\small (a) CE.} \hspace*{1.5cm} {\small (b) DivideMix.}
% % \vspace*{0.1cm}
 \caption{ \textbf{OSD distribution for all noisy training examples} after training CIFAR-10 with symmetric noise of $50\%$.
 }
 \label{appendix:fig:od_dist}
 \end{figure}

% \begin{figure}[t!]
% \centering{
% \includegraphics[width=1.0\columnwidth]{fig/examples.pdf}
% %\vspace*{-0.8cm}
% \caption{ \textbf{Mislabeled examples detected by \our from CIFAR-N\,(Worst) pre-trained by DivideMix.} The tag below figures denotes `true label' $\rightarrow$ `incorrect label' caused by human labeling error. All of the examples exhibit small losses.
% %During training, some samples have small losses and high confidence score that small-loss tricks (i.e., DivideMix) fail to detect them.
% %However, \our could detect these samples since they have high influence scores. Therefore, \our is complementary to the existing methods.
% }
% \label{fig:mislabeledexamples}
% }
% \end{figure}

\subsection{Effects of hyperparameter} \label{appendix:sec:gamma}
Choosing a high $\gamma$ increases the precision of the detected noisy label since it means that a training point exerts inconsistent influences to many classes (Figure~\ref{appendix:fig:gamma}).
On the other hand, to meet the high standard (\emph{e.g.,} unanimous consensus among all classes), it cannot but select less noisy samples, which results in the ratio of the remaining noisy labels to be high.
Therefore, choosing $\gamma$ is a tradeoff between the more accurate detection and the faster cleansing.
Therefore, in our experiments, the gamma is set to 5 in order to fix the data faster when the noise ratio is more than 40\%, and set to 8 in the other cases.

Furthermore, setting $\gamma = 0$ is equivalent to using only OSM in the algorithm.
Hence, it is verified that OSD helps to increase the precision of noisy label detection.

% \subsection{Effect of components}
% OSM alone, OSM+OSD 비교.

\subsection{Distribution of OSD} \label{appendix:sec:osd_dist}
To find noisy candidates, we fit a two-modality Gaussian mixture model (GMM) to ${\mathcal O_D^k} (x_i;\hat{\theta})$ for $k$-th class.
To justify if GMM can detect noisy candidates, we plot the distribution of training samples' ${\mathcal O_D^k} (x_i;\hat{\theta})$ (\emph{i.e.,} 6th class) in Figure~\ref{appendix:fig:od_dist}.
We calculate OSD from two models trained on CIFAR-10 (Symm-50) with CE and DivideMix, respectively. 
As shown in Figure~\ref{appendix:fig:od_dist}, OSD of clean and noisy samples is bi-modal and separable.
Thus, we fit the two-modality GMM into the OSD of all training examples to choose noisy candidates in the proposed algorithm.
This observation is consistent even when the model is trained with the existing robust methods. 
%In the heavy label noise, the model is prone to overfit to many examples with incorrect labels, which are misclassified as clean ones due to high error in sample selection.

\subsection{Noisy Label Detection with \our }\label{appendix:sec:noise_detection}

We report the noise ratio change after applying \our (2 rounds) on CIFAR and CIFAR-N in Table~\ref{appendix:table:cifar-noise}.
As can be seen from the tables, the original noise ratio has been largely alleviated after applying \our.
As can be seen from Section ~\ref{appendix:sec:exp_multi}, applying \our for more rounds can further alleviate the noise ratio in datasets.

Furthermore, we present the noisy label detection precision in Table~\ref{appendix:table:cifar-precision}.
We can observe that mislabeled samples are detected with high precision on both symmetric and real-world noisy data.

%The alleviated noise ratio along with the improved accuracy in Table~\ref{table:cifar-symm} and \ref{table:cifar-n} prove that \our is effective in detecting mislabeled samples.

\begin{table*}[t]
\vspace{0.8cm}
\caption{\textbf{Averaged noise ratio (\%) after \our (2 rounds).}}
\label{appendix:table:cifar-noise}
\hspace*{3cm}  {\small (a) CIFAR with symmetric noise.} \hspace*{5.7cm} {\small (b) CIFAR-N.}

\begin{minipage}{1.2\columnwidth}
% \caption{
% \textbf{Averaged noise Ratio (\%) after \our (2 rounds) on CIFAR with symmetric noise.}
% }
%\vspace{0cm}
\centering
\footnotesize
\resizebox{1\columnwidth}{!}
{
\begin{tabular}{lccc|ccc}
\toprule
                  & \multicolumn{3}{c|}{CIFAR-10} & \multicolumn{3}{c}{CIFAR-100} \\ \hline
                  & Symm-20  & Symm-50  & Symm-70 & Symm-20  & Symm-50  & Symm-70 \\ \midrule
No Post-processing & 20       & 50       & 70      & 20       & 50       & 70      \\
CE                 & 8.83     & 37.01     & 62.71    & 12.15     & 45.11     & 64.67    \\
VolMinNet~\cite{ref:volminnet_2021_icml} & 4.42     & 36.68     & 61.02    & 5.53      & 31.46     & 55.15    \\
Co-teaching~\cite{ref:coteacing_nips_18} & 3.99     & 35.26     & 64.95    & 6.63     & 31.81     & 59.40    \\
ELR~\cite{ref:elr_neurips_2020}     & 5.51      & 39.99     & 63.73    & 6.52      & 30.34     & 60.48    \\
DivideMix~\cite{ref:dividemix_iclr_2019} & 3.84      & 22.79     & 42.11    & 7.18      & 26.53     & 49.81    \\
UNICON~\cite{ref:unicon_2022} & 4.34      & 29.36     & 54.72    & 5.26      & 30.43     & 55.31    \\ \bottomrule
\end{tabular}
}
%\label{appendix:table:cifar-noise}
\end{minipage} \hfill
\begin{minipage}{0.89\columnwidth}
\centering
\footnotesize
\resizebox{1\columnwidth}{!}
{
\begin{tabular}{lccc|c}
\toprule
                  & \multicolumn{3}{c|}{CIFAR-10N} & CIFAR-100N \\ \hline
                  & Aggregate  & Random1  & Worst  & Noisy      \\ \midrule
No Post-processing & 9          & 18       & 40     & 40         \\
CE                 & 5.30          & 11.07       & 29.79     & 34.20       \\
VolMinNet~\cite{ref:volminnet_2021_icml}          & 2.48        & 4.54     & 38.61   & 31.24       \\
Co-teaching~\cite{ref:coteacing_nips_18}        & 2.32        & 4.33      & 26.55   & 28.56       \\
ELR~\cite{ref:elr_neurips_2020}                & 1.05        & 3.26      & 26.51   & 27.67       \\
DivideMix~\cite{ref:dividemix_iclr_2019}          & 1.11        & 2.98      & 18.32   & 27.28       \\ 
UNICON~\cite{ref:unicon_2022}          & 1.55        & 6.52      & 23.63   & 25.23       \\ \bottomrule
\end{tabular}
}
%\captionof{table}{CIFAR-N.}
%\label{appendix:table:cifarN-noise}
%\end{table*}
\end{minipage}
%\vspace*{0.9cm}
% \hspace*{3cm}  {\small (a) CIFAR with symmetric noise.} \hspace*{5.7cm} {\small (b) CIFAR-N.}
% \vspace*{-0.1cm}
%\caption{\textbf{Averaged noise ratio (\%) after \our (2 rounds).}}
%\label{appendix:table:cifar-noise}
% \vspace*{-0.1cm}
%\vspace*{-0.2cm}
\end{table*}

\begin{table*}[h]
% \vspace*{-0.1cm}
\caption{\textbf{Averaged precision (\%) of noise detection after \our (2 rounds).}}
\label{appendix:table:cifar-precision}
\hspace*{3cm}  {\small (a) CIFAR with symmetric noise.} \hspace*{5.7cm} {\small (b) CIFAR-N.}

\begin{minipage}{1.2\columnwidth}
% \caption{
% \textbf{Averaged noise Ratio (\%) after \our (2 rounds) on CIFAR with symmetric noise.}
% }
%\vspace{0cm}
\centering
\footnotesize
\resizebox{1\columnwidth}{!}
{
\begin{tabular}{lccc|ccc}
\toprule
                  & \multicolumn{3}{c|}{CIFAR-10} & \multicolumn{3}{c}{CIFAR-100} \\ \hline
                  & Symm-20  & Symm-50  & Symm-70 & Symm-20  & Symm-50  & Symm-70 \\ \midrule
CE                 & 82.84     & 92.23     & 93.36    & 62.38     & 73.35     & 86.34    \\
VolMinNet~\cite{ref:volminnet_2021_icml} & 91.56     & 99.79     & 94.01    & 94.80      & 98.63     & 97.13    \\
Co-teaching~\cite{ref:coteacing_nips_18} & 96.41     & 99.90     & 88.82    & 94.37     & 98.52     & 92.42    \\
ELR~\cite{ref:elr_neurips_2020}     & 96.37      & 99.55     & 99.58    & 83.14      & 92.48     & 92.76    \\
DivideMix~\cite{ref:dividemix_iclr_2019} & 93.91      & 96.97     & 98.69    & 74.56      & 91.95     & 96.06   \\
UNICON~\cite{ref:unicon_2022} & 86.88      & 97.96     & 99.44    & 85.35      & 97.47     & 97.64   \\ \bottomrule
\end{tabular}
}
%\label{appendix:table:cifar-noise}
\end{minipage} \hfill
\begin{minipage}{0.89\columnwidth}
\centering
\footnotesize
\resizebox{1\columnwidth}{!}
{
\begin{tabular}{lccc|c}
\toprule
                  & \multicolumn{3}{c|}{CIFAR-10N} & CIFAR-100N \\ \hline
                  & Aggregate  & Random1  & Worst  & Noisy      \\ \midrule
CE                 & 59.48          & 85.07       & 90.19     & 74.32       \\
VolMinNet~\cite{ref:volminnet_2021_icml}          & 61.35        & 74.33     & 98.96   & 89.38       \\
Co-teaching~\cite{ref:coteacing_nips_18}        & 64.46        & 88.49      & 98.89   & 89.71       \\
ELR~\cite{ref:elr_neurips_2020}                & 65.31        & 91.48      & 98.92   & 89.29       \\
DivideMix~\cite{ref:dividemix_iclr_2019}          & 70.19        & 90.68      & 95.51   & 86.55       \\ 
UNICON~\cite{ref:unicon_2022}          & 54.77        & 80.11      & 96.51   & 85.54       \\ \bottomrule
\end{tabular}
}
%\captionof{table}{CIFAR-N.}
%\label{appendix:table:cifarN-noise}
%\end{table*}
\end{minipage}
\vspace*{0.2cm}
\end{table*}

\begin{table*}[!h]
\caption{
\textbf{Comparison on CIFAR with varying levels of label noises (1 round).}
The averaged test accuracy\,(\%) with LNL methods and their combination with \our. 
The mean accuracy is computed over three different noise realizations.
%where the values in parentheses are the relative performance gain (\%) compared with the pre-trained methods. 
%The best results are marked in bold.
}
% \vspace{-0.2cm}
\label{table:cifar-symm-oneround}
\centering
%\small
\footnotesize
\resizebox{1\textwidth}{!}
{
\begin{tabular}{l|cccccc||cccccc}
\toprule
\multirow{3}{*}{Method} & \multicolumn{6}{c||}{CIFAR-10}                                                                                                                                                                                                                                                                                                                                                                      & \multicolumn{6}{c}{CIFAR-100}                                                                                                                                                                                                                                                                                                                                                                    \\ \cline{2-13} 
                        & \multicolumn{2}{c|}{Symm-20}                                                                                                          & \multicolumn{2}{c|}{Symm-50}                                                                                                           & \multicolumn{2}{c||}{Symm-70}                                                                                      & \multicolumn{2}{c|}{Symm-20}                                                                                                          & \multicolumn{2}{c|}{Symm-50}                                                                                                          & \multicolumn{2}{c}{Symm-70}                                                                                      \\ \cline{2-13} 
                        & Original                                               & \multicolumn{1}{c|}{+Inf. Rank}                                              & Original                                               & \multicolumn{1}{c|}{+Inf. Rank}                                               & Original                                               & +Inf. Rank                                               & Original                                               & \multicolumn{1}{c|}{+Inf. Rank}                                              & Original                                               & \multicolumn{1}{c|}{+Inf. Rank}                                              & Original                                               & +Inf. Rank                                              \\ \midrule
CE                      & \begin{tabular}[c]{@{}c@{}}80.46\\ (+0.0)\end{tabular} & \multicolumn{1}{c|}{\begin{tabular}[c]{@{}c@{}}\textbf{87.46}\\ \textbf{(+7.00)}\end{tabular}} & \begin{tabular}[c]{@{}c@{}}48.84\\ (+0.0)\end{tabular} & \multicolumn{1}{c|}{\begin{tabular}[c]{@{}c@{}}\textbf{78.14}\\ \textbf{(+29.31)}\end{tabular}} & \begin{tabular}[c]{@{}c@{}}28.42\\ (+0.0)\end{tabular} & \begin{tabular}[c]{@{}c@{}}\textbf{65.33}\\ \textbf{(+36.91)}\end{tabular} & \begin{tabular}[c]{@{}c@{}}64.35\\ (+0.0)\end{tabular} & \multicolumn{1}{c|}{\begin{tabular}[c]{@{}c@{}}\textbf{67.20}\\ \textbf{(+2.85)}\end{tabular}} & \begin{tabular}[c]{@{}c@{}}39.43\\ (+0.0)\end{tabular} & \multicolumn{1}{c|}{\begin{tabular}[c]{@{}c@{}}\textbf{47.36}\\ \textbf{(+7.93)}\end{tabular}} & \begin{tabular}[c]{@{}c@{}}15.50\\ (+0.0)\end{tabular} & \begin{tabular}[c]{@{}c@{}}\textbf{25.26}\\ \textbf{(+9.76)}\end{tabular} \\ \hline
VolMinNet~\cite{ref:volminnet_2021_icml}               & \begin{tabular}[c]{@{}c@{}}88.26\\ (+0.0)\end{tabular} & \multicolumn{1}{c|}{\begin{tabular}[c]{@{}c@{}}\textbf{90.90}\\ \textbf{(+2.64)}\end{tabular}} & \begin{tabular}[c]{@{}c@{}}71.13\\ (+0.0)\end{tabular} & \multicolumn{1}{c|}{\begin{tabular}[c]{@{}c@{}}\textbf{82.05}\\ \textbf{(+10.92)}\end{tabular}} & \begin{tabular}[c]{@{}c@{}}33.69\\ (+0.0)\end{tabular} & \begin{tabular}[c]{@{}c@{}}\textbf{63.50}\\ \textbf{(+29.82)}\end{tabular} & \begin{tabular}[c]{@{}c@{}}65.11\\ (+0.0)\end{tabular} & \multicolumn{1}{c|}{\begin{tabular}[c]{@{}c@{}}\textbf{68.48}\\ \textbf{(+3.37)}\end{tabular}} & \begin{tabular}[c]{@{}c@{}}48.77\\ (+0.0)\end{tabular} & \multicolumn{1}{c|}{\begin{tabular}[c]{@{}c@{}}\textbf{56.15}\\ \textbf{(+7.38)}\end{tabular}} & \begin{tabular}[c]{@{}c@{}}28.64\\ (+0.0)\end{tabular} & \begin{tabular}[c]{@{}c@{}}\textbf{36.86}\\ \textbf{(+8.22)}\end{tabular} \\ \hline
Co-teaching~\cite{ref:coteacing_nips_18}             & \begin{tabular}[c]{@{}c@{}}91.85\\ (+0.0)\end{tabular} & \multicolumn{1}{c|}{\begin{tabular}[c]{@{}c@{}}\textbf{92.77}\\ \textbf{(+0.92)}\end{tabular}} & \begin{tabular}[c]{@{}c@{}}85.44\\ (+0.0)\end{tabular} & \multicolumn{1}{c|}{\begin{tabular}[c]{@{}c@{}}\textbf{87.04}\\ \textbf{(+1.61)}\end{tabular}}  & \begin{tabular}[c]{@{}c@{}}52.63\\ (+0.0)\end{tabular} & \begin{tabular}[c]{@{}c@{}}\textbf{56.92}\\ \textbf{(+4.30)}\end{tabular}  & \begin{tabular}[c]{@{}c@{}}70.85\\ (+0.0)\end{tabular} & \multicolumn{1}{c|}{\begin{tabular}[c]{@{}c@{}}\textbf{71.42}\\ \textbf{(+0.56)}\end{tabular}} & \begin{tabular}[c]{@{}c@{}}59.14\\ (+0.0)\end{tabular} & \multicolumn{1}{c|}{\begin{tabular}[c]{@{}c@{}}\textbf{61.01}\\ \textbf{(+1.87)}\end{tabular}} & \begin{tabular}[c]{@{}c@{}}35.78\\ (+0.0)\end{tabular} & \begin{tabular}[c]{@{}c@{}}\textbf{37.56}\\ \textbf{(+2.16)}\end{tabular} \\ \hline
ELR~\cite{ref:elr_neurips_2020}                     & \begin{tabular}[c]{@{}c@{}}91.88\\ (+0.0)\end{tabular} & \multicolumn{1}{c|}{\begin{tabular}[c]{@{}c@{}}\textbf{92.52}\\ \textbf{(+0.64)}\end{tabular}} & \begin{tabular}[c]{@{}c@{}}88.48\\ (+0.0)\end{tabular} & \multicolumn{1}{c|}{\begin{tabular}[c]{@{}c@{}}\textbf{89.13}\\ \textbf{(+0.65)}\end{tabular}}  & \begin{tabular}[c]{@{}c@{}}77.26\\ (+0.0)\end{tabular} & \begin{tabular}[c]{@{}c@{}}\textbf{79.20}\\ \textbf{(+1.94)}\end{tabular}  & \begin{tabular}[c]{@{}c@{}}72.58\\ (+0.0)\end{tabular} & \multicolumn{1}{c|}{\begin{tabular}[c]{@{}c@{}}\textbf{73.41}\\ \textbf{(+0.83)}\end{tabular}} & \begin{tabular}[c]{@{}c@{}}64.01\\ (+0.0)\end{tabular} & \multicolumn{1}{c|}{\begin{tabular}[c]{@{}c@{}}\textbf{64.36}\\ \textbf{(+0.36)}\end{tabular}} & \begin{tabular}[c]{@{}c@{}}38.78\\ (+0.0)\end{tabular} & \begin{tabular}[c]{@{}c@{}}\textbf{38.89}\\ \textbf{(+0.11)}\end{tabular} \\ \hline
ELR+~\cite{ref:elr_neurips_2020}                    & \begin{tabular}[c]{@{}c@{}}93.75\\ (+0.0)\end{tabular} & \multicolumn{1}{c|}{\begin{tabular}[c]{@{}c@{}}\textbf{94.07}\\ \textbf{(+0.32)}\end{tabular}} & \begin{tabular}[c]{@{}c@{}}92.05\\ (+0.0)\end{tabular} & \multicolumn{1}{c|}{\begin{tabular}[c]{@{}c@{}}\textbf{92.40}\\ \textbf{(+0.35)}\end{tabular}}  & \begin{tabular}[c]{@{}c@{}}86.94\\ (+0.0)\end{tabular} & \begin{tabular}[c]{@{}c@{}}\textbf{87.56}\\ \textbf{(+0.62)}\end{tabular}  & \begin{tabular}[c]{@{}c@{}}74.15\\ (+0.0)\end{tabular} & \multicolumn{1}{c|}{\begin{tabular}[c]{@{}c@{}}\textbf{74.93}\\ \textbf{(+0.78)}\end{tabular}} & \begin{tabular}[c]{@{}c@{}}65.66\\ (+0.0)\end{tabular} & \multicolumn{1}{c|}{\begin{tabular}[c]{@{}c@{}}\textbf{68.52}\\ \textbf{(+2.86)}\end{tabular}} & \begin{tabular}[c]{@{}c@{}}50.19\\ (+0.0)\end{tabular} & \begin{tabular}[c]{@{}c@{}}\textbf{52.55}\\ \textbf{(+2.36)}\end{tabular} \\ \hline
DivideMix~\cite{ref:dividemix_iclr_2019}               & \begin{tabular}[c]{@{}c@{}}95.64\\ (+0.0)\end{tabular} & \multicolumn{1}{c|}{\begin{tabular}[c]{@{}c@{}}\textbf{95.96}\\ \textbf{(+0.32)}\end{tabular}} & \begin{tabular}[c]{@{}c@{}}94.02\\ (+0.0)\end{tabular} & \multicolumn{1}{c|}{\begin{tabular}[c]{@{}c@{}}\textbf{94.61}\\ \textbf{(+0.59)}\end{tabular}}  & \begin{tabular}[c]{@{}c@{}}91.27\\ (+0.0)\end{tabular} & \begin{tabular}[c]{@{}c@{}}\textbf{93.28}\\ \textbf{(+2.01)}\end{tabular}  & \begin{tabular}[c]{@{}c@{}}76.57\\ (+0.0)\end{tabular} & \multicolumn{1}{c|}{\begin{tabular}[c]{@{}c@{}}\textbf{77.83}\\ \textbf{(+1.25)}\end{tabular}} & \begin{tabular}[c]{@{}c@{}}72.29\\ (+0.0)\end{tabular} & \multicolumn{1}{c|}{\begin{tabular}[c]{@{}c@{}}\textbf{73.49}\\ \textbf{(+1.20)}\end{tabular}} & \begin{tabular}[c]{@{}c@{}}62.43\\ (+0.0)\end{tabular} & \begin{tabular}[c]{@{}c@{}}\textbf{64.43}\\ \textbf{(+2.00)}\end{tabular} \\ \hline
UNICON~\cite{ref:unicon_2022}               & \begin{tabular}[c]{@{}c@{}}91.95\\ (+0.0)\end{tabular} & \multicolumn{1}{c|}{\begin{tabular}[c]{@{}c@{}}\textbf{94.52}\\ \textbf{(+2.56)}\end{tabular}} & \begin{tabular}[c]{@{}c@{}}93.59\\ (+0.0)\end{tabular} & \multicolumn{1}{c|}{\begin{tabular}[c]{@{}c@{}}\textbf{94.75}\\ \textbf{(+1.16)}\end{tabular}}  & \begin{tabular}[c]{@{}c@{}}91.44\\ (+0.0)\end{tabular} & \begin{tabular}[c]{@{}c@{}}\textbf{92.84}\\ \textbf{(+1.40)}\end{tabular}  & \begin{tabular}[c]{@{}c@{}}74.82\\ (+0.0)\end{tabular} & \multicolumn{1}{c|}{\begin{tabular}[c]{@{}c@{}}\textbf{79.22}\\ \textbf{(+4.40)}\end{tabular}} & \begin{tabular}[c]{@{}c@{}}73.96\\ (+0.0)\end{tabular} & \multicolumn{1}{c|}{\begin{tabular}[c]{@{}c@{}}\textbf{75.36}\\ \textbf{(+1.40)}\end{tabular}} & \begin{tabular}[c]{@{}c@{}}68.61\\ (+0.0)\end{tabular} & \begin{tabular}[c]{@{}c@{}}\textbf{69.63}\\ \textbf{(+1.02)}\end{tabular} \\ \bottomrule
\end{tabular}
}
%\vspace{-0.1cm}
% \vspace{0.5cm}
\end{table*}

\begin{table*}[!h]
\caption{
\textbf{Comparison on CIFAR-N with varying levels of real-world noises (1 round).}
The averaged test accuracy\,(\%) with robust methods and their combination with RoG and \our{}. 
The mean accuracy is computed over three different noise realizations.
}
% \vspace{-0.2cm}
\label{table:cifar-n-oneround}
\centering
\footnotesize
%\small
\resizebox{0.7\textwidth}{!}
{
\begin{tabular}{l|cccccc||cc}
\toprule
\multirow{3}{*}{Method} & \multicolumn{6}{c||}{CIFAR-10N}                                                                                                                                                                                                                                                                                                                                                                    & \multicolumn{2}{c}{CIFAR-100N}                                                                                   \\ \cline{2-9} 
                        & \multicolumn{2}{c|}{Aggre}                                                                                                            & \multicolumn{2}{c|}{Rand1}                                                                                                            & \multicolumn{2}{c||}{Worst}                                                                                        & \multicolumn{2}{c}{Noisy}                                                                                        \\ \cline{2-9} 
                        & Original                                               & \multicolumn{1}{c|}{+Inf. Rank}                                              & Original                                               & \multicolumn{1}{c|}{+Inf. Rank}                                              & Original                                               & +Inf. Rank                                               & Original                                               & +Inf. Rank                                              \\ \midrule
CE                      & \begin{tabular}[c]{@{}c@{}}89.81\\ (+0.0)\end{tabular} & \multicolumn{1}{c|}{\begin{tabular}[c]{@{}c@{}}\textbf{90.79}\\ \textbf{(+0.98)}\end{tabular}} & \begin{tabular}[c]{@{}c@{}}83.80\\ (+0.0)\end{tabular} & \multicolumn{1}{c|}{\begin{tabular}[c]{@{}c@{}}\textbf{87.98}\\ \textbf{(+4.18)}\end{tabular}} & \begin{tabular}[c]{@{}c@{}}64.86\\ (+0.0)\end{tabular} & \begin{tabular}[c]{@{}c@{}}\textbf{78.56}\\ \textbf{(+13.70)}\end{tabular} & \begin{tabular}[c]{@{}c@{}}54.71\\ (+0.0)\end{tabular} & \begin{tabular}[c]{@{}c@{}}\textbf{59.77}\\ \textbf{(+5.06)}\end{tabular} \\ \hline
VolMinNet~\cite{ref:volminnet_2021_icml}               & \begin{tabular}[c]{@{}c@{}}88.59\\ (+0.0)\end{tabular} & \multicolumn{1}{c|}{\begin{tabular}[c]{@{}c@{}}\textbf{90.72}\\ \textbf{(+2.14)}\end{tabular}} & \begin{tabular}[c]{@{}c@{}}85.37\\ (+0.0)\end{tabular} & \multicolumn{1}{c|}{\begin{tabular}[c]{@{}c@{}}\textbf{88.95}\\ \textbf{(+3.58)}\end{tabular}} & \begin{tabular}[c]{@{}c@{}}72.35\\ (+0.0)\end{tabular} & \begin{tabular}[c]{@{}c@{}}\textbf{78.97}\\ \textbf{(+6.63)}\end{tabular}  & \begin{tabular}[c]{@{}c@{}}54.32\\ (+0.0)\end{tabular} & \begin{tabular}[c]{@{}c@{}}\textbf{56.94}\\ \textbf{(+4.36)}\end{tabular} \\ \hline
Co-teaching~\cite{ref:coteacing_nips_18}             & \begin{tabular}[c]{@{}c@{}}92.79\\ (+0.0)\end{tabular} & \multicolumn{1}{c|}{\begin{tabular}[c]{@{}c@{}}\textbf{93.28}\\ \textbf{(+0.49)}\end{tabular}} & \begin{tabular}[c]{@{}c@{}}91.59\\ (+0.0)\end{tabular} & \multicolumn{1}{c|}{\begin{tabular}[c]{@{}c@{}}\textbf{92.13}\\ \textbf{(+0.54)}\end{tabular}} & \begin{tabular}[c]{@{}c@{}}84.30\\ (+0.0)\end{tabular} & \begin{tabular}[c]{@{}c@{}}\textbf{86.03}\\ \textbf{(+1.72)}\end{tabular}  & \begin{tabular}[c]{@{}c@{}}61.07\\ (+0.0)\end{tabular} & \begin{tabular}[c]{@{}c@{}}\textbf{62.36}\\ \textbf{(+1.29)}\end{tabular} \\ \hline
ELR~\cite{ref:elr_neurips_2020}                     & \begin{tabular}[c]{@{}c@{}}92.09\\ (+0.0)\end{tabular} & \multicolumn{1}{c|}{\begin{tabular}[c]{@{}c@{}}\textbf{92.78}\\ \textbf{(+0.69)}\end{tabular}} & \begin{tabular}[c]{@{}c@{}}91.59\\ (+0.0)\end{tabular} & \multicolumn{1}{c|}{\begin{tabular}[c]{@{}c@{}}\textbf{92.09}\\ \textbf{(+0.50)}\end{tabular}} & \begin{tabular}[c]{@{}c@{}}86.07\\ (+0.0)\end{tabular} & \begin{tabular}[c]{@{}c@{}}\textbf{87.21}\\ \textbf{(+1.14)}\end{tabular}  & \begin{tabular}[c]{@{}c@{}}62.72\\ (+0.0)\end{tabular} & \begin{tabular}[c]{@{}c@{}}\textbf{64.02}\\ \textbf{(+1.31)}\end{tabular} \\ \hline
ELR+~\cite{ref:elr_neurips_2020}                   & \begin{tabular}[c]{@{}c@{}}94.36\\ (+0.0)\end{tabular} & \multicolumn{1}{c|}{\begin{tabular}[c]{@{}c@{}}\textbf{94.40}\\ \textbf{(+0.04)}\end{tabular}} & \begin{tabular}[c]{@{}c@{}}93.60\\ (+0.0)\end{tabular} & \multicolumn{1}{c|}{\begin{tabular}[c]{@{}c@{}}\textbf{93.85}\\ \textbf{(+0.25)}\end{tabular}} & \begin{tabular}[c]{@{}c@{}}89.74\\ (+0.0)\end{tabular} & \begin{tabular}[c]{@{}c@{}}\textbf{90.39}\\ \textbf{(+0.65)}\end{tabular}  & \begin{tabular}[c]{@{}c@{}}63.20\\ (+0.0)\end{tabular} & \begin{tabular}[c]{@{}c@{}}\textbf{64.28}\\ \textbf{(+1.07)}\end{tabular} \\ \hline
DivideMix~\cite{ref:dividemix_iclr_2019}               & \begin{tabular}[c]{@{}c@{}}94.99\\ (+0.0)\end{tabular} & \multicolumn{1}{c|}{\begin{tabular}[c]{@{}c@{}}\textbf{95.35}\\ \textbf{(+0.35)}\end{tabular}} & \begin{tabular}[c]{@{}c@{}}94.90\\ (+0.0)\end{tabular} & \multicolumn{1}{c|}{\begin{tabular}[c]{@{}c@{}}\textbf{95.37}\\ \textbf{(+0.47)}\end{tabular}} & \begin{tabular}[c]{@{}c@{}}92.24\\ (+0.0)\end{tabular} & \begin{tabular}[c]{@{}c@{}}\textbf{93.24}\\ \textbf{(+1.00)}\end{tabular}  & \begin{tabular}[c]{@{}c@{}}69.29\\ (+0.0)\end{tabular} & \begin{tabular}[c]{@{}c@{}}\textbf{70.67}\\ \textbf{(+1.38)}\end{tabular} \\ \hline
UNICON~\cite{ref:unicon_2022}               & \begin{tabular}[c]{@{}c@{}}90.82\\ (+0.0)\end{tabular} & \multicolumn{1}{c|}{\begin{tabular}[c]{@{}c@{}}\textbf{93.49}\\ \textbf{(+2.67)}\end{tabular}} & \begin{tabular}[c]{@{}c@{}}91.87\\ (+0.0)\end{tabular} & \multicolumn{1}{c|}{\begin{tabular}[c]{@{}c@{}}\textbf{93.96}\\ \textbf{(+2.09)}\end{tabular}} & \begin{tabular}[c]{@{}c@{}}92.33\\ (+0.0)\end{tabular} & \begin{tabular}[c]{@{}c@{}}\textbf{93.79}\\ \textbf{(+1.46)}\end{tabular}  & \begin{tabular}[c]{@{}c@{}}68.33\\ (+0.0)\end{tabular} & \begin{tabular}[c]{@{}c@{}}\textbf{70.68}\\ \textbf{(+2.35)}\end{tabular} \\ \bottomrule
\end{tabular}
}
%\vspace{-0.1cm}
% \vspace{0.5cm}
\end{table*}

\begin{figure*}[t]
\centering
\subfigure[labeled as `Jump'.]{
    \includegraphics[width=0.31\linewidth]{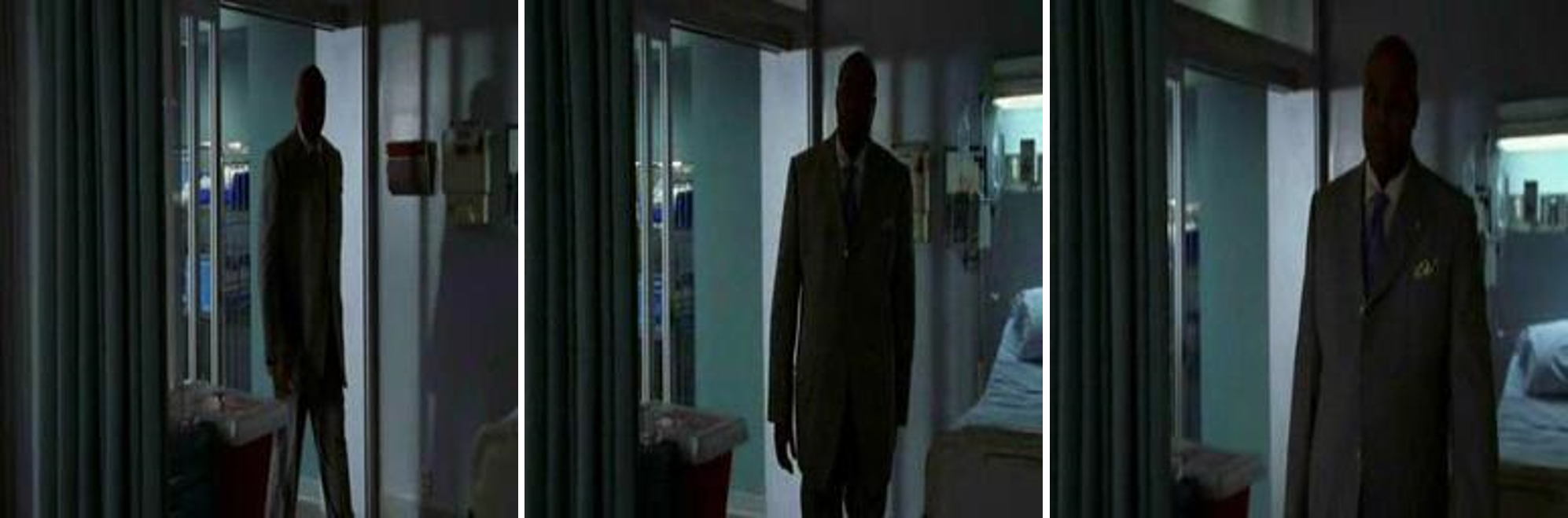}
}
\subfigure[labeled as `Dive'.]{
    \includegraphics[width=0.31\linewidth]{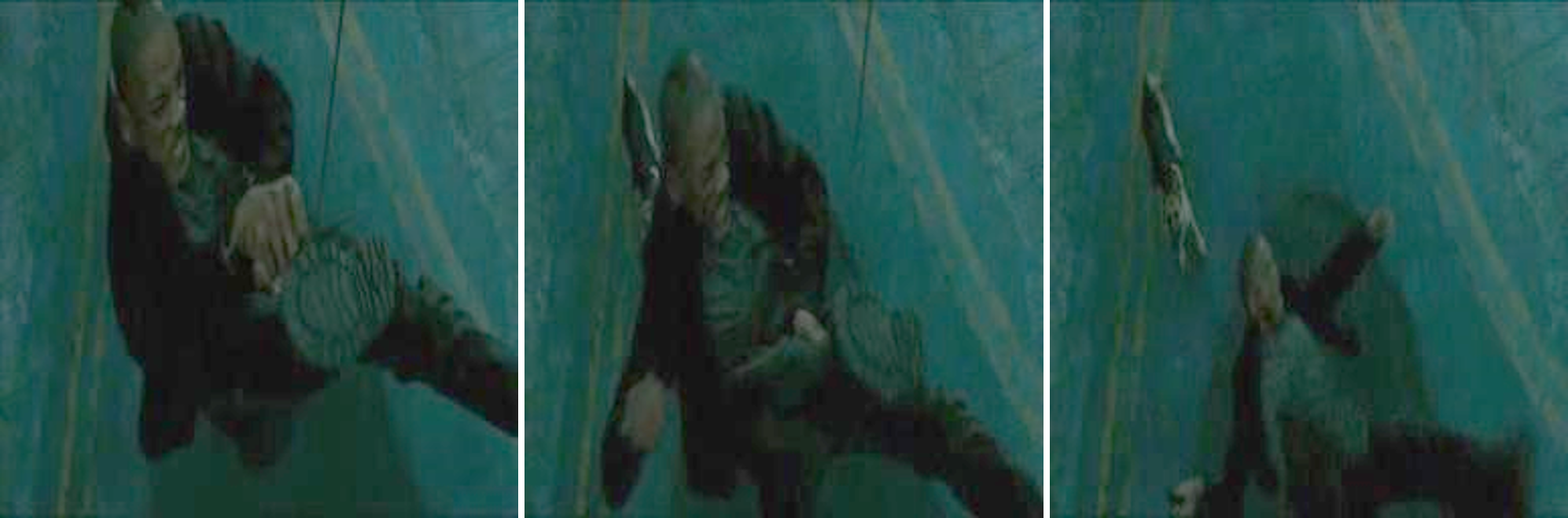}
}
\subfigure[labeled as `Hit'.]{
    \includegraphics[width=0.31\linewidth]{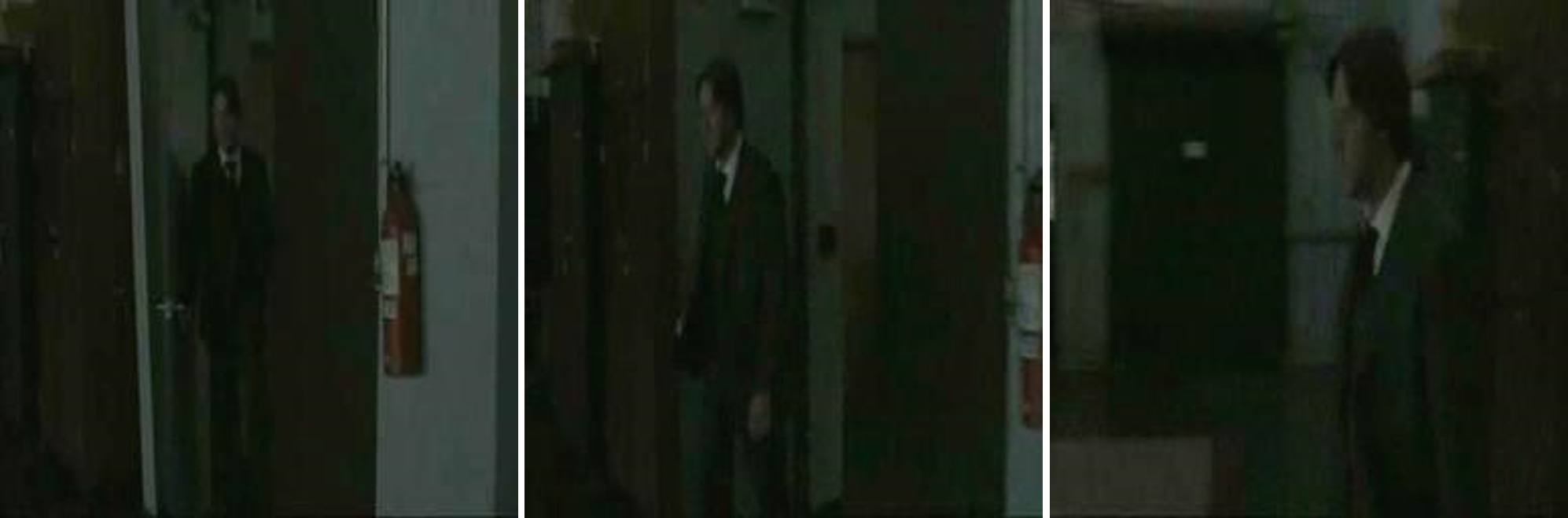}
}
\subfigure[labeled as `Dive'. (Jump?)]{
    \includegraphics[width=0.31\linewidth]{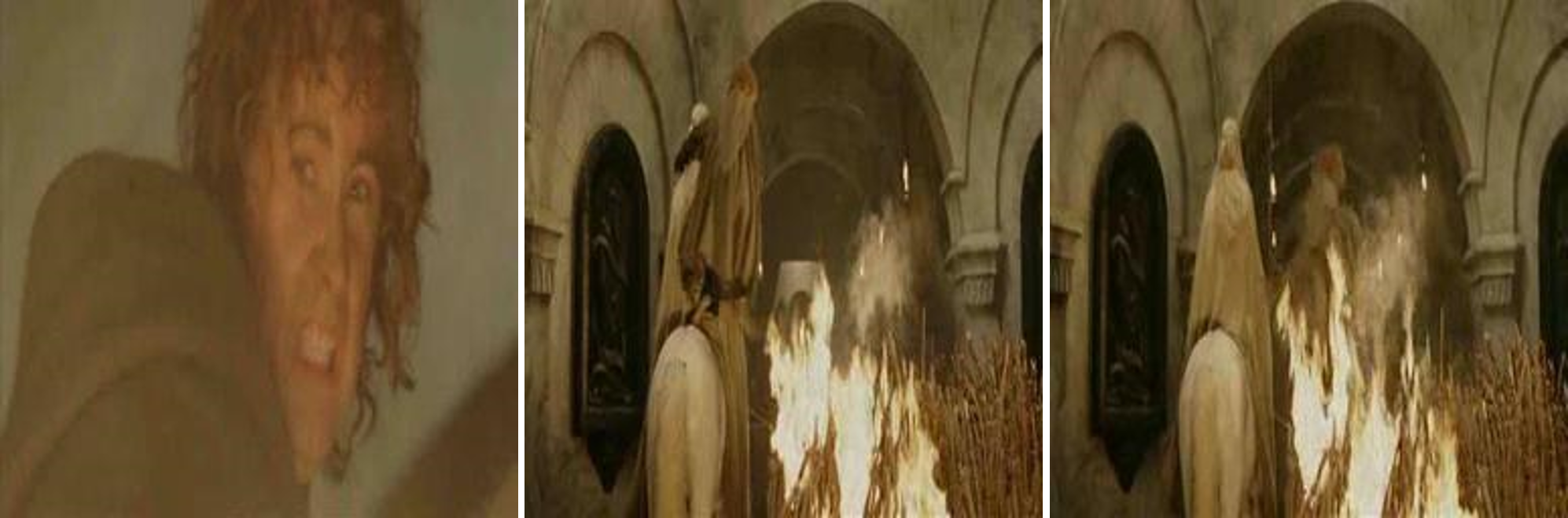}
}
\subfigure[labeled as `Punch'. (Kick?)]{
    \includegraphics[width=0.31\linewidth]{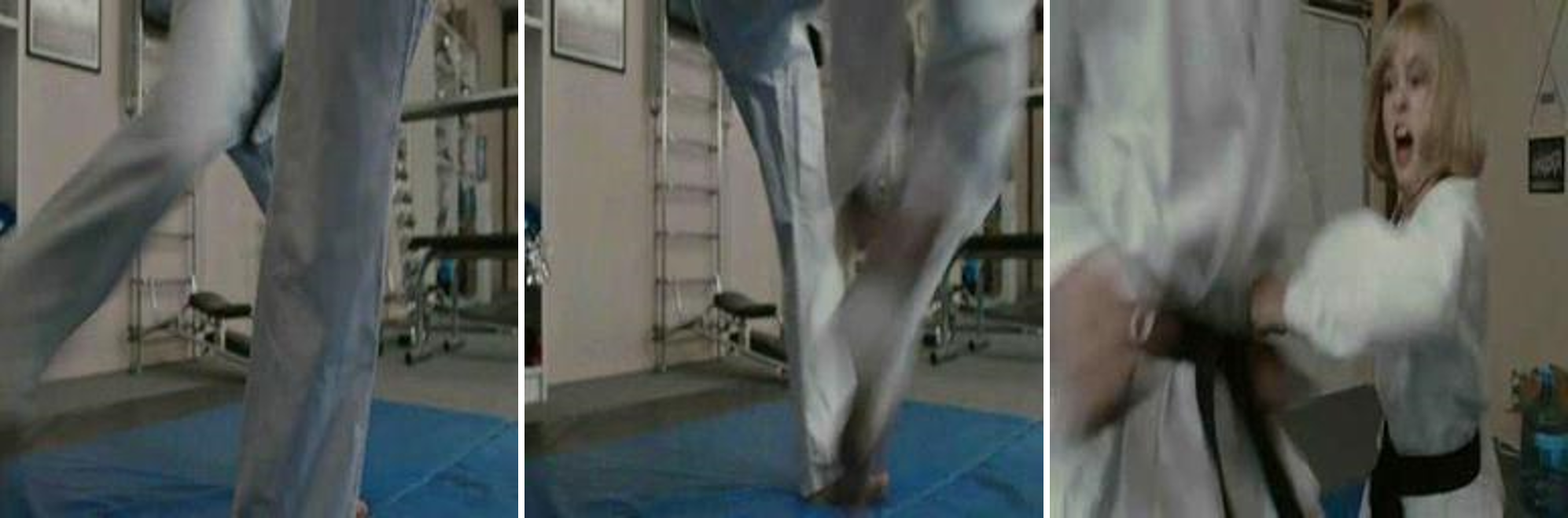}
}
\subfigure[labeled as `Punch'. (Hit? Sword?)]{
    \includegraphics[width=0.31\linewidth]{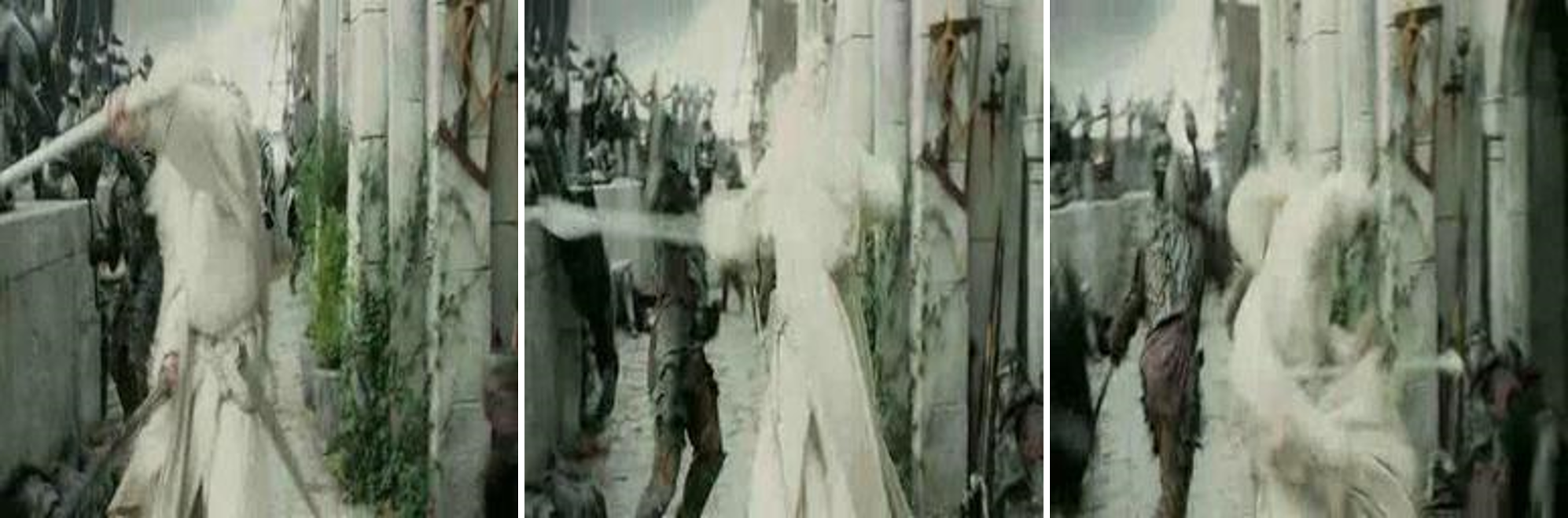}
}
\subfigure[labeled as `Kick'. (Run?)]{
    \includegraphics[width=0.31\linewidth]{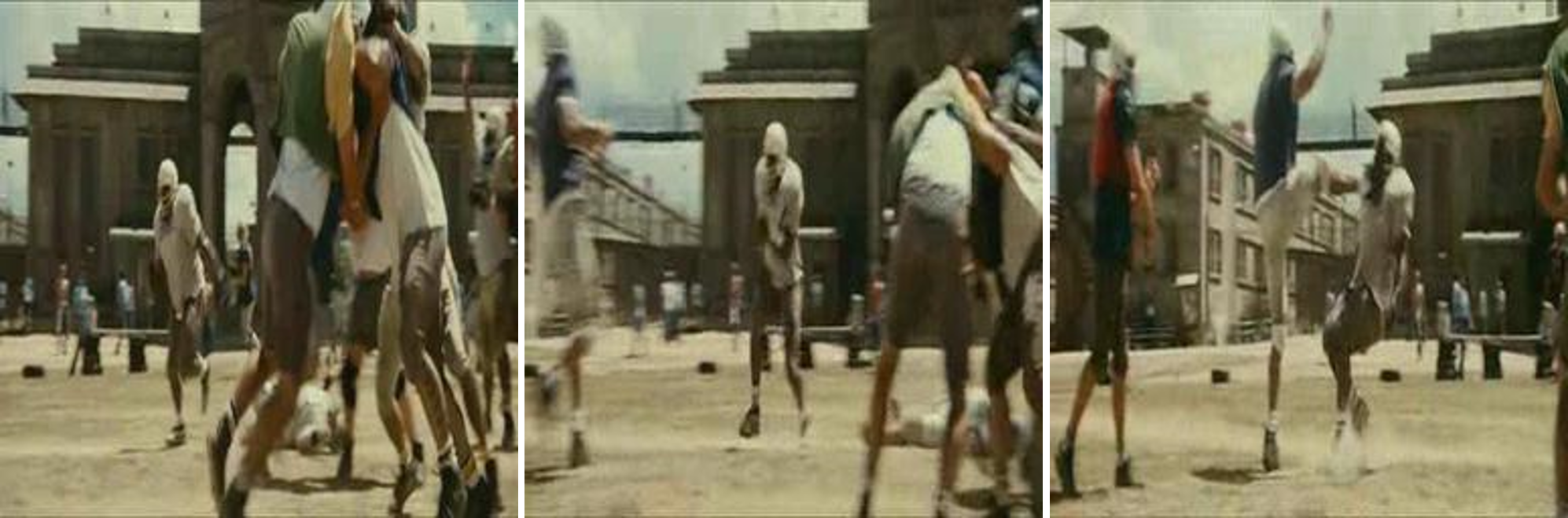}
}
\subfigure[labeled as `Kick'. (Sword?)]{
    \includegraphics[width=0.31\linewidth]{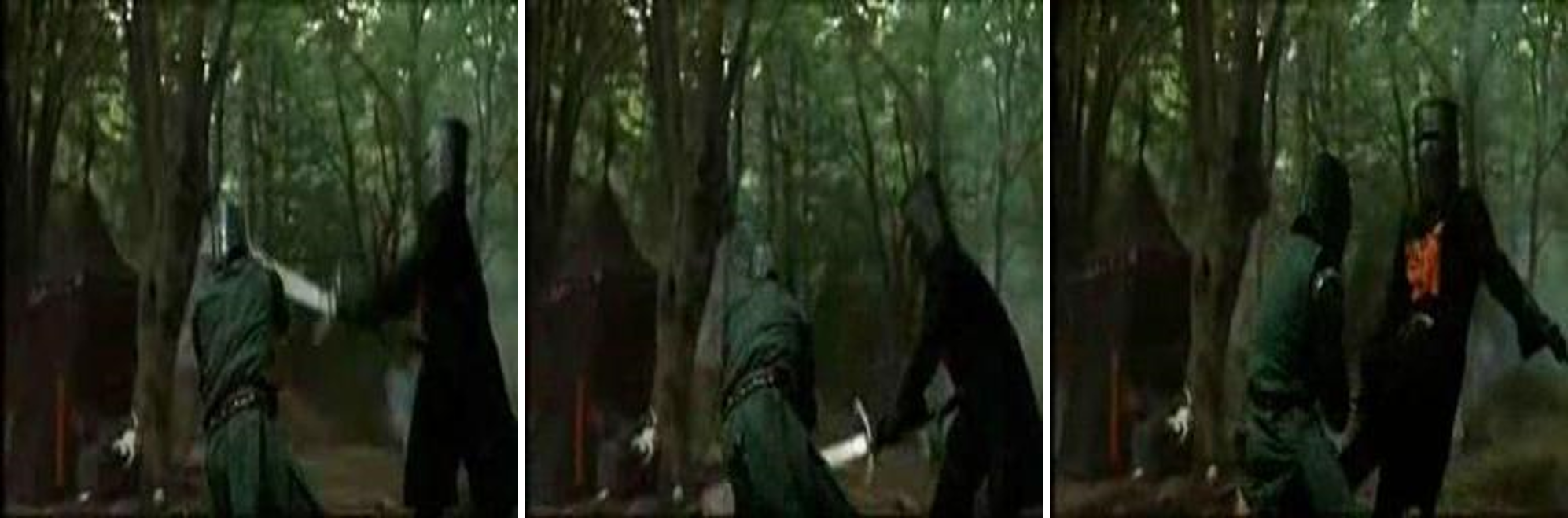}
}
\subfigure[labeled as `Kick'. (Ride bike?)]{
    \includegraphics[width=0.31\linewidth]{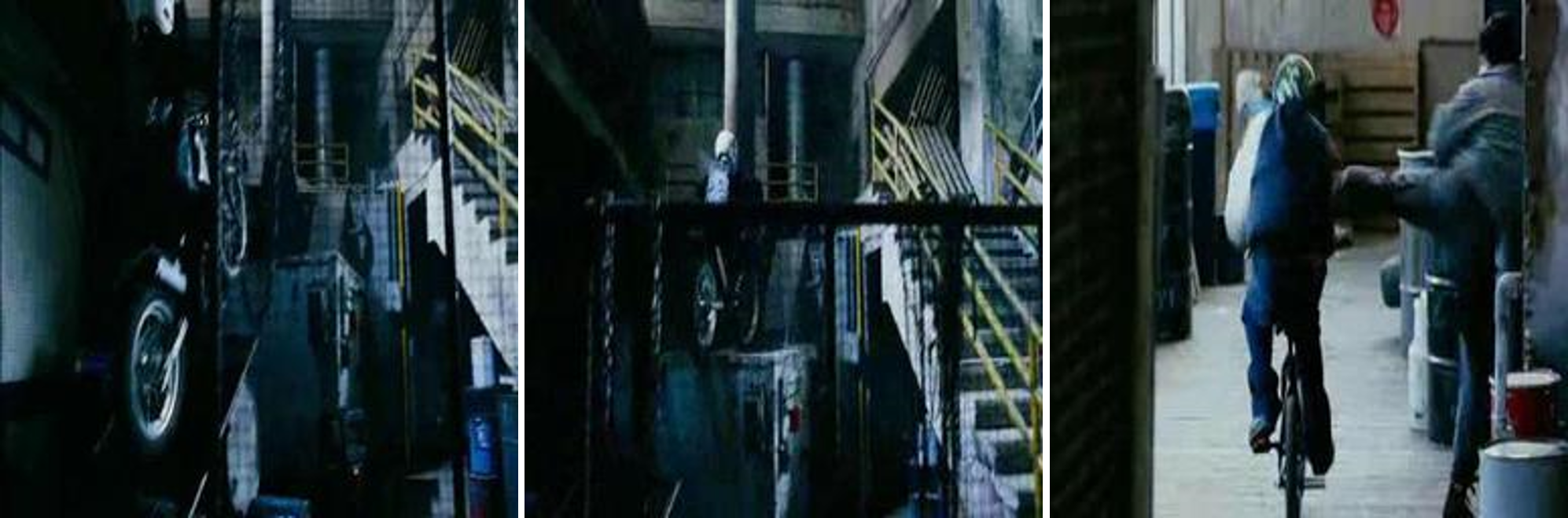}
}
\subfigure[labeled as `Kick'. (Turn?)]{
    \includegraphics[width=0.31\linewidth]{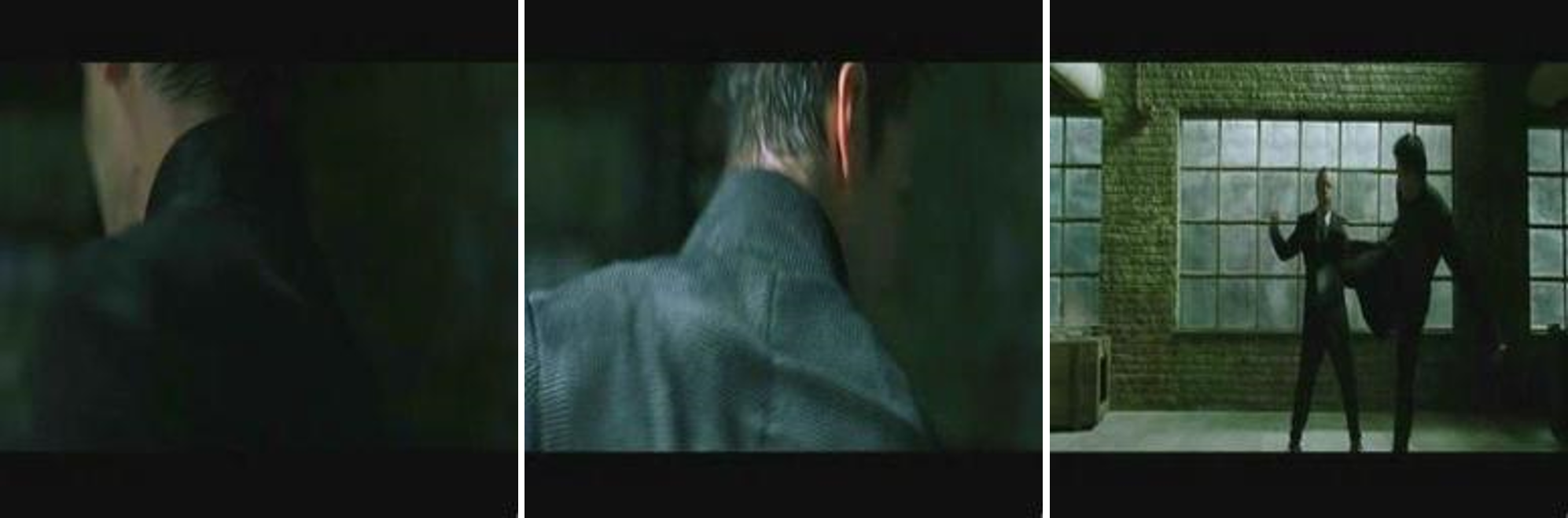}
}
\subfigure[labeled as `Sword'. (Shoot bow?)]{
    \includegraphics[width=0.31\linewidth]{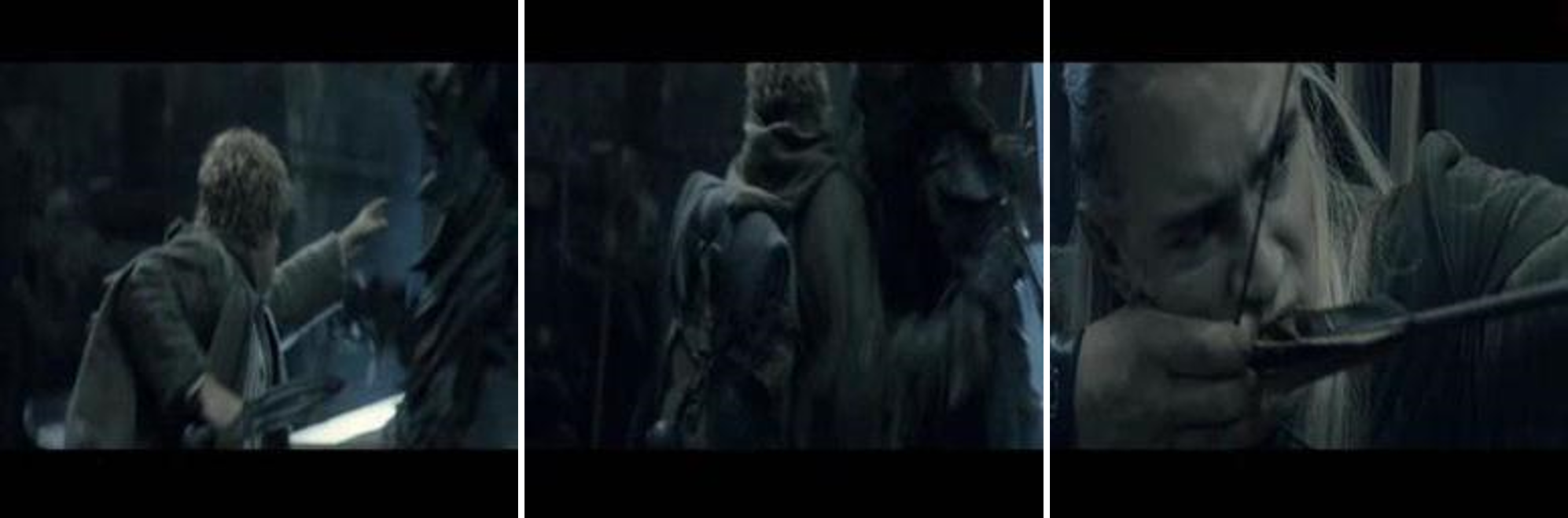}
}
\subfigure[labeled as `Wave'. (Smile?)]{
    \includegraphics[width=0.31\linewidth]{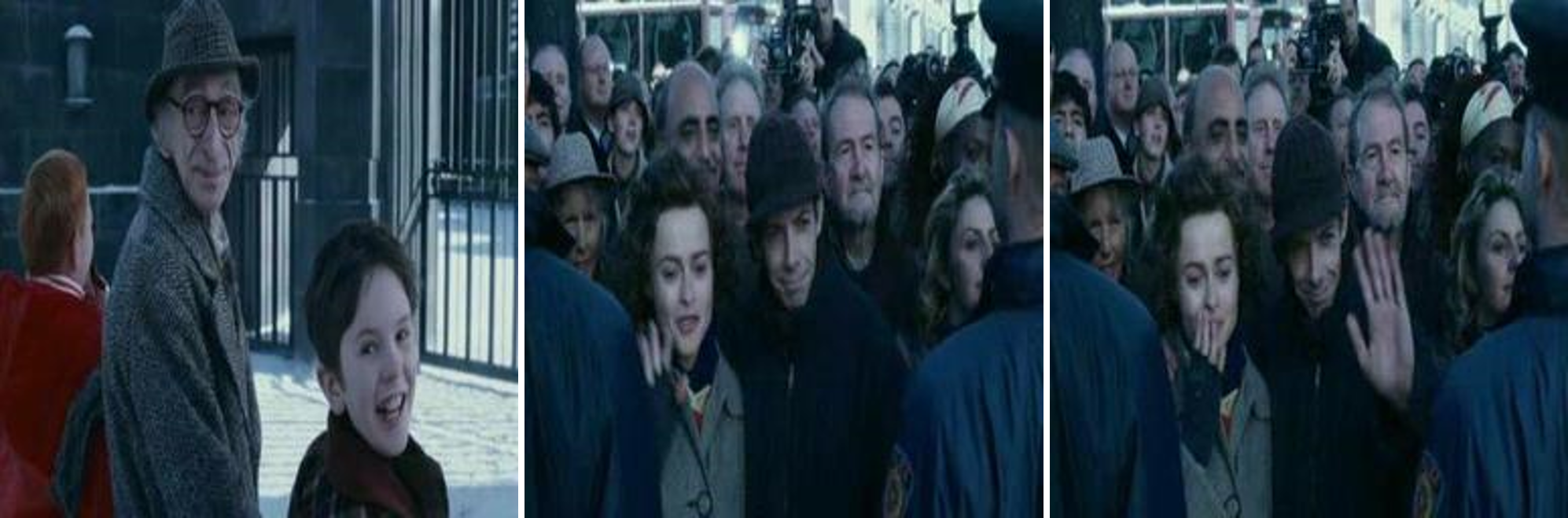}
}
\caption{\textbf{Training examples with the highest ${\cal O(\cdot)}$ (HMDB-51).} Some videos are incorrectly labeled and do not contain any scene corresponding to the label. The other videos are partly noisy and include scenes corresponding to other labels that seem more suitable. The other possible labels are shown in parentheses. (Best viewed magnified on screen.)}
\label{fig:noisy_video_example}
\end{figure*}

\subsection{Experimental results after one-round}\label{appendix:sec:oneround}
In this section, we present the results after one round of \our on CIFAR in Table~\ref{table:cifar-symm-oneround}, and on CIFAR-N in Table~\ref{table:cifar-n-oneround}.
We can observe that applying only one round of \our can considerably improve the classification accuracy. 
Thus, when time budget is limited, applying \our for once can be sufficient.

\section{Detector for Data Cleaning on Real-world Video Data}
\label{appendix:sec:video}

A few seconds of a video consists of a sequence of frames, ranging from tens to hundreds of consecutive images. 
Therefore, in general, when predicting the action class, frames are sampled and predicted for each frame. Then, the prediction scores of sampled frames are averaged and the action with the highest prediction score is determined as the final action class.

Consider a video action recognition task with $n$ training videos $(v_1, y_1), \cdots , (v_n, y_n)$, where $v_i$ is the $i$th video and $y_i$ is its label. 
Let $m_i$ be the number of sampled frames in the $i$th video, and $x_{ij}$ be the $j$th frame in the $i$th video. 
Then, the empirical risk for the video dataset is given by $R(\theta) = \frac{1}{n} \sum_{i=1}^n (\frac{1}{m_i} \sum _{j=1}^{m_i} \ell(y_i, f(x_{ij},\theta)))$, where $\ell(x_{ij}, \theta)$ is the loss for a frame $x_{ij}$.
Now, when we denote the loss of a video $v_i$ as $\ell(y_i, f(v_i, \theta)) = \frac{1}{m_i} \sum _{j=1}^{m_i} \ell(y_i, f(x_{ij},\theta))$, the empirical risk can be rewritten as $R(\theta) = \frac{1}{n} \sum_{i=1}^n \ell(y_i, f(v_i, \theta))$.
Given the empirical risk $R(\theta) = \frac{1}{n}\sum_{i=1}^n \ell({y}_i, f(v_i, \theta))$, the fully optimized\,(overfitted) model parameters $\hat{\theta}$ minimizes the given empirical risk $R(\theta)$ as $\hat{\theta} \eqdef  \text{argmin}_{\theta} R(\theta)$.
Then, a new parameter when removing the video $v$ is derived as
$\hat{\theta}_{v, \omega} \eqdef  \text{argmin}_{\theta} R(\theta) + \omega \ell(y, f(v, \theta))$.
Then, we can use equation~({\color{red}1}) by definition.
Therefore, a video in the video action recognition task can be easily mapped to an image in the image classification problem, and we can simply use the equations derived in this paper for the video dataset.

In this paper, we used Temporal Segment Networks (TSN)~\cite{ref:Xiong_Gool_2016}, which is one of the representative video action recognition models.
We train the networks based on public code by Xiong \footnote{\url{https://github.com/yjxiong/tsn-pytorch}}, without changing the given hyperparameter settings, except the addition of a hidden layer as explained in \ref{appendix:sec:implementation}. 
We train the network for 300 epochs. 
We set the initial learning rate to 0.001 and drop it by a factor of 0.1 after 30 and 200 epochs. 
To deal with both spatial information and long-range temporal structure, TSN adopts two-stream networks that each network processes an RGB image and the stacked optical flows \cite{ref:opticalflow}, respectively.
Therefore, we compute ${\cal O} (v_i;\hat{\theta})$ for both networks and analyze the commonly influential video clips from both networks. 
We present examples of the detected noisy-label videos in Figure \ref{fig:noisy_video_example}.

% \section{Ethics Statement} \label{appendix:sec:limitation}
\noindent\textbf{Limitation.}
Although our main baseline, RoG~\cite{ref:rog_2019_icml}, also uses validation set to optimize generative classifiers for ensemble, a limitation of our method is that it requires a small number of clean validation samples to calculate the OSD.
It can be difficult to collect image data in some domains despite the limited number of 5 images per class in our experiments.

% A limitation of our method is that it requires a Hessian to calculate the proposed overfitting scores.
% To reduce the computational burden, we reduce the number of parameters, sample the training data, and use optimization methods as mentioned in ~\ref{appendix:sec:hessian}.
% However, we believe that the recent advance on computing power and data parallel processing can solve the problem. 
% Future work could be designed to develop overfitting scores that do not use a Hessian.

% \noindent\textbf{Potential negative societal impact.}
% Since our method can detect mislabeled images from data crawling, the cost of collecting data can be decreased and thus more data can be easily collected.
% Training on larger data can require more computational power, which can accelerate global warming.

{\small
\bibliographystyle{ieee_fullname}
\bibliography{egbib}
}

\end{document}